\documentclass[11pt]{article}
\pdftrailerid{}
\usepackage[T1]{fontenc}
\usepackage[utf8]{inputenc}
\usepackage{lmodern,microtype}
\usepackage[margin=1in]{geometry}
\usepackage{amsmath,amssymb,amsthm,mathtools,bm}
\usepackage{enumitem,booktabs,array}
\usepackage[numbers,sort&compress]{natbib}
\usepackage{xcolor}
\usepackage{algorithm}
\usepackage[noend]{algpseudocode}
\usepackage[colorlinks=true,linkcolor=blue!45!black,citecolor=blue!45!black,urlcolor=blue!45!black,bookmarksnumbered=true]{hyperref}
\usepackage[capitalise,nameinlink]{cleveref}
\hypersetup{
  pdftitle={Learning from First Discrepancies with Joint Reset and Pilot Guarantees},
  pdfauthor={Michal Certik},
  pdfsubject={Statistical learning theory; stopped feedback},
  pdfkeywords={first-discrepancy feedback, proper learning, VC dimension, exploration, depth-robust graphs}}

\numberwithin{equation}{section}
\newtheorem{theorem}{Theorem}[section]
\newtheorem{lemma}[theorem]{Lemma}
\newtheorem{proposition}[theorem]{Proposition}
\newtheorem{corollary}[theorem]{Corollary}
\theoremstyle{definition}
\newtheorem{definition}[theorem]{Definition}
\newtheorem{assumption}[theorem]{Assumption}
\theoremstyle{remark}
\newtheorem{remark}[theorem]{Remark}
\newtheorem{example}[theorem]{Example}
\crefname{assumption}{Assumption}{Assumptions}
\crefname{example}{Example}{Examples}
\crefname{algorithm}{Procedure}{Procedures}
\Crefname{algorithm}{Procedure}{Procedures}
\floatname{algorithm}{Procedure}
\algrenewcommand\algorithmicrequire{\textbf{Input:}}
\algrenewcommand\algorithmicensure{\textbf{Output:}}

\newcommand{\E}{\mathbb E}
\newcommand{\1}{\mathbf 1}
\newcommand{\eps}{\varepsilon}
\newcommand{\cH}{\mathcal H}
\newcommand{\cZ}{\mathcal Z}
\newcommand{\cP}{\mathcal P}
\newcommand{\cG}{\mathcal G}
\newcommand{\cN}{\mathcal N}
\newcommand{\Ot}{\widetilde O}
\newcommand{\KL}{\operatorname{KL}}
\newcommand{\TV}{\operatorname{TV}}
\newcommand{\Ber}{\operatorname{Ber}}
\newcommand{\Bad}{\mathrm{Bad}}
\newcommand{\Vis}{\mathrm{Vis}}

\newcommand{\nat}{\mathrm{nat}}
\newcommand{\Lam}{\Lambda_b}
\newcommand{\CEG}{C_{\mathrm{EG}}}
\newcommand{\Csp}{B_{\mathrm{sp}}}
\newcommand{\epsb}{\bar\varepsilon}

\AddToHook{env/theorem/begin}{\crefalias{theorem}{theorem}}
\AddToHook{env/lemma/begin}{\crefalias{theorem}{lemma}}
\AddToHook{env/proposition/begin}{\crefalias{theorem}{proposition}}
\AddToHook{env/corollary/begin}{\crefalias{theorem}{corollary}}
\AddToHook{env/definition/begin}{\crefalias{theorem}{definition}}
\AddToHook{env/assumption/begin}{\crefalias{theorem}{assumption}}
\AddToHook{env/remark/begin}{\crefalias{theorem}{remark}}
\AddToHook{env/example/begin}{\crefalias{theorem}{example}}
\AddToHook{cmd/appendix/after}{\crefalias{section}{appendix}}
\setlist{itemsep=2pt,topsep=4pt}

\title{Learning from First Discrepancies with\\ Joint Reset and Pilot Guarantees}
\author{Michal Certik\\[2pt] The World bank Group\\[2pt] \texttt{michal.certik@gmail.com}}
\date{November 2026}

\begin{document}
\AddToHookNext{shipout/foreground}{%
  \put(0,-36){\makebox[\paperwidth]{\small Code archive DOI: \href{https://doi.org/10.5281/zenodo.24104680}{10.5281/zenodo.24104680}}}%
}
\maketitle

\begin{abstract}
A run that stops at its first wrong decision shows the queries and their correct labels up to and including that decision, and nothing after it. We study how many such runs are needed for learning. Each run tests a hypothesis, the pilot, chosen beforehand from a fixed binary class of VC dimension at most $d$, on a fresh record of at most $D$ queries drawn from one unknown distribution that does not depend on the pilot. The terminal information of a record is shown only if the pilot makes no discrepancy, and a bounded loss, fixed in advance, is attached to the position where a hypothesis stops. If the discrepancy probabilities of members of the class have infimum at most $f$ and their risks have infimum at most $r$, one learner uses $\Ot((d+1)[D(f+\eps)+r]/\eps^2)$ runs and $\Ot(D(1+f/\eps))$ distinct pilots on every history. It returns a member of the class and an interval of width at most $\eps$ containing, with probability at least $1-\delta$, both the output's risk and the optimal risk. We also treat unknown ceilings and expected costs. Lower bounds for finite classes show that each order is necessary in the worst case, possibly on different instances, at positive dimension, small accuracy and fixed confidence. The learner enumerates label patterns of the class on displayed queries, possibly in exponential time; the results, under standard measurability conditions, concern statistical resources only. The proof combines a gain computable from stopped displays, exponential weights, a sparse depth-robust graph and a harmonic-mean bound.
\end{abstract}

\section{Introduction}\label{sec:intro}

Many checking procedures go through a sequence of items and stop at the first item that is handled incorrectly. When a candidate decision rule is run on such a sequence, the run reveals the items and their correct labels up to and including the first item on which the rule is wrong; the later items, and any outcome attached to the end of the sequence, remain hidden. This paper studies how to learn a good rule when this is the only feedback. The learner must choose which rule to run before it sees the sequence, and an early mistake hides everything that follows. Even when every revealed label is correct, minimizing the empirical error on the prefixes observed so far need not produce a good rule, because the behaviour that matters for the loss of a candidate may lie beyond those prefixes, which were determined by the rules that happened to be run. Learning therefore requires deliberate exploration.

We formalize this as follows (\cref{sec:model} gives the precise model). A fresh run uses a record of at most $D$ queries, drawn from one unknown distribution, independently of the rule that is run on it. The rule, called the \emph{pilot}, is a member of a fixed binary class $\cH$ of VC dimension at most $d$; it is selected before the record is drawn and does not change during the run. The run displays the queries and their true labels through the pilot's first discrepancy, inclusive; if the pilot makes no discrepancy, it also displays a terminal descriptor of the record. A bounded loss, fixed before learning, is attached to the position at which a hypothesis stops: a hypothesis that errs at query $j$ pays a loss that depends on the first $j$ queries and labels, and a hypothesis that survives pays a loss that may depend on the whole record and its descriptor. We count the number $N$ of runs, also called resets, and the number $V$ of distinct members of $\cH$ that are actually used as pilots during training. The second count measures exploration: every rule played while a cover is being acquired counts, not only the members of a final mixture.

\paragraph{Two optima.}
Two infima govern the problem:
\[
 F^*=\inf_{h\in\cH}\Pr(h\text{ errs on the record}),\qquad
 R^*=\inf_{h\in\cH}\E[\text{loss of }h\text{ where it stops}].
\]
Neither needs to be attained, and they may be approached by different hypotheses. The first controls how much of a record a good pilot can reveal; the second controls the variance of the final estimate. Given numbers $f,r$ with $F^*\le f$ and $R^*\le r$, our main result (\cref{thm:main}) is a learner whose numbers of resets and distinct pilots are at most
\begin{equation}\label{eq:intro-rate}
 N=\Ot\!\left((d+1)\,\frac{D(f+\eps)+r}{\eps^2}\right),
 \qquad
 V=\Ot\!\left(D\left(1+\frac f\eps\right)\right)
\end{equation}
on every history. Here $\Ot$ hides universal constants and polylogarithmic factors in $D$, $d+1$, $1/\eps$ and $1/\delta$. On every history the learner returns a member $\widehat h\in\cH$ and an interval of width at most $\eps$, computed from the observations; with probability at least $1-\delta$ the interval contains both $R^*$ and the risk $R(\widehat h)$ of the output, so in particular $R(\widehat h)\le R^*+\eps$.

When no ceilings are supplied, a second learner (\cref{thm:adaptive}) attains the same orders with $F^*$ and $R^*$ in place of $f$ and $r$, on the same event as the coverage guarantee; on every history it satisfies the coarser caps $\Ot((d+1)D/\eps^2)$ and $\Ot(D/\eps)$, and with a smaller internal failure level the local orders also hold in expectation. For the natural discrepancy loss, which charges one unit for a discrepancy and nothing for survival, $R^*=F^*$, so \eqref{eq:intro-rate} holds with $f=r$ (\cref{cor:natural}); in particular the reset scale is $\Ot((d+1)D/\eps)$ when the natural infimum is zero, even if it is not attained.

Both orders in \eqref{eq:intro-rate} are necessary in the worst case (\cref{thm:lower}): for every $d,D\ge1$, accuracy $\eps\le1/16$ and success probability $4/5$, there are finite classes of VC dimension exactly $d$, record laws and fixed losses satisfying the two ceilings on which every successful learner has expected reset count $\Omega(d[D(f+\eps)+r]/\eps^2)$ in some instance, and others on which it has expected distinct-pilot count $\Omega(D(1+f/\eps))$ in some instance. These are coordinatewise ceiling-minimax statements at fixed confidence, not a trade-off between $N$ and $V$ and not a lower bound for every class. Natural discrepancy has a separate statement under its single ceiling $F^*=R^*\le r$ (\cref{thm:natural-lower}), with an exception at $D=1$, where every query and its label are displayed whatever pilot is used and one fixed training pilot suffices.

\subsection{How the learner works}\label{sec:intro-overview}

The construction has six steps; the arguments specific to stopped feedback are in Steps~2--6.

\begin{enumerate}[label=\arabic*.,leftmargin=*]
\item \emph{A coarse anchor.} The learner first finds a member $a$ of $\cH$ and a number $w=O(f+\eps)$ such that, if $f$ is a valid ceiling, $F(a)\le w$ on a success event. It uses an ownership argument: the loss of a candidate on a record is assigned to the first pilot, in order of use, that reaches the candidate's stopping position. When the numerical inputs of an ownership call are valid, a bounded potential limits the number of failed validations on a concentration event; on every history a cap on the number of candidates bounds the cost (\cref{sec:anchor}, \cref{app:anchor}).
\item \emph{A legal gain at truncated ranks.} Fix a frozen mixture $L$ of already available members, with weight at least $1/2$ on $a$; we call it a logger. We define, for every candidate $h$, a score that can be computed from one display of a pilot drawn from $L$. Its conditional mean is a potential evaluated at the candidate's stopping rank, truncated one position after the last rank that $L$ reaches with probability at least a threshold $\zeta$. The key point is that a displayed pair at position $j$ is used to estimate an increment of the potential at the virtual position $j+1$. This is legal because the discrepancy pair itself is displayed (\cref{lem:gain}).
\item \emph{An observed dictionary.} The class may be infinite. The learner plays $L$ on fresh resets and calls the pattern oracle on the queries that were actually displayed. A VC missed-set argument shows that every member of $\cH$ has a representative in the resulting finite dictionary whose truncated rank differs from its own with small probability.
\item \emph{Exponential weights and one sparsification.} Classical exponential weights run numerically on the dictionary, while fresh data still come from the frozen logger $L$. A single random sparsification turns the averaged weights into a small lottery $p$, and a single transfer from the dictionary to the whole class yields a bound on the potential of $p$ at the truncated rank of every member of $\cH$ (\cref{lem:acquisition}).
\item \emph{Propagation along a robust graph.} The four steps above learn a lottery that is good only up to the logger's visibility frontier. To reach the true stopping ranks without paying a polynomial factor in $D$, we run $S=4D$ acquisition stages arranged along a sparse depth-robust directed graph of polylogarithmic indegree. The logger of a stage mixes the lotteries of its graph predecessors. A successful predecessor moves the frontier one position further. Deleting any set of unsuccessful stages leaves a long path, so the argument allows the failures of different stages to be arbitrarily correlated. On most records, a constant fraction of the stages see the true stopping rank of a given candidate (\cref{lem:visibility}).
\item \emph{Harmonic recovery and a final estimate.} The final lottery puts weight $1/2$ on $a$ and spreads the rest uniformly over the $S$ stages. A naive bound would lose the factor $S$. The harmonic--arithmetic mean inequality over the visible stages recovers it (\cref{lem:final-certificate}). Inverse-propensity estimates on fresh runs of this frozen lottery, a uniform relative deviation bound, proper empirical minimization over the patterns on the displayed queries, and explicit clipping then give the output and its interval (\cref{prop:precision}).
\end{enumerate}

The depth-robust graph, the exponential-weights optimizer, inverse-propensity estimation, the rank-tail inequality and the VC arguments are established tools, and we do not claim them as new. What the proof establishes is that they can be combined using only the observations available in this experiment, while counting every reset and every played function. Complete latent records appear only in the proofs; the learner sees a whole record only when its pilot survives, and has no other source of complete records.

\subsection{Relation to the closest work}

Two earlier results are especially close. Li and Zhang \citep{li_zhang_v1} give a fast trajectory-wise learner with first-step mixtures for a supplied finite evaluable class in the attained-realizable setting; an absorbing-state reduction turns it into a legal stopped-feedback baseline with a single proper output. Balcan, Blum, Li and Sharma \citep{balcan_v1} already give proper agnostic learning of verifiers with coefficient one, and its extension to ordinary VC classes, when complete annotated traces are supplied. Neither properness, agnostic competition nor finite VC dimension is therefore the distinguishing feature here. What we add is the acquisition of the missing trace information when the pilot's own earlier discrepancy censors the record, including its terminal information, under original-proper frozen pilots and with both training resources controlled. \Cref{sec:related} compares the exact interfaces, including an earlier proper policy search with resets under a mistake-bound assumption \citep{mansour1999} and a legal empirical specialization of the planning engine of Amortila, Foster and Krishnamurthy \citep{amortila2024}.

\subsection{Scope and organization}

The results concern statistical resources under a strict observation interface: one common record law that does not depend on the pilot, exact and fixed labels, a pilot frozen for the whole run with no prefix resets and no continuations after an error, and an oracle that returns representatives of the patterns of $\cH$ on displayed queries and may need exponential time. \Cref{sec:model,sec:results} give the model and the results, \cref{sec:anchor,sec:stage,sec:composition} prove the main upper bound, \cref{sec:lower-mechanisms} explains the lower-bound constructions, \cref{sec:related} discusses related work and \cref{sec:limitations} the assumptions and constants; the appendices contain the remaining proofs.

\section{The experiment}\label{sec:model}

\subsection{Records, pilots and displays}

\begin{definition}[Records]\label{def:records}
Let $\cZ$ be a measurable query space and $y:\cZ\to\{0,1\}$ a measurable truth function. Fix a known integer $D\ge1$. A \emph{record} is
\[
 T=\bigl((z_1,y_1),\ldots,(z_M,y_M),A\bigr),\qquad 0\le M\le D,
\]
where $z_j\in\cZ$, $y_j=y(z_j)$, and $A$ is a terminal descriptor taking values in a measurable space. A \emph{world} consists of the truth function $y$ and a probability distribution $\cP$ on records with $y_j=y(z_j)$ almost surely. Both are unknown to the learner.
\end{definition}

Queries in a record may repeat, may depend on the earlier labels, and the length $M$ and the descriptor $A$ may depend arbitrarily on the queries. The truth is exact: the label of a query is a fixed function of the query. The descriptor carries the information that is available only at the end of a record, for instance the outcome of a completed task.

Let $\cH$ be a nonempty class of functions $\cZ\to\{0,1\}$ whose VC dimension is at most a known integer $d\ge0$. For a function $g:\cZ\to\{0,1\}$ and a record $T$ define the \emph{stopping rank}
\[
 \tau_g(T)=\min\{j\le M:\ g(z_j)\ne y_j\},
\]
with $\tau_g(T)=M+1$ if the set is empty. We say that $g$ \emph{errs} at $\tau_g$ if $\tau_g\le M$, and that $g$ \emph{survives} if $\tau_g=M+1$.

\begin{definition}[Display]\label{def:display}
The \emph{display} of $g$ on $T$ consists of the pairs $(z_j,y_j)$ for $j\le\min(\tau_g(T),M)$, together with an end tag and the descriptor $A$ if and only if $\tau_g(T)=M+1$. We write $n_g=\min(\tau_g,M)$ for the number of displayed pairs.
\end{definition}

Two cases deserve attention. A discrepancy at the last query, $\tau_g=M$, displays the pair $(z_M,y_M)$ but neither the end tag nor $A$; in general the display does not reveal that the record has ended (it does when $M=D$, since $D$ is known). On an empty record ($M=0$) every function survives with $\tau_g=1$, and the display consists of the end tag and $A$.

\begin{definition}[Interaction]\label{def:interaction}
A learner proceeds in resets $t=1,2,\ldots$. Before reset $t$ it selects a pilot $g_t\in\cH$ as a measurable function of the earlier displays and of its private randomness. (The selection may consist in sampling from a finite distribution on $\cH$; the sampled function is fixed before the record is drawn.) Then a fresh record $T_t\sim\cP$ is drawn, independently of everything before, and the learner observes the display of $g_t$ on $T_t$. After any reset the learner may stop; $N$ denotes the number of resets. At the end it returns $\widehat h\in\cH$ and an interval, both computed from its observations and private randomness.
\end{definition}

No other access is available. In particular there is no access to the hidden suffix of a record, no continuation after a discrepancy, no reset to an arbitrary prefix, no change of the pilot within a run, no oracle for $\cP$, and no deployment oracle. The record law does not depend on the pilot or on the private coins. This common law is what allows the analysis to evaluate the stopping ranks of several hypotheses on the same fresh reference record; such a reference record is an object of the analysis, not a source of observations.

\subsection{Losses and risks}

\begin{definition}[Prefix-observable loss]\label{def:loss}
A \emph{loss} is a known measurable function $\lambda$ assigning to a record $T$ and a rank $j\in\{1,\ldots,M+1\}$ a value $\lambda(T,j)\in[0,1]$, such that for $j\le M$ the value $\lambda(T,j)$ is a function of the first $j$ pairs $(z_1,y_1),\ldots,(z_j,y_j)$, while $\lambda(T,M+1)$ may depend on the whole record including $A$. The loss is fixed before learning.
\end{definition}

For $h\in\cH$ write
\[
 \ell_h(T)=\lambda(T,\tau_h(T)),\qquad R(h)=\E\,\ell_h(T),\qquad F(h)=\Pr(\tau_h(T)\le M),
\]
\[
 R^*=\inf_{h\in\cH}R(h),\qquad F^*=\inf_{h\in\cH}F(h).
\]
The dependence on the candidate enters only through its stopping rank. The \emph{natural discrepancy loss} is $\lambda_\nat(T,j)=\1\{j\le M\}$; for it $R=F$. The learner may use the natural loss as an auxiliary objective, for instance to find an anchor, even when the target loss $\lambda$ is different.

The following observation is used many times. Let $g$ be a pilot and $h$ any function whose values at the displayed queries are known.
\begin{itemize}
\item If $h$ disagrees with some displayed pair, then its first such disagreement is $\tau_h\le n_g$, and $\ell_h(T)=\lambda(T,\tau_h)$ is determined by the displayed pairs.
\item If $g$ survives, the whole record is displayed, so $\tau_h$ and $\ell_h(T)$ are determined.
\item If $g$ errs and $h$ agrees with every displayed pair, then $h$ is correct at $\tau_g$, so $\tau_h>\tau_g$; in general $\ell_h(T)$ is not determined.
\end{itemize}

\subsection{The pattern oracle and regularity}

\begin{assumption}[Observed-pattern oracle]\label{ass:oracle}
For every finite list $(z_1,\ldots,z_q)$ of queries that have actually been displayed, the learner can obtain a finite set $\cG\subseteq\cH$ containing, for every $h\in\cH$, exactly one $h'\in\cG$ with $h'(z_i)=h(z_i)$ for $1\le i\le q$. For $q=0$ the set is a single member of $\cH$. The returned members are evaluable at every query, and their selection may depend on the list. A call draws no record and plays no function.
\end{assumption}

The oracle is stronger than an empirical risk minimizer: it lists one representative for each label pattern that $\cH$ realizes on the displayed queries. It supplies nothing on queries that have not been displayed, and it may require exponential time.

\begin{assumption}[Measurability]\label{ass:measurable}
The evaluation maps $(h,z)\mapsto h(z)$ of the functions used, including the evaluation of functions selected from data at later queries, the loss, the oracle's selections and finite tie-breaks, and the uniform events and witness selections in all VC arguments of this paper (\cref{app:vc,app:anchor,app:acquisition,app:estimation,app:lower}) are measurable, so that all events used have ordinary probabilities.
\end{assumption}

\Cref{ass:measurable} is the usual measurable-VC convention. It does not follow from finite VC dimension and pointwise measurability alone.

\subsection{Resources and the required output}

\begin{definition}[Resources]\label{def:resources}
$N$ is the total number of resets, over all phases of the learner: anchor acquisition, validation, design acquisition and final estimation. $V$ is the number of distinct functions in $\{g_1,\ldots,g_N\}$, that is, the size of the union of all functions actually played as pilots during training.
\end{definition}

Computing with a finite distribution over $\cH$, enumerating a dictionary and privately drawing the identity of a function play nothing by themselves. A function counts towards $V$ when it is first actually used as a pilot, whatever other roles it has; it may, for instance, also belong to a dictionary. The returned $\widehat h$ need not have been played. The number of functions from which a procedure can draw its pilots, for instance the anchor together with all slots of all designs in a precision stage, bounds the number of distinct functions it plays; the support of the final lottery alone need not.

\begin{definition}[Joint certificate]\label{def:certificate}
For $\eps\in(0,1]$, an \emph{$\eps$-certificate} is a pair $(\widehat h,\widehat I)$ where $\widehat h\in\cH$ and $\widehat I=[\widehat l,\widehat u]$ with $0\le\widehat l\le\widehat u\le1$ and $\widehat u-\widehat l\le\eps$, both computed from the observations. It \emph{covers} if $R^*\in\widehat I$ and $R(\widehat h)\in\widehat I$.
\end{definition}

A covering $\eps$-certificate gives $R(\widehat h)\le R^*+\eps$, and it also locates $R^*$ within $\eps$. Our learners return an $\eps$-certificate on every history; only the covering property is probabilistic.

\begin{example}[A censored terminal]\label{ex:censored}
Let every record consist of a gate query, a second query and a descriptor. A pilot that errs at the gate displays the gate and its label and nothing else. A pilot that errs at the second query displays both pairs, but not the descriptor. A candidate that survives both queries can have a loss that depends on the descriptor. Thus, already at horizon two, knowing every observed label does not give an unbiased sample of the loss of every candidate.
\end{example}

\paragraph{Conventions.}
A \emph{lottery} is a finitely supported probability distribution on $\cH$, and $\delta_h$ is the point mass at $h$. Logarithms are natural. A statement holds \emph{on every history} if it holds for every realization of the records and of the private randomness. We write $\Ot(\cdot)$ for an upper bound up to a universal constant and a factor that is polynomial in $\log(eD)$, $\log(e(d+1))$, $\log(e/\eps)$ and $\log(e/\delta)$; it never hides a polynomial factor in $d$ or $D$. Every statement concerns the supplied horizon $D$; we do not assert one confidence event that holds simultaneously for all possible horizons.

\section{Results}\label{sec:results}

\subsection{Upper bounds}

\begin{theorem}[Joint guarantees with supplied ceilings]\label{thm:main}
Let $0<\eps\le1$, $0<\delta<1$ and $f,r\in[0,1]$. There is a learner, receiving $d,D,\eps,\delta,f,r$, with the following properties.
\begin{enumerate}[label=(\alph*)]
\item On every history it returns an $\eps$-certificate $(\widehat h,\widehat I)$ and uses
\[
 N=\Ot\!\left((d+1)\,\frac{D(f+\eps)+r}{\eps^2}\right)
 \quad\text{resets and}\quad
 V=\Ot\!\left(D\left(1+\frac f\eps\right)\right)
 \quad\text{distinct pilots.}
\]
\item If $F^*\le f$ and $R^*\le r$, then with probability at least $1-\delta$ the certificate covers: $R^*\in\widehat I$ and $R(\widehat h)\in\widehat I$. In particular $R(\widehat h)\le R^*+\eps$.
\end{enumerate}
The counts include every reset and every function played in all phases, including anchor acquisition, exploration, all intermediate loggers and final estimation.
\end{theorem}

The bounds in (a) hold on every history: the sample sizes of the final precision stage are fixed numerical functions of $(d,D,\eps,\delta,f,r)$, and those of the anchor, which depend on observed endpoints, are bounded by such functions. The proof does not use that $f$ and $r$ are valid ceilings; validity is used only for the covering property~(b).

\begin{theorem}[Unknown ceilings and expected costs]\label{thm:adaptive}
Let $0<\eps\le1$ and $0<\delta<1$. There is a learner, receiving $d,D,\eps,\delta$, with the following properties.
\begin{enumerate}[label=(\alph*)]
\item On every history it returns an $\eps$-certificate and uses at most $\Ot((d+1)D/\eps^2)$ resets and $\Ot(D/\eps)$ distinct pilots.
\item With probability at least $1-\delta$, the certificate covers and, on the same event,
\[
 N=\Ot\!\left((d+1)\,\frac{D(F^*+\eps)+R^*}{\eps^2}\right),
 \qquad
 V=\Ot\!\left(D\left(1+\frac{F^*}{\eps}\right)\right).
\]
\item If the learner is run with internal failure level $\delta'=\min(\delta,\eps^2)$, with all logarithms recalculated at that level, then (a) and (b) hold and in addition $\E N$ and $\E V$ obey the two local orders displayed in~(b).
\end{enumerate}
\end{theorem}

The bounds in (b) hold on the success event, not on every history. Part~(c) is a separate expectation argument based on the caps in~(a); it is not a relabelling of~(b).

\begin{proposition}[Auxiliary accounting]\label{prop:auxiliary}
The learners of \cref{thm:main,thm:adaptive} have the following properties on every history.
\begin{enumerate}[label=(\roman*)]
\item At most $DN$ true labels are displayed.
\item The number of oracle calls is $\Ot(D(1+\log(1/\eps)))$, including the initial call that produces a first member of $\cH$.
\item Within one precision stage (\cref{alg:precision}), the $S=4D$ design dictionaries contain at most $\sum_{s=1}^S(Dm_s+1)^d$ representatives in total, where $m_s$ is the dictionary sample size of design $s$. The final selection uses a separate enumeration with at most $(Dn_F+1)^d$ representatives, where $n_F$ is the final estimation sample size. These numbers are not a bound on the total memory including the anchor work.
\item Within one precision stage, every pilot is the anchor or a sparse slot of one of its designs. Enumerating a dictionary and computing numerical weights play nothing by themselves; a dictionary member can still be played, for instance when it coincides with the anchor or with a slot, and every actual use is counted in $V$.
\item If $\eps=1$, the learners return the representative obtained from one empty-list oracle call together with $\widehat I=[0,1]$, using zero resets and zero training pilots.
\end{enumerate}
No bound on computation time or on the encoding of real numbers is asserted.
\end{proposition}

\begin{corollary}[Natural discrepancy and unit-error losses]\label{cor:natural}
Suppose the loss satisfies $F(h)\le R(h)$ for every $h\in\cH$. This holds for the natural discrepancy loss, and for every loss that charges one unit at each discrepancy. If $R^*\le r$, the learner of \cref{thm:main} with $f=r$ uses on every history
\[
 N=\Ot\!\left((d+1)\,\frac{D(r+\eps)}{\eps^2}\right),
 \qquad
 V=\Ot\!\left(D\left(1+\frac r\eps\right)\right),
\]
and covers with probability at least $1-\delta$. The learner of \cref{thm:adaptive} satisfies these bounds with $r$ replaced by $R^*$, in the sense of parts (b) and (c) of that theorem.
\end{corollary}

\begin{proof}
Taking infima in $F(h)\le R(h)$ gives $F^*\le R^*\le r$, so $f=r$ is a valid ceiling for $F^*$; no common minimizer of $F$ and $R$ is needed. Substituting $f=r$ in \cref{thm:main} and using $D(r+\eps)+r\le 2D(r+\eps)$ gives the bounds. For \cref{thm:adaptive}, use $F^*\le R^*$ in the same way.
\end{proof}

\subsection{Lower bounds}

In the lower bounds a learner is a procedure as in \cref{def:interaction}; it may depend on the known class and loss, and the pilots and the output must belong to the class. An \emph{admissible instance} for parameters $(d,D,f,r)$ consists of a finite class of VC dimension exactly $d$, a world whose records have length at most $D$, and a fixed known prefix-observable loss, such that $F^*\le f$ and $R^*\le r$. The constructions below come in finite families of instances that share a class and a loss rule and differ in the world.

\begin{theorem}[Two-ceiling lower bounds]\label{thm:lower}
Let $d,D\ge1$ be integers, $0<\eps\le1/16$ and $f,r\in[0,1]$. Let a learner return, on every admissible instance, a member $\widehat h$ of the class with $\Pr\{R(\widehat h)\le R^*+\eps\}\ge4/5$. Then
\[
 \sup\E N\ \ge\ c\,d\,\frac{D(f+\eps)+r}{\eps^2},
 \qquad
 \sup\E V\ \ge\ c\,D\left(1+\frac f\eps\right),
\]
where $c>0$ is a universal constant and each supremum is over admissible instances. For the second inequality we assume, as usual, that the learner terminates almost surely on each instance of the finite families used in the proof; every learner with deterministic resource caps does. The suprema may be attained on different families for different summands and for the two resources. Each family uses one loss rule, known and common to all its worlds.
\end{theorem}

\begin{theorem}[Natural discrepancy under its single ceiling]\label{thm:natural-lower}
Let $d,D\ge1$ be integers, $0<\eps\le1/128$ and $r\in[0,1]$. Consider the natural discrepancy loss and finite classes of VC dimension exactly $d$, with the single ceiling $F^*=R^*\le r$. Let a learner satisfy $\Pr\{F(\widehat h)\le F^*+\eps\}\ge4/5$ on every such instance with records of length at most $D$. Then:
\begin{enumerate}[label=(\alph*)]
\item $\sup\E N\ge c\,dD(r+\eps)/\eps^2$;
\item if $D\ge2$, then $\sup\E V\ge c\,D(1+r/\eps)$, under the almost-sure termination convention of \cref{thm:lower};
\item if $D=1$, then for every class satisfying \cref{ass:oracle,ass:measurable} (in particular for the finite classes above), every $\eps\in(0,1]$ and every $\delta\in(0,1)$, there is a learner for the natural discrepancy loss that plays one fixed member of the class on all of its resets and returns an $\eps$-certificate that covers with probability at least $1-\delta$.
\end{enumerate}
Here $c>0$ is a universal constant, and the suprema may again be attained on different families.
\end{theorem}

For the natural loss, two simultaneous ceilings $F^*\le f$ and $R^*\le r$ amount to the single ceiling $\min(f,r)$; \cref{thm:natural-lower} must be applied with that ceiling, not with $r$. In particular, if $f=0<r$, the natural discrepancy infimum is zero (in an infinite class it need not be attained), and \cref{thm:main} applied with both ceilings equal to $\min(f,r)=0$ uses $\Ot((d+1)D/\eps)$ resets; no lower bound of order $r/\eps^2$ holds there.

\begin{remark}[How to read the lower bounds]\label{rem:lower-reading}
The lower bounds use only the excess-risk requirement. An $\eps$-certificate that covers implies it, so they apply to the certificates of \cref{thm:main,thm:adaptive}. Since deterministic caps bound expectations, \cref{thm:main,thm:lower} show that the learner of \cref{thm:main} is coordinatewise ceiling-minimax optimal up to logarithmic factors, at fixed confidence and in the stated range of $\eps$, at positive dimension; part~(c) of \cref{thm:adaptive} gives the same comparison for expected costs without supplied ceilings. The lower bounds assert neither a trade-off between $N$ and $V$ on a common instance, nor a lower bound for every class or every loss, nor sharp dependence on $\delta$, nor anything for $d=0$. A nonempty class of VC dimension zero consists of a single function (two functions that differ at some query shatter that query), so excess-risk learning needs no data, and a certificate only requires estimating one bounded mean using at most one pilot; the upper bounds remain valid for $d=0$ through the factor $d+1$, but are not sharp there. At $D=1$ the general-loss pilot lower bound $\Omega(1+f/\eps)$ remains valid, because a terminal descriptor is still hidden after a discrepancy, whereas for natural discrepancy at $D=1$ one training pilot suffices.
\end{remark}

\Cref{app:adaptation} proves \cref{thm:adaptive,prop:auxiliary}; \cref{sec:composition} proves \cref{thm:main}; \cref{app:lower} proves \cref{thm:lower,thm:natural-lower}.

\section{Propensities, potentials and the anchor}\label{sec:anchor}

This section collects the objects used by the learner and the elementary facts about them. Throughout, $T$ denotes a record with length $M$.

\subsection{Propensities}

\begin{definition}[Propensity]\label{def:propensity}
For a lottery $\pi$ and a rank $l\in\{1,\ldots,M+1\}$ let
\[
 Q_\pi(T,l)=\pi\bigl(\{g\in\cH:\ \tau_g(T)\ge l\}\bigr).
\]
\end{definition}

\begin{lemma}[Properties of propensities]\label{lem:propensity}
For every lottery $\pi$ and record $T$:
\begin{enumerate}[label=(\roman*)]
\item $Q_\pi(T,1)=1$, and $l\mapsto Q_\pi(T,l)$ is nonincreasing on $\{1,\ldots,M+1\}$;
\item $Q_\pi(T,l)$ depends on $T$ only through the pairs $(z_j,y_j)$ with $j<l$, so it can be computed from every display that contains these pairs;
\item $\pi\mapsto Q_\pi(T,l)$ is linear under mixtures of lotteries;
\item if a pilot $g$ is drawn from $\pi$ independently of $T$, then $\Pr(\tau_g\ge l\mid T)=Q_\pi(T,l)$, and for $l\le M$ this is the conditional probability that the pair at position $l$ is displayed.
\end{enumerate}
\end{lemma}

\begin{proof}
We have $\tau_g(T)\ge l$ if and only if $g(z_j)=y_j$ for all $j<l$. This gives (i) and (ii). Part (iii) holds because $Q_\pi(T,l)$ is the $\pi$-measure of a fixed set. For (iv), independence gives $\Pr(\tau_g\ge l\mid T)=\pi(\{g:\tau_g(T)\ge l\})$, and for $l\le M$ the pair at position $l$ is displayed exactly when $l\le\min(\tau_g,M)$, that is, when $\tau_g\ge l$.
\end{proof}

The algorithm uses lotteries in two ways. Some are used to draw pilots. Others are purely numerical: they enter only through propensities in computations and are never used to draw a pilot, although their members may be played in other roles.

\subsection{The potential}

Fix an anchor $a\in\cH$ and a regularizer $b\in(0,1/2]$, and write $\tau_a=\tau_a(T)$. For a lottery $\pi$ define
\[
 \Phi_\pi(T,l)=\frac{\1\{l>\tau_a\}}{Q_\pi(T,l)+b}\qquad(1\le l\le M+1),
 \qquad
 \Lam=\log\frac{1+b}{b}.
\]
Since $\tau_a\ge1$, the potential vanishes for $l\le\tau_a$, in particular at $l=1$. For $\tau_a<l\le M+1$ it is nondecreasing in $l$ by \cref{lem:propensity}(i). It takes values in $[0,1/b]$. If the anchor survives, $\tau_a=M+1$, then $\Phi_\pi(T,l)=0$ for all $l\le M+1$. Finally, for fixed $T$ and $l$, the map $\pi\mapsto\Phi_\pi(T,l)$ is convex, because $x\mapsto1/(x+b)$ is convex and $\pi\mapsto Q_\pi(T,l)$ is linear.

\subsection{Loggers, frontiers and truncated ranks}

\begin{definition}[Logger, frontier, truncated rank]\label{def:logger}
A \emph{logger} is a lottery $L$ with $L(a)\ge1/2$ that is fixed before the resets on which it is used. Fix a threshold $\zeta\in(0,1/2]$. The \emph{frontier} of $L$ on $T$ and the \emph{truncated rank} of $h\in\cH$ are
\begin{equation}\label{eq:frontier}
 J_L(T)=\max\{l\in\{1,\ldots,M+1\}:\ Q_L(T,l)\ge\zeta\},
 \qquad
 \rho_L(h)=\rho_L(h,T)=\min\bigl(\tau_h(T),\,J_L(T)+1\bigr).
\end{equation}
\end{definition}

By \cref{lem:propensity}(i), $\{l:Q_L(T,l)\ge\zeta\}=\{1,\ldots,J_L\}$. Since the anchor survives up to its own stopping rank, $Q_L(T,\tau_a)\ge L(a)\ge1/2\ge\zeta$, hence $J_L\ge\tau_a$. Also $\rho_L(h)\le\tau_h\le M+1$. The frontier and the truncated rank are used only in the analysis; the algorithm never needs their values, only the local tests $Q_L(T,j)\ge\zeta$ at displayed positions $j$.

\begin{lemma}[Rank tail]\label{lem:rank-tail}
For every record $T$, every lottery $\pi$ and every logger $L$,
\[
 \sum_{h}\pi(h)\,\Phi_\pi\bigl(T,\rho_L(h)\bigr)\ \le\ \1\{\tau_a\le M\}\,\Lam,
\]
where the sum runs over the support of $\pi$.
\end{lemma}

\begin{proof}
If $\tau_a=M+1$, every potential vanishes. Assume $\tau_a\le M$ and put $\bar l=\min(J_L+1,M+1)$. Under $h\sim\pi$ the truncated rank takes values in $\{1,\ldots,\bar l\}$. Let $\nu_l=\pi(\rho_L(h)=l)$ and $q_l=\pi(\rho_L(h)\ge l)$ for $1\le l\le\bar l$, and $q_{\bar l+1}=0$. For $l\le\bar l$ we have $l\le J_L+1$, so $\rho_L(h)\ge l$ holds exactly when $\tau_h\ge l$; hence $q_l=Q_\pi(T,l)$. (The convention $q_{\bar l+1}=0$ concerns the truncated rank only; it says nothing about $Q_\pi(T,\bar l+1)$.) Dropping the indicator in $\Phi$ and using $\nu_l=q_l-q_{l+1}$,
\[
 \sum_h\pi(h)\Phi_\pi(T,\rho_L(h))
 \le\sum_{l=1}^{\bar l}\frac{\nu_l}{q_l+b}
 \le\sum_{l=1}^{\bar l}\int_{q_{l+1}}^{q_l}\frac{dx}{x+b}
 \le\int_0^1\frac{dx}{x+b}=\Lam .
\]
The second inequality holds because $1/(x+b)\ge1/(q_l+b)$ on $[q_{l+1},q_l]$, and the third because these intervals have disjoint interiors and lie in $[0,1]=[0,q_1]$.
\end{proof}

The rank-tail inequality is elementary and is related to rank-cover arguments in the literature; we use it as a tool and do not claim it as new.

\subsection{The anchor}\label{sec:anchor-lemmas}

The first phase of the learner finds, on a success event, an anchor with small natural discrepancy probability. The anchor makes most records easy to observe, independently of the target loss. It is obtained by an ownership procedure, proved in full in \cref{app:anchor}. On one record, the loss of a candidate is assigned to the first pilot, in the order of use, whose run reaches the candidate's stopping rank. These assignments are computable from displays, and their predictable means add up to at most the candidate's risk. Consider a call whose numerical input $r$ is a valid risk ceiling, $R^*\le r$. On the joint concentration event of \cref{app:anchor}, a candidate whose validation fails then stops later than all earlier pilots with probability more than a fixed amount, so it increases the potential $\E\max_g\tau_g$ by that amount. This potential exceeds its value for the initial pilot $a$ of the procedure by at most $DF(a)$; so if moreover $F(a)\le w$, not all allowed candidates fail and a candidate is accepted. On every history the number of candidates is capped, and if all allowed tests are exhausted without acceptance, the procedure returns its initial pilot with the interval $[0,0]$.

\begin{lemma}[Anchor with a supplied ceiling; \cref{app:anchor}]\label{lem:anchor-supplied}
Let $0<\eps\le1$, $0<\alpha<1$, $f\in[0,1]$ and $\epsb=\min(1,f+\eps)$. There is a legal procedure that returns a member $a\in\cH$ and the number
\[
 w=\max\bigl(\eps,\min(1,f+\epsb)\bigr),
 \qquad\text{which satisfies}\qquad
 \eps\le w\le1,\quad w\le2(f+\eps).
\]
If $F^*\le f$, then $\Pr\{F(a)\le w\}\ge1-\alpha$. On every history the procedure uses $\Ot((d+1)D/\epsb)=\Ot((d+1)D/(f+\eps))$ resets and $\Ot(D)$ distinct pilots and oracle calls, where the hidden factors are polylogarithmic in $d+1$, $D$, $1/\eps$ and $1/\alpha$. If $\epsb=1$ it uses no reset: it returns an empty-list representative and $w=1$.
\end{lemma}

\begin{lemma}[Anchor without a supplied ceiling; \cref{app:anchor}]\label{lem:anchor-adaptive}
Let $0<\eps\le1$ and $0<\alpha<1$. There is a legal procedure that returns a member $a\in\cH$ and an observable number $w_0\in[\eps,1]$ with
\[
 \Pr\{F(a)\le w_0\le2(F^*+\eps)\}\ge1-\alpha .
\]
On every history it uses $\Ot((d+1)D/\eps)$ resets and $\Ot(D)$ distinct pilots and oracle calls. If its last scale is $t_c$, it uses $\Ot((d+1)D/t_c)$ resets, and on the success event $t_c\ge(F^*+\eps)/5$, so that the reset count is $\Ot((d+1)D/(F^*+\eps))$.
\end{lemma}

When an ownership schedule is executed, its initial pilot and every tested candidate, including an accepted output, are played and counted. The bypasses for $\epsb=1$ in \cref{lem:anchor-supplied} and for $\eps=1$ in \cref{lem:anchor-adaptive} make no reset and return an unplayed representative, which is counted only when it is first used as a pilot (\cref{rem:anchor-accounting}).

\begin{table}[t]
\centering\small
\caption{Principal notation of \cref{sec:anchor,sec:stage,sec:composition}.}\label{tab:notation}
\begin{tabular}{@{}l p{0.70\textwidth}@{}}
\toprule
Symbol & Meaning \\
\midrule
$\tau_h$, $n_g$ & stopping rank of $h$; number $\min(\tau_g,M)$ of pairs displayed by pilot $g$\\
$\ell_h$, $R$, $F$ & loss of $h$ at its stopping rank; risk; natural discrepancy probability\\
$Q_\pi(T,l)$ & propensity: $\pi$-mass of functions that are correct at positions $1,\ldots,l-1$\\
$a$, $w$ & anchor and the numerical bound $F(a)\le w$ used by the learner\\
$b$, $\Phi_\pi$, $\Lam$ & regularizer; potential $\1\{l>\tau_a\}/(Q_\pi(T,l)+b)$; $\log((1+b)/b)$\\
$L$, $\zeta$ & logger (a frozen lottery with $L(a)\ge1/2$); visibility threshold\\
$J_L$, $\rho_L(h)$ & frontier of $L$; truncated rank $\min(\tau_h,J_L+1)$ (analysis only)\\
$\widehat G(h)$, $O_{hh'}$ & displayed gain of $h$; observable disagreement event of $h,h'$\\
$\cG$, $p$ & observed dictionary; sparse lottery returned by an acquisition stage\\
$\CEG$, $C(b)$ & $(2+(e-1)e)/(1-e^{-1})$; $2\CEG\Lam+2$\\
$S$, $A_s$, $k_A$ & number $4D$ of stages; predecessor set of stage $s$; deterministic indegree cap\\
$L_s$, $p_s$, $\Pi$ & logger and design of stage $s$; final lottery $\tfrac12\delta_a+\tfrac1{2S}\sum_sp_s$\\
$\rho_s(h)$, $\Bad$, $\Vis$ & truncated rank at stage $s$; unsuccessful and visible stages of $(h,T)$\\
$t$, $v$, $\beta$ & precision of a stage; supplied bound on $R^*$; failure allowance\\
$b_F$, $C_*$, $Y_h$ & $b/2$; $64C(b)$; inverse-propensity score of $h$ in the final estimate\\
\bottomrule
\end{tabular}
\end{table}

\section{A legal gain and one acquisition stage}\label{sec:stage}

Fix an anchor $a$, a logger $L$, thresholds $\zeta,b\in(0,1/2]$, and a lottery $\pi$, all determined before a fresh reset. The lottery $\pi$ is numerical: it enters only through the potential $\Phi_\pi$ and is never used to draw a pilot. The pilot of the reset is drawn from $L$.

\subsection{The displayed gain}

\begin{lemma}[Displayed gain]\label{lem:gain}
Draw $g\sim L$ before the record $T$, and for $h\in\cH$ let
\begin{equation}\label{eq:gain}
 \widehat G(h)=\sum_{j=1}^{n_g}
 \1\bigl\{Q_L(T,j)\ge\zeta,\ \ h(z_i)=y_i\ \text{for all } i\le j\bigr\}\,
 \frac{\Phi_\pi(T,j+1)-\Phi_\pi(T,j)}{Q_L(T,j)} ,
\end{equation}
where each summand is defined to be zero unless its indicator equals one, and the division is carried out only in that case. Then:
\begin{enumerate}[label=(\alph*)]
\item $\widehat G(h)$ is a function of the display of $g$, of the fixed objects $a,L,\pi,\zeta,b$, and of the values of $h$ at the displayed queries;
\item $0\le\widehat G(h)\le 2/(\zeta b)$;
\item $\E[\widehat G(h)\mid T]=\Phi_\pi(T,\rho_L(h))$;
\item for every pair $h,h'\in\cH$ there is an event $O_{hh'}$, decided by the display of $g$ and the values of $h$ and $h'$ at the displayed queries, such that
\begin{equation}\label{eq:observable-disagreement}
 \Pr(O_{hh'}\mid T)\ \ge\ \zeta\,\1\{\rho_L(h)\ne\rho_L(h')\}.
\end{equation}
\end{enumerate}
\end{lemma}

\begin{proof}
(a) Let $j\le n_g$. The pairs at positions $1,\ldots,j$ are displayed. By \cref{lem:propensity}(ii), $Q_L(T,j)$ depends on the pairs before $j$, while $Q_\pi(T,j)$ and $Q_\pi(T,j+1)$ depend on the pairs up to $j$. The indicators $\1\{j>\tau_a\}$ and $\1\{j+1>\tau_a\}$ in the potentials depend only on whether $a$ agrees with the pairs up to $j$, and the test on $h$ uses the pairs up to $j$. Hence every summand is computable. The position $j+1$ is virtual: only the predictions at positions up to $j$ enter $\Phi_\pi(T,j+1)$, so no query at position $j+1$ and no end-of-record information is needed. In particular, if $g$ errs at the last actual query, then $j=n_g=M$ is processed in exactly the same way, and the learner never needs to know that the record has ended. Summands with $j>n_g$ are not included, and the descriptor is never used.

(c) Fix $T$. For $1\le j\le M$, the $j$th summand is present exactly when $j\le n_g$, that is, when $\tau_g\ge j$, which by \cref{lem:propensity}(iv) has conditional probability $Q_L(T,j)$. Its two tests say $j\le J_L$ (by monotonicity of $Q_L$) and $\tau_h\ge j+1$, that is, together, $j+1\le\min(J_L+1,\tau_h)=\rho_L(h)$. On this branch $Q_L(T,j)\ge\zeta>0$. Since $\rho_L(h)\le M+1$, every $j$ with $j+1\le\rho_L(h)$ satisfies $j\le M$. Therefore
\[
 \E[\widehat G(h)\mid T]
 =\sum_{l=2}^{\rho_L(h)}\bigl(\Phi_\pi(T,l)-\Phi_\pi(T,l-1)\bigr)
 =\Phi_\pi(T,\rho_L(h))-\Phi_\pi(T,1)=\Phi_\pi(T,\rho_L(h)).
\]

(b) The increments $\Phi_\pi(T,j+1)-\Phi_\pi(T,j)$ are nonnegative: both potentials vanish when $j+1\le\tau_a$, the increment at $j=\tau_a$ equals $\Phi_\pi(T,\tau_a+1)\ge0$, and for $j>\tau_a$ the potential is nondecreasing. Every present summand has $j+1\le\rho_L(h)$. The summand with $j=\tau_a$, if present, has divisor $Q_L(T,\tau_a)\ge1/2$ and is at most $2\Phi_\pi(T,\tau_a+1)\le2/b$. The summands with $j>\tau_a$ have divisors at least $\zeta$, and their increments add up to at most $\Phi_\pi(T,\rho_L(h))\le1/b$, so they contribute at most $1/(\zeta b)$. Summands with $j<\tau_a$ vanish. Since $\zeta\le1/2$, the total is at most $2/b+1/(\zeta b)\le2/(\zeta b)$.

(d) Let $O_{hh'}$ be the event that some displayed position $j\le n_g$ has $Q_L(T,j)\ge\zeta$, both $h$ and $h'$ agree with the pairs at positions $1,\ldots,j-1$, and exactly one of them agrees with the pair at position $j$. It is decided by the display and by the values of $h,h'$ at displayed queries. Suppose $\rho_L(h)<\rho_L(h')$ (the other case is symmetric). Then $\rho_L(h)\le J_L$, so $\rho_L(h)=\tau_h\le J_L$, and $\tau_h<\rho_L(h')\le M+1$, so $\tau_h\le M$. Put $j^*=\tau_h$. Then $h$ is correct before $j^*$ and errs at $j^*$, while $\tau_{h'}\ge\rho_L(h')>j^*$, so $h'$ is correct at $1,\ldots,j^*$. Since $j^*\le J_L$, we have $Q_L(T,j^*)\ge\zeta$, and with conditional probability $\Pr(\tau_g\ge j^*\mid T)=Q_L(T,j^*)\ge\zeta$ the pair at $j^*$ is displayed; on this event $O_{hh'}$ occurs with $j=j^*$. This includes the case in which $g$ itself errs at $j^*$, because discrepancies are displayed.
\end{proof}

The legal content of \eqref{eq:gain} is the one-position shift: a pair displayed at position $j$ is used to estimate the increment of the potential at the virtual position $j+1$. The shift is possible because the pilot's discrepancy pair is displayed, and it requires dividing by the propensity $Q_L(T,j)$ of position $j$, not of position $j+1$.

\subsection{One acquisition stage}

The numerical optimization in a stage is the classical exponential-weights update \citep[Eq.~(3.3)]{helmbold1998} with the multiplicative-potential analysis of \citet{freund1997}. Once the legal bounded gain of \cref{lem:gain} and a finite observed dictionary are available, the reset and support orders of one stage follow from a standard multiplicative-potential bound, bounded martingale concentration and one randomized sparsification. We prove this conversion for the displayed gain in \cref{app:acquisition}; the optimizer itself is not a new primitive. The logger stays fixed while the numerical weights change, so the gains form an adapted sequence and their concentration needs martingale arguments rather than independence.

\begin{algorithm}[t]
\caption{$\textsc{Acquire}(L,a,w,\zeta,b,\alpha)$: one acquisition stage}\label{alg:acquire}
\begin{algorithmic}[1]
\Require logger $L$ with $L(a)\ge1/2$; numbers $w\in(0,1]$, $\zeta,b\in(0,1/2]$, $\alpha\in(0,1)$
\State $W_0\gets 1/(\zeta wb)$, \ $W\gets 2/(\zeta bw\Lam)$, \ $\CEG\gets(2+(e-1)e)/(1-e^{-1})$
\State $m\gets$ least integer $m\ge8W_0$ with $2(2Dm+1)^{2d}\,2^{-m/(2W_0)}\le\alpha/4$
\For{$i=1,\ldots,m$} draw $g\sim L$, play it on a fresh reset, store the display \EndFor
\State call the oracle on all queries displayed in these $m$ resets; let $\cG$ be the result; $K_0\gets(Dm+1)^d$
\State $\iota\gets\log(4K_0/\alpha)$, \ $n\gets\lceil W\iota\rceil$, \ $\pi_1\gets$ uniform on $\cG$
\For{$t=1,\ldots,n$}
 \State draw $g_t\sim L$ and play it on a fresh reset
 \State for every $h\in\cG$: $\gamma_{t,h}\gets\widehat G_t(h)/(w\Lam)$, with $\widehat G_t$ given by \eqref{eq:gain} for $\pi=\pi_t$
 \State $\pi_{t+1}(h)\gets\pi_t(h)e^{\gamma_{t,h}/W}\big/\sum_{u\in\cG}\pi_t(u)e^{\gamma_{t,u}/W}$ for $h\in\cG$
\EndFor
\State $\bar\pi\gets\frac1n\sum_{t=1}^n\pi_t$ \Comment{average of the pre-update weights}
\State $k\gets\bigl\lceil\frac8b\log\frac{4(D+1)}{\alpha b}\bigr\rceil$; draw $k$ slots independently from $\bar\pi$ with private coins
\State \Return the uniform lottery $p$ on the $k$ slots (with multiplicity)
\end{algorithmic}
\end{algorithm}

\begin{lemma}[One acquisition stage]\label{lem:acquisition}
Condition on a history that fixes $a,L,w,\zeta,b,\alpha$ with $L(a)\ge1/2$, $0<w\le1$, $0<\zeta,b\le1/2$ and $0<\alpha<1$, and let $p$ be the output of \cref{alg:acquire}. Put $C(b)=2\CEG\Lam+2$.
\begin{enumerate}[label=(\alph*)]
\item On every history the procedure uses exactly $m+n$ resets, makes one oracle call and returns a lottery with at most $k$ slots. The pilot of every reset is drawn from $L$. Enumerating $\cG$, updating the numerical weights and drawing the slots play nothing by themselves; a member of $\cG$ is played only if it is drawn from $L$.
\item If $F(a)\le w$, then with conditional probability at least $1-\alpha$, simultaneously for every $h\in\cH$,
\begin{equation}\label{eq:truncated-certificate}
 \E_T\,\Phi_p\bigl(T,\rho_L(h)\bigr)\le C(b)\,w,
\end{equation}
where $\E_T$ integrates a fresh reference record with $a,L,p$ held fixed.
\item The numbers $m,n,k$ are deterministic functions of $(d,D,w,\zeta,b,\alpha)$, and
\[
 m+n=\Ot\!\left(\frac{d+1+\log(1/\alpha)}{\zeta wb}\right),\qquad k=O\!\left(\frac1b\log\frac{D}{\alpha b}\right),
\]
where the hidden factors are polylogarithmic in $D$, $d+1$, $1/(\zeta wb)$ and $1/\alpha$.
\end{enumerate}
\end{lemma}

The proof is in \cref{app:acquisition}. Two transfers are worth pointing out. First, the dictionary sample controls, uniformly over all pairs, the observable disagreement events of \cref{lem:gain}(d) before any representative of a particular comparator is chosen. Together with \eqref{eq:observable-disagreement} this gives $\Pr(\rho_L(h)\ne\rho_L(h'))\le wb$ for every $h$ and its representative $h'$, at the price of exactly the visibility factor $\zeta$. Second, after the numerical pass, one random sparse bad set of records controls all ranks of a record on which the anchor fails, so the dense and sparse potentials can be compared there; on records where the anchor survives both potentials vanish and no comparison is needed. The transfer from the dictionary to the whole class is paid once, after the sparsification, and costs at most $w$. The returned slots are available functions that have not necessarily been played. A later logger may play them, and each such use is counted in $V$.

\section{From truncated visibility to a joint learner}\label{sec:composition}

\Cref{lem:acquisition} gives a certificate only at ranks truncated at the logger's frontier. This section moves the frontier forward through a sequence of stages, recovers the weight lost by mixing the stages, and turns the resulting lottery into an output and an interval.

\subsection{The stage graph}

The stages are arranged along a sparse directed acyclic graph whose long paths survive the deletion of any set of vertices. Such graphs are classical. The multiscale connector argument is due to Erd\H{o}s, Graham and Szemer\'edi \citep{EGS1975_egs1975_journal_static_renyi}; the local-expander formulation that we follow is that of Alwen, Blocki and Pietrzak \citep[Section~4, Lemmas~4.2--4.4]{ABP2017_2018_abp2017_arxiv_v2}. \Cref{app:graph} gives a self-contained proof with the constants we need.

\begin{lemma}[Robust forward graph; \cref{app:graph}]\label{lem:graph-interface}
Let $S\ge2$ and $0<\beta<1$, and put
\begin{equation}\label{eq:degree-cap}
 k_G=\Bigl\lceil40\Bigl(2\log(10e)+\log\frac{2S^2}{\beta}\Bigr)\Bigr\rceil,
 \qquad
 k_A=\min\bigl\{S-1,\ 19+k_G\bigl(\lfloor\log_2(2S)\rfloor+1\bigr)\bigr\}.
\end{equation}
Using private coins only, one can draw a directed graph $\Gamma$ on $[S]=\{1,\ldots,S\}$, with every edge $(x,y)$ satisfying $x<y$, such that on every draw each $s\ge2$ has a predecessor set $A_s\subseteq[s-1]$ with $s-1\in A_s$ and $|A_s|\le k_A$. With probability at least $1-\beta$ the graph has the \emph{path property}: for every set $B\subseteq[S]$ there is a directed path in $\Gamma$ that avoids $B$ and has at least $S-4|B|$ vertices (a vacuous statement when $S-4|B|\le0$).
\end{lemma}

The set $B$ in the path property is arbitrary; it may depend on the graph and on all data collected later. The cap $k_A$ is a deterministic number, and every parameter below uses $k_A$ rather than the realized indegrees. Note that $k_A=O((1+\log S)\log(S/\beta))$.

More generally, we call a \emph{stage graph} any directed graph $\Gamma$ on $[S]$ with edges $(x,y)$, $x<y$, together with a number $k_A$ fixed before $\Gamma$ is drawn, such that on every draw $s-1\in A_s$ and $|A_s|\le k_A$ for $s\ge2$. Besides the graph of \cref{lem:graph-interface}, the complete graph with all edges $(x,y)$, $x<y$, and $k_A=S-1$ is a stage graph; it has the path property on every draw, since $[S]\setminus B$ listed in increasing order is a path with $S-|B|$ vertices.

\subsection{One precision stage}

\begin{algorithm}[t]
\caption{$\textsc{PrecisionStage}(a,t,w,v,\beta,\Gamma)$}\label{alg:precision}
\begin{algorithmic}[1]
\Require anchor $a\in\cH$; numbers $0<t\le w\le1$, $v\in[0,1]$, $\beta\in(0,1)$; a stage graph $\Gamma$ on $[S]$, $S=4D$, with predecessor sets $A_s$ and indegree cap $k_A$
\State $b\gets$ the largest $b\in(0,1/2]$ with $b\,C(b)\le t/(1024w)$, where $C(b)=2\CEG\Lam+2$; \ $\zeta\gets b/(2k_A)$
\For{$s=1,\ldots,S$}
 \State $L_s\gets\delta_a$ if $s=1$, and $L_s\gets\frac12\delta_a+\frac1{2|A_s|}\sum_{r\in A_s}p_r$ if $s\ge2$
 \State $p_s\gets\textsc{Acquire}(L_s,a,w,\zeta,b,\beta/S)$
\EndFor
\State $\Pi\gets\frac12\delta_a+\frac1{2S}\sum_{s=1}^Sp_s$, \ $b_F\gets b/2$, \ $C_*\gets64C(b)$
\State $n_F\gets$ the sample size \eqref{eq:est-n} for $W_Y=2/b$, $c_1=2$, $c_2=2C_*w$, accuracy $t$, ceiling $v$ and failure $\beta$
\State play $n_F$ pilots drawn independently from $\Pi$ on fresh resets and store their displays
\State call the oracle on all queries displayed in these resets; for each returned member $h$ compute $\widehat R(h)=\frac1{n_F}\sum_{i=1}^{n_F}Y_{h,i}$ with $Y_h$ from \eqref{eq:final-score}; let $\widehat h$ be a returned member minimizing $\widehat R$ (fixed tie-break)
\State $q\gets\widehat R(\widehat h)$, \ $l\gets(q-t/4)/(1+\theta)$, \ $u\gets(q+t/4)/(1-\theta)$ \Comment{$\theta$ from \eqref{eq:est-s}}
\State clip $l$ and $u$ separately to $[0,1]$; if the result is unordered or wider than $t$, replace it by $[0,0]$
\State \Return $\widehat h$ and the interval
\end{algorithmic}
\end{algorithm}

\Cref{alg:precision} describes one precision stage. It receives a proper anchor $a$, a precision $t$, a numerical bound $w$ intended to satisfy $F(a)\le w$, a numerical bound $v$ intended to satisfy $R^*\le v$, a failure allowance $\beta$, and a stage graph. The procedure and all its sample sizes are defined whether or not $F(a)\le w$ and $R^*\le v$ hold; these inequalities are needed only for the coverage guarantee.

Every logger $L_s$ consists of the anchor and of the sparse designs $p_r$ of earlier stages, so it uses only functions that are available when it is formed. The dense dictionaries of the designs are never loggers. Averaging earlier exploration distributions is an established device; for example, PC-PG averages the visitation distributions of earlier policies \citep[Section~3, Eq.~(3)]{PCPG2020_pcpg2020_arxiv_v1}. What has to be shown here is that the average $\Pi$ has a good certificate at the true stopping ranks, using only legal acquisition.

\begin{lemma}[Choice of $b$]\label{lem:b-choice}
Let $0<t\le w\le1$. The function $b\mapsto bC(b)$ is continuous and strictly increasing on $(0,1/2]$, tends to $0$ as $b\to0$, and exceeds $t/(1024w)$ at $b=1/2$. Hence the $b$ of \cref{alg:precision} is the unique root of $bC(b)=t/(1024w)$ in $(0,1/2)$. It satisfies
\begin{equation}\label{eq:b-lower}
 \frac{t}{65536\,w\log(1024w/t)}\ \le\ b\ \le\ \frac{t}{2048\,w}\le\frac1{2048},
\end{equation}
so that $1/b=O\bigl((w/t)\log(ew/t)\bigr)$, $\Lam=O(\log(ew/t))$, $C(b)=O(\log(ew/t))$, and $\Lam\ge\log2049>2$.
\end{lemma}

\begin{proof}
The derivative of $bC(b)=b(2\CEG\log(1+1/b)+2)$ is $2\CEG[\log(1+1/b)-1/(1+b)]+2$, which is at least $2$ because $\log(1+x)\ge x/(1+x)$ with $x=1/b$. Since $b\log(1+1/b)\to0$, $bC(b)\to0$ as $b\to0$. At $b=1/2$ we have $bC(b)=\CEG\log3+1>1>t/(1024w)$. This proves the first claims. As $C(b)\ge2$, the root satisfies $2b\le t/(1024w)$, which is the upper bound in \eqref{eq:b-lower}; then $\Lam=\log(1+1/b)\ge\log2049$.

For the lower bound put $x=1024w/t\ge1024$ and $b'=1/(64x\log x)\le1/2$. Since $65\log x\le x^2$ for $x\ge1024$, we have $1+1/b'=1+64x\log x\le x^3$, hence $\log(1+1/b')\le3\log x$ and
\[
 b'C(b')\le\frac{6\CEG\log x+2}{64x\log x}=\frac{6\CEG+2/\log x}{64x}\le\frac1x ,
\]
because $\CEG=10.5530\ldots$ and $\log x\ge\log1024$ give $6\CEG+2/\log x<63.61<64$. Thus $b'C(b')\le t/(1024w)$, and monotonicity gives $b\ge b'=t/(65536w\log(1024w/t))$. The order statements follow from $\log(1024w/t)\le\log1024+\log(w/t)$ and $\Lam\le\log(1+1/b')\le3\log x$.
\end{proof}

\subsection{Recordwise propagation with correlated failures}

Freeze the graph and the designs $p_1,\ldots,p_S$. Fix $h\in\cH$ and a reference record $T$ with $\tau_h>\tau_a$. For each stage $s$ let
\[
 \rho_s(h)=\min\bigl(\tau_h,J_{L_s}+1\bigr),\qquad
 \Delta=\tau_h-\tau_a,\qquad
 x_s=\rho_s(h)-\tau_a ,
\]
where $J_{L_s}$ is the frontier of $L_s$ with threshold $\zeta$. Call stage $s$ \emph{successful} for $(h,T)$ if $Q_{p_s}(T,\rho_s(h))\ge b$, and let $\Bad(h,T)\subseteq[S]$ be the set of unsuccessful stages. Call $s$ \emph{visible} for $(h,T)$ if $\rho_s(h)=\tau_h$, and let $\Vis(h,T)$ be the set of visible stages.

\begin{lemma}[Propagation and visibility]\label{lem:visibility}
In the situation above:
\begin{enumerate}[label=(\roman*)]
\item $1\le x_s\le\Delta\le D$ for every $s$;
\item if $r\in A_s\setminus\Bad(h,T)$, then $x_s\ge\min(\Delta,x_r+1)$;
\item if $\Gamma$ has the path property, then $|\Vis(h,T)|\ge S-4|\Bad(h,T)|-\Delta+1$ whenever the right side is positive.
\end{enumerate}
If moreover the certificate \eqref{eq:truncated-certificate} holds for every stage $s$ (with logger $L_s$), then for every $h\in\cH$ and every $s$,
\begin{equation}\label{eq:bad-stage-mass}
 \Pr_T\bigl(\tau_h>\tau_a,\ s\in\Bad(h,T)\bigr)\le2b\,C(b)\,w,
\end{equation}
and if in addition $\Gamma$ has the path property, then
\begin{equation}\label{eq:invisible-mass}
 \Pr_T\bigl(\tau_h>\tau_a,\ |\Vis(h,T)|<S/4\bigr)\le16\,b\,C(b)\,w .
\end{equation}
\end{lemma}

\begin{proof}
(i) Since $J_{L_s}\ge\tau_a$ and $\tau_h>\tau_a$, we have $\rho_s(h)\ge\tau_a+1$, so $x_s\ge1$; and $\rho_s(h)\le\tau_h$ gives $x_s\le\Delta$. Finally $\Delta\le(M+1)-1\le D$ because $\tau_a\ge1$.

(ii) Let $r\in A_s\setminus\Bad(h,T)$. Since $L_s$ gives weight $1/(2|A_s|)$ to $p_r$ and propensities are linear in the lottery,
\[
 Q_{L_s}(T,\rho_r(h))\ \ge\ \frac{Q_{p_r}(T,\rho_r(h))}{2|A_s|}\ \ge\ \frac b{2k_A}=\zeta .
\]
As $\rho_r(h)\le M+1$, the definition of the frontier gives $J_{L_s}\ge\rho_r(h)$, hence $\rho_s(h)\ge\min(\tau_h,\rho_r(h)+1)$, which is the claim.

(iii) Let $s_1<\cdots<s_\ell$ be the vertices of a path avoiding $\Bad(h,T)$ with $\ell\ge S-4|\Bad(h,T)|$, so $s_i\in A_{s_{i+1}}\setminus\Bad(h,T)$. By (i) and (ii), $x_{s_i}\ge\min(\Delta,i)$ by induction on $i$. Hence $s_\Delta,\ldots,s_\ell$ are visible, which gives $\ell-\Delta+1$ visible stages when this number is positive. If the claimed bound is nonpositive there is nothing to prove.

For \eqref{eq:bad-stage-mass}, note that on $\{\tau_h>\tau_a\}$ we have $\rho_s(h)>\tau_a$, so $\Phi_{p_s}(T,\rho_s(h))=1/(Q_{p_s}(T,\rho_s(h))+b)$. For $q\ge0$, $\1\{q<b\}\le2b/(q+b)$. Hence
\[
 \Pr_T\bigl(\tau_h>\tau_a,\ s\in\Bad(h,T)\bigr)\le2b\,\E_T\Phi_{p_s}(T,\rho_s(h))\le2bC(b)w
\]
by \eqref{eq:truncated-certificate}. For \eqref{eq:invisible-mass}, suppose $\tau_h>\tau_a$ and $|\Vis(h,T)|<S/4=D$. If $S-4|\Bad|-\Delta+1>0$, then (iii) gives $D>4D-4|\Bad|-\Delta+1\ge3D-4|\Bad|+1$, so $|\Bad|>D/2$. Otherwise $4|\Bad|\ge4D-\Delta+1\ge3D+1$, and again $|\Bad|>D/2$. Markov's inequality and linearity of expectation give
\begin{align*}
 \Pr_T\bigl(\tau_h>\tau_a,\ |\Bad(h,T)|>D/2\bigr)
 &\le\frac2D\sum_{s=1}^S\Pr_T\bigl(\tau_h>\tau_a,\ s\in\Bad(h,T)\bigr)\\
 &\le\frac2D\cdot4D\cdot2bC(b)w=16bC(b)w .
\end{align*}
No independence between the failures of different stages is used: $\Bad(h,T)$ may be any set, and it may depend on the graph, the designs, $h$ and $T$.
\end{proof}

\subsection{Harmonic recovery of the mixture weight}

\begin{lemma}[Final certificate]\label{lem:final-certificate}
Suppose $\Gamma$ has the path property and the certificate \eqref{eq:truncated-certificate} holds for every stage. Then, simultaneously for every $h\in\cH$,
\begin{equation}\label{eq:final-certificate}
 \E_T\,\frac{\1\{\tau_h>\tau_a\}}{Q_\Pi(T,\tau_h)+b_F}\ \le\ C_*\,w,
 \qquad C_*=64\,C(b),\ b_F=b/2 .
\end{equation}
\end{lemma}

\begin{proof}
Fix $h$ and a record with $\tau_h>\tau_a$ and $|\Vis(h,T)|\ge S/4$. Then $Q_{\delta_a}(T,\tau_h)=0$, so with $q_s=Q_{p_s}(T,\tau_h)$,
\[
 Q_\Pi(T,\tau_h)+\frac b2=\frac1{2S}\Bigl(\sum_{s=1}^Sq_s+Sb\Bigr)\ \ge\ \frac1{2S}\sum_{s\in\Vis(h,T)}(q_s+b).
\]
For positive numbers $y_s$, the Cauchy--Schwarz inequality gives $\bigl(\sum_{s\in\Vis}y_s\bigr)^{-1}\le|\Vis|^{-2}\sum_{s\in\Vis}y_s^{-1}$. With $y_s=q_s+b$ and $|\Vis|\ge S/4$,
\begin{equation}\label{eq:harmonic}
 \frac1{Q_\Pi(T,\tau_h)+b/2}\ \le\ \frac{2S\cdot16}{S^2}\sum_{s\in\Vis(h,T)}\frac1{q_s+b}
 =\frac{32}S\sum_{s\in\Vis(h,T)}\Phi_{p_s}(T,\rho_s(h)),
\end{equation}
because $\rho_s(h)=\tau_h>\tau_a$ for visible $s$. Integrate \eqref{eq:harmonic} over the records with $\tau_h>\tau_a$ and $|\Vis|\ge S/4$, enlarge the nonnegative sum to all $s\in[S]$, and apply \eqref{eq:truncated-certificate} to each stage; this part contributes at most $(32/S)\cdot S\cdot C(b)w=32C(b)w$. On the records with $\tau_h>\tau_a$ and $|\Vis|<S/4$, bound the integrand by $1/b_F=2/b$ and use \eqref{eq:invisible-mass}; this contributes at most $(2/b)\cdot16bC(b)w=32C(b)w$. The sum is $C_*w$.
\end{proof}

The factor $1/(2S)$ with which each design enters $\Pi$ is recovered because a constant fraction of the stages is visible; bounding the contribution of a single stage would lose a further factor of order $D$. The certificate holds simultaneously for all $h$, but its exceptional set of records with low propensity depends on $h$. We never take a union of these sets over the class.

\subsection{The final estimate and the observable interval}

For the final estimate, pilots are drawn from the frozen lottery $\Pi$ before fresh records. For $h\in\cH$ write $Q_h=Q_\Pi(T,\tau_h)$ and define the score
\begin{equation}\label{eq:final-score}
 Y_h=\frac{\lambda(T,\tau_h)}{Q_h}\ \ \text{if}\ \tau_g\ge\tau_h\ \text{and}\ Q_h\ge b_F,
 \qquad Y_h=0\ \ \text{otherwise},
\end{equation}
where the test is carried out first and the division only on the first branch. This is inverse-propensity estimation under a frozen lottery; inverse-action-probability exploration is classical, see for instance \citet[Section~2.2, Lemma~1 and Algorithm~1]{dudik2011_v1}.

\begin{lemma}[The final score]\label{lem:final-score}
\begin{enumerate}[label=(\alph*)]
\item $Y_h$ is a function of the display of $g$ and of the values of $h$ at the displayed queries, and it depends on $h$ only through $\tau_h(T)$.
\item Conditionally on $T$, $\E[Y_h\mid T]=\ell_h(T)\,\1\{Q_h\ge b_F\}$, $0\le Y_h\le2/b$ and
\[
 \E[Y_h^2\mid T]=\begin{cases}\ell_h(T)^2/Q_h&\text{if }Q_h\ge b_F,\\ 0&\text{if }Q_h<b_F.\end{cases}
\]
\item If \eqref{eq:final-certificate} holds and $b$ is chosen as in \cref{alg:precision}, then for every $h\in\cH$,
\begin{equation}\label{eq:final-moments}
 0\le R(h)-\E Y_h\le2b_FC_*w=64\,b\,C(b)\,w\le\frac t{16},
 \qquad
 \E Y_h^2\le2R(h)+2C_*w .
\end{equation}
\end{enumerate}
\end{lemma}

\begin{proof}
(a) If $h$ disagrees with some displayed pair, its first disagreement is $\tau_h\le n_g\le\tau_g$; then $Q_h$ depends on the pairs before $\tau_h$, $\lambda(T,\tau_h)$ on the pairs up to $\tau_h$, and all are displayed. If $g$ survives, the whole record and its descriptor are displayed, so $\tau_h$, $Q_h$ and $\lambda(T,\tau_h)$ are known, including on an empty record, where $\tau_h=1$ and $Q_h=1$. If $g$ errs and $h$ agrees with every displayed pair, then $\tau_h>\tau_g$, the test fails and $Y_h=0$; this case is recognized from the display (no end tag, no disagreement), and the later loss of $h$ is never evaluated. In all cases $Y_h$ depends on $h$ only through $\tau_h$.

(b) Given $T$, the rank $\tau_h$ and the number $Q_h$ are fixed. If $Q_h<b_F$, then $Y_h=0$. If $Q_h\ge b_F$, then $Y_h=\lambda(T,\tau_h)/Q_h$ on the event $\tau_g\ge\tau_h$, which has conditional probability $Q_h$ by \cref{lem:propensity}(iv), and $Y_h=0$ otherwise. This gives both moments, and on the nonzero branch $Y_h\le1/b_F=2/b$.

(c) By (b), $R(h)-\E Y_h=\E[\ell_h\1\{Q_h<b_F\}]\in[0,\Pr(Q_h<b_F)]$. If $\tau_h\le\tau_a$, the anchor survives up to $\tau_h$, so $Q_h\ge\frac12\ge b_F$. If $\tau_h>\tau_a$, use $\1\{Q_h<b_F\}\le2b_F/(Q_h+b_F)$ and \eqref{eq:final-certificate}. Thus the bias is at most $2b_FC_*w=64bC(b)w\le64\,t/1024=t/16$ by the choice of $b$. For the second moment, on $\{\tau_h\le\tau_a\}$ we have $\ell_h^2/Q_h\le2\ell_h$. On $\{\tau_h>\tau_a,\ Q_h\ge b_F\}$ we have $Q_h+b_F\le2Q_h$, hence $\ell_h^2/Q_h\le2/(Q_h+b_F)$, and \eqref{eq:final-certificate} bounds the integral of this part by $2C_*w$.
\end{proof}

None of these bounds assumes $R(h)\le v$, neither for all candidates nor for the selected output.

\paragraph{Estimation parameters.}
Let $W_Y=2/b$, $c_1=2$, $c_2=2C_*w$, and
\begin{equation}\label{eq:est-s}
 s_E=\min\Bigl\{\frac1{W_Y},\ \frac t{16c_1(v+t)},\ \frac t{16\max(c_2,t)}\Bigr\},
 \qquad\theta=s_Ec_1 .
\end{equation}
For an integer $n\ge1$ define the law-free numbers
\begin{equation}\label{eq:net-parameters}
\begin{aligned}
 \nu_n&=\min\bigl\{v'\in\mathbb N:\ v'\ge8n,\ 2(2v'D+1)^{2d}e^{-v'/(4n)}<1\bigr\},
 \qquad H_n=(\nu_nD+1)^d,\\
 \iota_n&=\log\frac{4H_n}{\beta},
 \qquad
 \kappa_n=\max\Bigl\{4,\ 18\log\frac{4(2nD+1)^{2d}}{\beta}\Bigr\},
\end{aligned}
\end{equation}
and let $n_F\ge2$ be the least integer $n$ with
\begin{equation}\label{eq:est-n}
 \frac{16\,\iota_n}{s_En}\le t
 \qquad\text{and}\qquad
 \frac{16\,(W_Y\kappa_n+2)}{n}\le t .
\end{equation}
The quantities $\nu_n$ and $H_n$ are the size parameters of a latent-record net (\cref{lem:vc-net} with mass $1/n$), and $\kappa_n$ is a rare-count threshold (\cref{lem:vc-rare}). The net is used only in the analysis; the learner never constructs it.

\begin{lemma}[Uniform relative estimate and interval; \cref{app:estimation}]\label{lem:estimate-interval}
Condition on the history before the final estimate, and suppose that \eqref{eq:final-certificate} holds. Then with conditional probability at least $1-\beta$,
\begin{equation}\label{eq:relative-estimate}
 |\widehat R(h)-R(h)|\le\theta R(h)+\frac t4\qquad\text{for every }h\in\cH .
\end{equation}
On this event the raw endpoints of \cref{alg:precision} satisfy $l\le R^*\le R(\widehat h)\le u$. If moreover $R^*\le v$, then $u-l\le2t/3$, so the clipped interval is ordered, has width at most $t$, and contains $R^*$ and $R(\widehat h)$. On every history the returned interval is ordered, lies in $[0,1]$ and has width at most $t$, and $\widehat h\in\cH$.
\end{lemma}

The selection step enumerates the patterns of $\cH$ on the queries displayed in the final $n_F$ resets. Since $\widehat R(h)$ depends on $h$ only through its values on these queries, a minimizer over the returned representatives is a minimizer over all of $\cH$; no population minimizer is assumed. This enumeration is separate from the design dictionaries.

\subsection{The precision stage and the proof of \texorpdfstring{\cref{thm:main}}{Theorem 3.1}}

\begin{proposition}[One precision stage]\label{prop:precision}
Let $\Gamma$ be a stage graph with cap $k_A$, for instance the graph of \cref{lem:graph-interface} drawn with any allowance. Run \cref{alg:precision} with $0<t\le w\le1$, $v\in[0,1]$ and $\beta\in(0,1)$.
\begin{enumerate}[label=(\alph*)]
\item On every history the output is a member of $\cH$ and an ordered interval in $[0,1]$ of width at most $t$. The number of resets is
\begin{equation}\label{eq:stage-cost}
 \sum_{s=1}^S(m_s+n_s)+n_F=\Ot\!\left(\bigl(d+1+\log(D/\beta)\bigr)\,\frac{Dk_Aw+v+w+t}{t^2}\right),
\end{equation}
where the hidden factors are polylogarithmic in $d+1$, $D$, $k_A$, $1/t$ and $1/\beta$. Every function played is the anchor or a slot of one of the $S$ designs, so the stage plays at most
\[
 1+S\Bigl\lceil\frac8b\log\frac{4(D+1)S}{\beta b}\Bigr\rceil=\Ot\!\left(\frac{Dw}t\right)
\]
distinct functions. The stage makes $S$ design oracle calls and one selection call.
\item Suppose $F(a)\le w$ and $R^*\le v$. Conditionally on the history before the stage and on any realization of $\Gamma$ with the path property, the interval covers $R^*$ and $R(\widehat h)$ with probability at least $1-2\beta$. Consequently, if $\Gamma$ is drawn with allowance $\beta$ just before the stage, the interval covers with conditional probability at least $1-3\beta$.
\end{enumerate}
All sample sizes are numerical functions of $(d,D,t,w,v,\beta,k_A)$.
\end{proposition}

\begin{proof}
(b) Each call of \cref{alg:acquire} uses a logger with $L_s(a)\ge1/2$ and fresh resets. Since $F(a)\le w$, \cref{lem:acquisition}(b) gives, conditionally on every history before that call, failure probability at most $\beta/S$. A union bound over the $S$ calls shows that all certificates hold with probability at least $1-\beta$. On this event and on the path property, \cref{lem:final-certificate} gives \eqref{eq:final-certificate}. Conditionally on the designs, \cref{lem:estimate-interval} then gives coverage with probability at least $1-\beta$. The two failure probabilities add up to $2\beta$; drawing the graph adds $\beta$.

(a) The width and properness claims are part of \cref{lem:estimate-interval}. By \cref{lem:b-choice}, $1/b=O((w/t)\log(ew/t))$. For each design $W_0=1/(\zeta wb)=2k_A/(wb^2)$ and $W=2W_0/\Lam\le W_0$, since $\Lam>2$. Hence \cref{lem:acquisition}(c), with failure $\beta/S$, gives
\[
 m_s+n_s=\Ot\!\left((d+1+\log(S/\beta))\,\frac{k_A}{wb^2}\right)
 =\Ot\!\left((d+1+\log(D/\beta))\,\frac{k_Aw}{t^2}\right),
\]
and summing over $S=4D$ designs gives the first term of \eqref{eq:stage-cost}. By \cref{lem:est-order} in \cref{app:estimation},
\[
 n_F=\Ot\!\left((d+1+\log(1/\beta))\Bigl[\frac{W_Y}t+\frac{c_1(v+t)}{t^2}+\frac{c_2+t}{t^2}\Bigr]\right),
\]
with $W_Y/t=2/(bt)=\Ot(w/t^2)$ and $c_2=128C(b)w=\Ot(w)$; this is the remaining part of \eqref{eq:stage-cost}. By \cref{lem:acquisition}(a), every pilot of the stage is drawn from a logger $L_s$ or from $\Pi$, and these lotteries are built from the anchor and the slots of $p_1,\ldots,p_S$; so every function played in the stage is the anchor or a slot. With failure $\beta/S$ each design has $\lceil(8/b)\log(4(D+1)S/(\beta b))\rceil$ slots, which gives the bound on played functions; by \cref{lem:b-choice} it is $\Ot(Dw/t)$. Each design makes one oracle call and the selection makes one more.
\end{proof}

\begin{algorithm}[t]
\caption{Learner of \cref{thm:main} (supplied ceilings $f,r$)}\label{alg:main}
\begin{algorithmic}[1]
\Require $d,D,\eps,\delta,f,r$
\If{$\eps=1$} \Return an empty-list representative and $[0,1]$ \EndIf
\State run the procedure of \cref{lem:anchor-supplied} with failure $\alpha=\delta/4$; obtain $a$ and $w$
\State draw $\Gamma$ by \cref{lem:graph-interface} on $[4D]$ with allowance $\delta/16$
\State \Return $\textsc{PrecisionStage}(a,\eps,w,r,\delta/16,\Gamma)$
\end{algorithmic}
\end{algorithm}

\begin{proof}[Proof of \cref{thm:main}]
If $\eps=1$, \cref{alg:main} uses one oracle call, no reset and no pilot, and the interval $[0,1]$ covers every pair of numbers in $[0,1]$. Let $\eps<1$. By \cref{lem:anchor-supplied}, $\eps\le w\le1$ and $w\le2(f+\eps)$, so $t=\eps$ satisfies $t\le w$. If $F^*\le f$ and $R^*\le r$, the anchor satisfies $F(a)\le w$ except with probability $\delta/4$, and then \cref{prop:precision}(b) with $v=r$ and $\beta=\delta/16$ gives coverage except with probability $3\delta/16$. The total failure probability is at most $\delta/4+3\delta/16<\delta$.

For the resources, the anchor uses $\Ot((d+1)D/\epsb)$ resets with $\epsb=\min(1,f+\eps)\ge\eps$, and
\[
 \frac{D}{\epsb}\le\frac D\eps\le\frac{D(f+\eps)}{\eps^2}.
\]
In \eqref{eq:stage-cost}, $k_A=O((1+\log D)\log(eD/\delta))$ is polylogarithmic, $w\le2(f+\eps)$, $v=r$ and $t=\eps$, so the stage uses $\Ot((d+1)[D(f+\eps)+r]/\eps^2)$ resets; the term $\log(D/\beta)$ is absorbed into the polylogarithmic factor multiplying $d+1$. The anchor plays $\Ot(D)$ functions and the stage plays $\Ot(Dw/\eps)=\Ot(D(f+\eps)/\eps)$ functions, which gives $V=\Ot(D(1+f/\eps))$. The sample sizes and slot counts of the precision stage are numerical functions of $(d,D,\eps,\delta,f,r)$, since $w$ is a function of $(\eps,f)$. The batch sizes of the anchor depend on observed endpoints, but by the proof of \cref{lem:anchor-supplied} in \cref{app:anchor} they are bounded on every history by numbers that depend only on $(d,D,\eps,\delta,f)$. So every phase has a deterministic numerical upper bound, and the caps hold on every history, whether or not the supplied numbers are valid ceilings and whether or not the confidence events occur. \Cref{app:adaptation} proves \cref{prop:auxiliary}, including the complete accounting of oracle calls.
\end{proof}

\section{The lower-bound constructions}\label{sec:lower-mechanisms}

\Cref{app:lower} proves \cref{thm:lower,thm:natural-lower} with finite families of instances: a learner that succeeds with probability at least $4/5$ on every instance of a family must have a large expected cost on some instance. Each family uses one known loss rule, and its hidden parameters are carried by the truth function and the record law. Different terms use different families, and a maximum of separate lower bounds gives their sum up to a constant. We describe the mechanisms because they explain why the two ceilings and the two resources enter differently.

\paragraph{Three obstructions for resets.}
The term $dD/\eps$ comes from rare realizable threshold paths. With probability proportional to $\eps$, a record selects one of $d$ independent threshold blocks and follows the binary-search path of length $D$ towards the hidden threshold of that block. The address of each query depends only on the earlier branch decisions, so it reveals no future bit. A frozen pilot that has not yet seen a bit predicts it correctly with probability $1/2$, so it reveals at most two new bits in expectation before its first error. Identifying most of the $d$ thresholds requires $\Omega(dD)$ bits, hence $\Omega(dD/\eps)$ resets. These are familiar fresh-bit and symmetric-counterexample mechanisms; related decision-tree constructions appear in \citet[Example~2]{braverman2026}. We prove the exact stopped-record version rather than importing an equivalence-query theorem whose interface does not match.

For the term $dDf/\eps^2$, a hard record occurs with probability $f$. It selects a block and one of $K$ arms, asks the gate query of that arm, and then follows a threshold path on the arm. A member of the class opens exactly one gate per block. A discrepancy at a wrong gate costs nothing under the target loss, while a later discrepancy or a bad terminal arm costs one. Hence $R^*=0$ and $F^*\le f$, but finding enough good arms requires both matching the gate and paying the fresh-bit price on the path. Taking $K$ of order $f/\eps$ gives the term. Properness matters here: a pilot outside the class that opens every gate, or an output outside the class that rejects every gate, could avoid the obstruction. Contextual combination locks of a related kind appear in \citet{jia2023}; here both the observations and the target loss are those of the stopped-record experiment.

For the term $dr/\eps^2$, records have length one and carry hidden Bernoulli terminal descriptors. A good pilot can have zero discrepancy probability, but the terminal means of the candidates still have to be estimated to accuracy of order $\eps/r$ on records of probability $r$. Giving the learner complete records only makes it stronger, and a stopped-transcript relative-entropy bound gives the lower bound even then. This witness shows why the $r$ term carries no extra factor of the horizon.

\paragraph{Distinct functions rather than resets.}
The pilot floor $\Omega(D)$ comes from the threshold paths again: an unchanged pilot that meets the same deterministic record repeats the same stopped prefix and exposes nothing new, so only the first encounter of each pilot with each block can reveal new bits. The term $\Omega(Df/\eps)$ uses the free-gate family and counts completed arms in a comparison experiment in which every arm of one block is bad. The final output is allowed to be one additional unplayed guess. For expected costs, the proof truncates the number of distinct pilots and imposes a common finite reset cap, which exists because runs on the finite family terminate almost surely; no budget assumption is placed on the comparison experiment, which is not a promised instance. A lower bound on resets, or on the support of the final lottery, would not by itself bound the number of distinct functions used in training.

\paragraph{The natural-loss boundary.}
For natural discrepancy at $D\ge3$, a diluted family of gated threshold trees, in which some blocks have a winning arm and others have none, gives both the reset term and the pilot term. At $D=2$, a separate hidden-arm family gives the pilot term. In that family the class predicts one at every second query; adding arbitrary further functions to the class could remove the hardness, which is why it is a statement about a specific class. At $D=1$ every query and its label are displayed whatever pilot is used, so one fixed pilot already yields an ordinary sample of the natural losses of all candidates, and the natural pilot lower bound fails. A general loss at $D=1$ can still hide a terminal descriptor after a discrepancy, and the general pilot lower bound remains valid there. This is why the two pilot lower bounds have different horizon ranges.

\section{Related work}\label{sec:related}

We compare exact interfaces: who chooses what before a record is drawn, what is displayed, what is counted, and what the output must satisfy. Several close results become legal special cases of our experiment after a reduction. These reductions are our derivations, not theorems stated in the cited papers.

\paragraph{Interactive imitation with first-step mixtures.}
\citet[Algorithm~2 and Theorem~4]{li_zhang_v1} study finite-class realizable interactive imitation learning with trajectory-wise annotation, using first-step mixtures: a base policy is drawn at the start of an episode and kept fixed. Their Lemma~12 partitions trajectories by the first disagreement. Their learner takes the finite policy class itself as input and runs an online learner over the probability simplex on it. An absorbing-state specialization turns this into a legal stopped-feedback baseline for a supplied finite evaluable class in the attained-realizable setting, or under equivalent access to updates and sampling over the whole class. (Finiteness alone does not give this access under \cref{ass:oracle}, whose empty-list call returns a single representative.) Represent a truthful record as a trajectory, move to an uninformative absorbing state with a known action after the first mismatch, and let the trajectory annotation reveal the same informative prefix as our display; the tail after absorption is known padding, not an actual query after an error. A first-step draw commits to a member of the class before the trajectory, which is exactly a legal proper pilot. Sampling components of the returned mixture and validating each on independent resets in which it is its own pilot yields a single proper output with small discrepancy probability. Since the optimum is then known to be zero, the interval $[0,\eps]$ is a joint certificate for this output in the natural-loss specialization. What this baseline does not provide is the subject of \cref{thm:main}: classes of finite VC dimension accessed only through \cref{ass:oracle}, a general fixed loss with possibly unattained and nonzero infima, a certificate for an unknown optimum, and counting of all distinct functions played during acquisition.

\paragraph{Learning verifiers from annotated traces.}
\citet[Definition~C.1, Theorem~C.2 and the VC extension that follows it]{balcan_v1} already give proper agnostic learning of verifiers with coefficient one from complete annotated traces, together with an extension to classes of finite VC dimension; in the published version these are Definition~F.1 and Theorem~F.2 \citep{balcan_published}. Properness, agnostic competition and finite VC dimension are therefore not distinguishing features of the present work. Our additional obligation is to acquire enough information when the chosen pilot's earlier discrepancy censors the rest of the trace, including its terminal information. An empirical minimizer over complete traces does not implement this acquisition under \cref{ass:oracle}, because the traces are not supplied.

\paragraph{Proper policy search with resets.}
\citet[Sections~2.1--2.2, Theorem~4.2 and Figure~2]{mansour1999} gives near-best policy search over a restricted class of strategies, in a partially observable environment that is accessed only by actions and resets to the initial state, assuming a proper (in his terms, realizable) mistake-bounded online learner for the class. Every generated trajectory follows one fixed strategy of the class, and the output competes with the best expected bounded return in the class. Making the first discrepancy absorbing, paying the reward $1-\lambda(T,\tau_h)$ once at the observed stopping outcome (a discrepancy, or survival, including an empty record) and padding the rest with known steps of reward zero, we obtain from it a legal stopped-record special case with frozen proper pilots and agnostic bounded-loss competition: the expected return of $h$ is $1-R(h)$, and neither a hidden suffix nor a terminal descriptor after a discrepancy is used. The mistake-bound assumption is stronger than finite VC dimension: thresholds on the real line have VC dimension one, but an adversary that chooses each new query between the remaining candidate thresholds forces a mistake in every round. The restriction operations of the online learner are not supplied by \cref{ass:oracle}, and the running time stated there is exponential in the mistake bound. That result does not give the full statement of \cref{thm:main}: classes of finite VC dimension accessed only through \cref{ass:oracle}, with the simultaneous localized reset and distinct-pilot orders together with the joint certificate.

\paragraph{Stopping-rank covers.}
\citet[Lemma~3.1, Algorithms~1 and~2, Proposition~3.3]{goel_pei_wang_2605.09183} give rank and exchangeability covers, a minimax formulation and a no-regret game algorithm for selective imitation under dynamics shift; their rank argument in Lemma~B.2 is an important antecedent for reasoning about stopping positions. Their game is trained and evaluated on available complete trajectories. The no-regret subroutine (their Algorithm~2) returns a distribution over validator sets of bounded size, and the final selector (their Algorithm~1) uses the union of logarithmically many sets sampled from it. These per-set and final-selector size bounds are not bounds on all hypotheses used while a stopped-data cover is acquired. Our acquisition and propagation arguments address that difference; we do not claim to have invented rank covers or the no-regret construction.

\paragraph{Sparse policy covers and a legal empirical specialization.}
\citet[Algorithm~1, Eq.~(7), Proposition~4.1, Theorem~4.1 and Appendix~H.1]{amortila2024} provide a sparse policy-cover planning engine. Their planner can be applied to a known empirical one-step model built from legally displayed observations and observable nonnegative weights: the sample indices become the states, the two correctness outcomes become the actions, and the pattern oracle performs the policy optimizations. This reduction, followed by uniform relative concentration, recovers the rate powers of horizon one and of suitable horizon-two cases that are supplied with an evaluable finite list of proper pilots with pointwise first-query coverage, for example a list that realizes both labels at every possible first query. That the population record law is unknown is no obstacle, because the empirical model is known. This specialization is our derivation from their planning theorem; it does not by itself supply the truncated-rank acquisition and the growing-horizon composition used here. Inverse-propensity certificates, policy covers, and discovery followed by evaluation have a substantial literature. PC-PG averages the visitation distributions of earlier policies \citep{PCPG2020_pcpg2020_arxiv_v1}, but it uses rollout and continuation access and a critic transfer assumption that differ from our frozen original-proper pilots. POPLER and its extension to infinite policy classes \citep[Appendix~G.4]{jia2023} also precede discover-then-evaluate strategies; their sunflower assumptions and the switching between reacher and core policies within an episode are not available here.

\paragraph{Proper online learning with unknown base measure.}
Representatives of historical patterns and proper learning without knowledge of the law are also known. R-Cover is a proper algorithm for classes of finite VC dimension that commits to a function before each new query and does not know the base measure \citep[Theorem~4 and Algorithm~1]{blanchard2024_v1}; the later version, with a new title, states the classification result as Theorem~2 \citep{blanchard2024}. Its feedback is revealed query by query, whereas here one pilot is frozen for an entire stopped record. Our law-dependent nets are used only in the analysis, while the dictionaries used by the learner are acquired on displayed queries.

\paragraph{Classical tools.}
Given the legal gain and the observed dictionary, the reset and support orders of one acquisition stage follow from the standard exponential-weights potential, bounded martingale concentration and one sparsification \citep{helmbold1998,freund1997}; we credit this whole conversion, not only the form of the update. The propensities and mixtures are related to exploration in contextual bandits \citep{dudik2011_v1}. \Citet[Section~3.6, with Sections~2.4 and~3.4]{agarwal2014} also bound the total number of policies that ever receive positive weight in their warm-start algorithm, a form of accounting over the whole run rather than only the final support. That count differs from the resource $V$ counted here: positive weight is not the same as actual use as a pilot, their action distributions add uniform mass on individual actions, which a frozen proper pilot need not be able to implement, and their objective is regret. The stage graph is a classical depth-robust graph \citep{EGS1975_egs1975_journal_static_renyi,ABP2017_2018_abp2017_arxiv_v2,BCLS2021_2022_blocki2022_explicit}. The relevant lemmas of Alwen, Blocki and Pietrzak are already in their 2017 preprint, and the explicit construction of Blocki, Cinkoske, Lee and Son was public in 2021, before its 2022 proceedings version. Theorem~9 of the latter paper is credited there to the argument of Erd\H{o}s, Graham and Szemer\'edi, applied to the explicit bipartite expanders constructed in that paper. Apple tasting \citep{helmbold2000}, DAgger \citep{ross2011} and expert intervention learning \citep{spencer2021}, which includes Boolean stop signals as feedback, show that learning from selective feedback and from interventions is not a new paradigm; their action choices, annotation rules and timing within a run must be checked before a theorem is transferred to the present experiment.

\paragraph{Summary.}
The comparisons above concern the stated observation and oracle interfaces; none of the works discussed, and none of the reductions to them described here, gives the complete statement of \cref{thm:main}. The contribution is not a new optimization, concentration or graph primitive, but the joint stopped-data theorem: legal acquisition at truncated ranks, transfer to the whole class, propagation of visibility under correlated failures, harmonic recovery of the mixture weight, and the two simultaneous resource caps together with the joint certificate.

\section{Limitations and open questions}\label{sec:limitations}

\paragraph{Computation and constants.}
The learner calls an oracle that lists one representative for every label pattern of $\cH$ on the displayed queries. For a general class this may require exponential time, and we prove no bound on running time, memory or the precision of the real-number computations. The sufficient constants are large. For example, \cref{alg:precision} sizes every design with the predetermined cap $k_A$ of \eqref{eq:degree-cap}, not with the realized indegrees, and this cap equals the complete-graph value $S-1$ at all horizons below a few thousand. Evaluating \eqref{eq:degree-cap} with $S=4D$, the first strict crossing of this predetermined budget, that is, the first horizon at which $k_A<S-1$, is $D=5086$ when $\beta=10^{-2}$ and $D=7802$ when $\beta=10^{-9}$; for every $\beta\in[10^{-9},10^{-2}]$ the first crossing lies between these two values, because at each horizon $k_A$ is nonincreasing in $\beta$. The strict inequality need not persist: for $\beta=10^{-9}$ the cap equals $S-1$ again for $8192\le D\le8309$, where the factor $\lfloor\log_2(2S)\rfloor+1$ increases. Wherever the cap equals $S-1$, in particular below the first crossing, the design budget in \eqref{eq:stage-cost} grows like $D^2$ up to logarithmic factors, so the linear dependence on $D$ in \cref{thm:main} is asymptotic. These statements concern the budget charged by the procedure; a cap equal to $S-1$ does not mean that the random graph of \cref{lem:graph-interface} is complete. Using the complete graph for every $D$ gives a deterministic stage graph, the reset bound $\Ot((d+1)[D^2(f+\eps)+r]/\eps^2)$ and the same pilot bound (\cref{rem:graph-variants}).

\paragraph{The model.}
Three assumptions are substantive. First, all runs use one common record law, which does not depend on the pilot. If the pilot could change the distribution of the rest of a record, the identities that evaluate several hypotheses on one reference record, for instance $\Pr(\tau_g\ge l\mid T)=Q_L(T,l)$, can fail. Second, the true labels are exact and fixed. A noisy checker that stops a run at a correct decision can invalidate the conditional-mean identity of the displayed gain and the moment and bias bounds of the final score. Third, the pilot is frozen for the whole run, and there are no continuations after an error and no resets to an arbitrary prefix. Extending the results beyond these assumptions requires new assumptions and new proofs. Nothing here is a statement about deployed systems or empirical performance.

\paragraph{Open questions.}
\begin{enumerate}[label=(\arabic*)]
\item Are there computationally efficient learners with the same two resource orders for structured classes, for instance thresholds or other classes with efficient consistency oracles?
\item Which weakening of the exact-truth and common-law assumptions still permits joint guarantees of the same type?
\item The lower bounds are coordinatewise at fixed confidence (\cref{rem:lower-reading}). What is the optimal dependence of the two resources on $\delta$, and what is the full set of achievable pairs $(N,V)$ on a common instance? The learners display at most $DN$ labels; is this upper bound on the number of displayed labels optimal?
\item Is there an explicit stage graph with small constants, so that the linear dependence on $D$ appears already at moderate horizons?
\end{enumerate}

\section*{Declarations}
Contributions and use of AI: AI tools assisted with mathematical exploration, proof development, computational checks and manuscript preparation. The author takes responsibility for the final content.

\bibliographystyle{plainnat}
\bibliography{references}

\clearpage
\appendix
\section{VC tools for latent records}\label{app:vc}

In this appendix a record $T=((z_j,y_j)_{j=1}^M,A)$, $0\le M\le D$, is a complete probability-space object; the learner sees only its displays. Write $\cP$ for the record law. For $h,h'\in\cH$ let
\[
 E(h,h')=\{T:\ h(z_j)\ne h'(z_j)\text{ for some }j\le M\}
\]
be the event that $h$ and $h'$ disagree somewhere on the record. \Cref{ass:measurable} makes all events below measurable.

\subsection{Trace counts}

\begin{lemma}[Trace counts]\label{lem:vc-traces}
Let $\cH$ have VC dimension at most $d\ge0$.
\begin{enumerate}[label=(\roman*)]
\item On any list of $q$ queries, $\cH$ realizes at most $(q+1)^d$ label patterns.
\item Consider a Boolean test of a pair $(h,h')$ on an observation which is determined by the values of $h$ and $h'$ at the at most $D$ queries of that observation, together with fixed data of the observation that do not depend on $(h,h')$. On $n$ observations, the family of such tests over all pairs realizes at most $(nD+1)^{2d}$ distinct indicator vectors. This applies to the events $E(h,h')$ on records, and to the events $O_{hh'}$ of \cref{lem:gain}(d) on displays.
\end{enumerate}
\end{lemma}

\begin{proof}
(i) Let $\mathfrak P$ be the set of patterns of $\cH$ on distinct points $x_1,\ldots,x_q$ (repeated queries only reduce the count), and write $\phi(q,d)=\sum_{j=0}^{\min(d,q)}\binom qj$. We show $|\mathfrak P|\le\phi(q,d)$ for every set $\mathfrak P$ of patterns that shatters no set of size $d+1$, by induction on $q+d$. If $q=0$ there is at most one pattern. If $d=0$, no point is shattered, so all patterns agree at every point and there is at most one pattern. Otherwise let $\mathfrak P'$ be the set of restrictions to $x_1,\ldots,x_{q-1}$, and $\mathfrak P''\subseteq\mathfrak P'$ the set of restrictions that have both extensions at $x_q$ in $\mathfrak P$. Then $|\mathfrak P|=|\mathfrak P'|+|\mathfrak P''|$. The set $\mathfrak P'$ shatters no set of size $d+1$. If $\mathfrak P''$ shattered a set $Y$, then $\mathfrak P$ would shatter $Y\cup\{x_q\}$, so $\mathfrak P''$ shatters no set of size $d$. By induction $|\mathfrak P|\le\phi(q-1,d)+\phi(q-1,d-1)=\phi(q,d)$. Finally, for $d\ge1$ the map sending a subset $\{i_1<\cdots<i_k\}\subseteq\{1,\ldots,q\}$ with $k\le d$ to $(i_1,\ldots,i_k,0,\ldots,0)\in\{0,\ldots,q\}^d$ is injective, so $\phi(q,d)\le(q+1)^d$; for $d=0$ both sides equal one.

(ii) The $n$ observations contain at most $nD$ query positions. The indicator vector of the test for $(h,h')$ is determined by the pattern of $h$ and the pattern of $h'$ on the list of these queries, and by (i) there are at most $((nD+1)^d)^2$ such pairs of patterns.
\end{proof}

\subsection{Missed sets and a latent net}

\begin{lemma}[Uniform missed-set bound]\label{lem:vc-missed}
Let $\mathcal A$ be a class of events on an observation space such that on any $m$ observations it realizes at most $G(m)$ indicator vectors. Let $X_1,\ldots,X_n$ be independent with common law $\mu$, and let $0<\chi\le1$ with $n\chi\ge8$. Then
\begin{equation}\label{eq:vc-missed}
 \Pr\bigl\{\exists A\in\mathcal A:\ \mu(A)>\chi,\ X_i\notin A\ (1\le i\le n)\bigr\}
 \le2G(2n)\,2^{-n\chi/2}\le2G(2n)\,e^{-n\chi/4}.
\end{equation}
The same bound holds conditionally on a previously frozen history, if the fresh observations are conditionally independent with a common conditional law and the bound $G$ is deterministic.
\end{lemma}

\begin{proof}
Let $\mathsf E$ be the event in \eqref{eq:vc-missed}. On $\mathsf E$ select a witness $A^*\in\mathcal A$ with $\mu(A^*)>\chi$ and no $X_i$ in $A^*$. Let $X_1',\ldots,X_n'$ be an independent copy of the sample. Conditionally on $X$, the number $Z'$ of $i$ with $X_i'\in A^*$ is binomial with mean $np$, $p=\mu(A^*)>\chi$, and variance at most $np$. By Chebyshev's inequality,
\[
 \Pr(Z'<np/2\mid X)\le\frac{np}{(np/2)^2}=\frac4{np}\le\frac4{n\chi}\le\frac12 .
\]
Hence $\Pr(\mathsf E)\le2\Pr(\mathsf E')$, where $\mathsf E'$ is the event that some $A\in\mathcal A$ contains none of the $X_i$ and at least $n\chi/2$ of the $X_i'$. Now exchange $X_i$ and $X_i'$ independently for each $i$ with probability $1/2$; this does not change the joint law. Condition on the unordered pairs $\{X_i,X_i'\}$. There are at most $G(2n)$ indicator vectors of events of $\mathcal A$ on the $2n$ pooled observations. Fix one with at least $n\chi/2$ hits. If some pair contains two hits, the first sample cannot avoid the event. Otherwise the hits lie in distinct pairs, and all of them land in the second sample with probability at most $2^{-n\chi/2}$. A union bound over indicator vectors gives $\Pr(\mathsf E'\mid\text{pairs})\le G(2n)2^{-n\chi/2}$, and the first inequality follows. The second uses $(\log2)/2\ge1/4$. The argument applies verbatim conditionally on a frozen history.
\end{proof}

For $0<\chi\le1$ let $\nu(\chi)$ be the least positive integer $v$ with
\begin{equation}\label{eq:net-size}
 v\chi\ge8\qquad\text{and}\qquad2(2vD+1)^{2d}e^{-v\chi/4}<1,
\end{equation}
and put $H_\chi=(\nu(\chi)D+1)^d$. With $\chi=1/n$ these are the numbers $\nu_n$ and $H_n$ of \eqref{eq:net-parameters}.

\begin{lemma}[A latent-record net]\label{lem:vc-net}
For every $0<\chi\le1$ there is a finite set $\cN_\chi\subseteq\cH$, possibly depending on the unknown law, with $|\cN_\chi|\le H_\chi$, such that every $h\in\cH$ has some $h'\in\cN_\chi$ with $\cP(E(h,h'))\le\chi$. Moreover
\[
 \nu(\chi)\le\Bigl\lceil\frac8\chi\max\Bigl\{1,\ \log4+2d\log\frac{48dD}{\chi}\Bigr\}\Bigr\rceil
 \quad(d\ge1),\qquad \nu(\chi)=\lceil8/\chi\rceil\quad(d=0),
\]
so that $\nu(\chi)=O((d+1)\chi^{-1}\log(e(d+1)D/\chi))$ and $\log H_\chi=O(d\log(e(d+1)D/\chi))$.
\end{lemma}

\begin{proof}
Let $v=\nu(\chi)$ and apply \cref{lem:vc-missed} to the events $E(h,h')$ on $v$ independent records, with $G(m)=(mD+1)^{2d}$ from \cref{lem:vc-traces}(ii). Since the right side of \eqref{eq:vc-missed} is less than one, there is a realization $T_1,\ldots,T_v$ such that every pair $(h,h')$ that agrees at all queries of these records has $\cP(E(h,h'))\le\chi$. Fix such a realization. Let $\cN_\chi$ contain one member of $\cH$ for each pattern of $\cH$ on the at most $vD$ queries of $T_1,\ldots,T_v$; by \cref{lem:vc-traces}(i), $|\cN_\chi|\le(vD+1)^d$. Every $h$ shares its pattern with some $h'\in\cN_\chi$, and this pair agrees at all sampled queries, so $\cP(E(h,h'))\le\chi$. This is an existence argument used only in proofs; the learner never constructs $\cN_\chi$ and makes no oracle call on latent queries.

For the bound on $\nu(\chi)$, let $d\ge1$. Since $2vD+1\le3vD$, the second condition in \eqref{eq:net-size} holds if $v\chi/4\ge\log4+2d\log(3Dv)$, that is, if $v\ge(4/\chi)(\log4+2d\log(3Dv))$. By \cref{lem:inversion} with $A=4/\chi$, $B=\log4$, $k=2d$ and $c=3D$, this holds for every $v\ge(8/\chi)(\log4+2d\log(48dD/\chi))$. Together with $v\ge8/\chi$ this gives the bound. For $d=0$ the second condition reads $v\chi>4\log2$, which follows from $v\chi\ge8$. The order statements follow by substitution.
\end{proof}

\subsection{Rare counts}

\begin{lemma}[Uniform rare-event counts]\label{lem:vc-rare}
Under the trace assumption of \cref{lem:vc-missed}, let $X_1,\ldots,X_N$ be independent with law $\mu$, let $\chi\ge0$ and $\kappa\ge\max\{4,4N\chi\}$. Then
\[
 \Pr\Bigl\{\exists A\in\mathcal A:\ \mu(A)\le\chi,\ \sum_{i=1}^N\1_A(X_i)>\kappa\Bigr\}\le2G(2N)\,e^{-\kappa/18}.
\]
In particular, for the events $E(h,h')$ of mass at most $1/N$ on $N$ independent records, all counts are at most
\[
 \kappa_N=\max\Bigl\{4,\ 18\log\frac{4(2ND+1)^{2d}}{\beta}\Bigr\}
\]
except on an event of probability at most $\beta/2$.
\end{lemma}

\begin{proof}
Let $\mathsf E$ be the event in question and on $\mathsf E$ select a witness $A^*$ with $\mu(A^*)\le\chi$ and count $O>\kappa$ in the sample. For an independent copy of the sample, the count $O'$ of $A^*$ has conditional mean at most $N\chi\le\kappa/4$, so by Markov's inequality $O'\le\kappa/2$ with conditional probability at least $1/2$. Hence $\Pr(\mathsf E)$ is at most twice the probability that some $A\in\mathcal A$ has $O_A>\kappa$ and $O_A'\le\kappa/2$. On this event put $L_A=O_A+O_A'$; then $L_A>\kappa$ and, since $O_A'<O_A/2$, also $O_A-O_A'>L_A/3$. Exchange the members of each pair independently as before, and condition on the unordered pairs. For a fixed indicator vector on the pooled sample, $L_A$ is fixed and $O_A-O_A'=\sum_{i\in I}\epsilon_i$, where $I$ is the set of pairs with exactly one hit, $|I|\le L_A$, and the $\epsilon_i$ are independent uniform signs. Since $\cosh x\le e^{x^2/2}$, the exponential Markov inequality with parameter $s/|I|$ gives $\Pr(\sum_{i\in I}\epsilon_i>s)\le e^{-s^2/(2|I|)}$, so with $s=L_A/3$ the conditional probability is at most $e^{-L_A/18}\le e^{-\kappa/18}$. (If $I$ is empty the sum is zero and the probability vanishes.) A union bound over the at most $G(2N)$ indicator vectors and the factor two prove the claim. For the last assertion take $\chi=1/N$, so that $4N\chi=4$, and note $2(2ND+1)^{2d}e^{-\kappa_N/18}\le\beta/2$.
\end{proof}

\begin{remark}[Transfer and adaptively censored data]\label{rem:vc-transfer}
If $h$ and $h'$ agree at every query of a record, their stopping ranks and losses on it agree, including the terminal loss. Every score computed from the record, from frozen auxiliary functions and lotteries, and from the candidate only through its stopping rank, also agrees. Hence a score with range $[0,W]$ differs between $h$ and $h'$ by at most $W\1_{E(h,h')}$.

Fresh records remain independent with law $\cP$ when pilots are chosen from past observations, because the record law does not depend on the pilot. One may therefore draw each complete record independently, reveal only the pilot's display, and pad a stopped procedure to its deterministic cap by arbitrary predictable pilots. \Cref{lem:vc-rare} then controls the number of score differences on this independent array of latent records even when the displays are adaptive. It does not make adaptive scores independent; their concentration uses predictable conditional means, as in \cref{app:anchor,app:acquisition}.
\end{remark}

\subsection{Two elementary tools}

\begin{lemma}[Inverting logarithmic conditions]\label{lem:inversion}
Let $A,k,c>0$ and $B\ge0$. Every real $m>0$ with $m\ge2A\bigl(B+k\log(2Ack)\bigr)$ satisfies $m\ge A\bigl(B+k\log(cm)\bigr)$.
\end{lemma}

\begin{proof}
Since $\log y\le y-1$,
\[
 \log(cm)=\log(2Ack)+\log\frac m{2Ak}\le\log(2Ack)+\frac m{2Ak}-1 .
\]
Hence $A(B+k\log(cm))\le AB+Ak\log(2Ack)+m/2-Ak<AB+Ak\log(2Ack)+m/2\le m$, where the last step is the hypothesis divided by two.
\end{proof}

\begin{lemma}[Exponential bounds for adapted bounded variables]\label{lem:supermartingale}
Let $(\mathcal F_t)_{t\ge0}$ be a filtration and $Y_t\in[0,1]$ be $\mathcal F_t$-measurable with $\mu_t=\E[Y_t\mid\mathcal F_{t-1}]$. Let $\vartheta>0$.
\begin{enumerate}[label=(\alph*)]
\item $\E[\exp(\vartheta Y_t)\mid\mathcal F_{t-1}]\le\exp((e^\vartheta-1)\mu_t)$ and $\E[\exp(-\vartheta Y_t)\mid\mathcal F_{t-1}]\le\exp(-(1-e^{-\vartheta})\mu_t)$.
\item For every $u\ge0$ and every integer $n\ge1$,
\[
 \Pr\Bigl\{\exists\,s\le n:\ \sum_{t=1}^s\bigl(\vartheta Y_t-(e^\vartheta-1)\mu_t\bigr)\ge u\Bigr\}\le e^{-u},
\]
and the same holds for $\sum_{t\le s}(-\vartheta Y_t+(1-e^{-\vartheta})\mu_t)$.
\end{enumerate}
\end{lemma}

\begin{proof}
(a) By convexity, $e^{\vartheta y}\le1+(e^\vartheta-1)y$ and $e^{-\vartheta y}\le1-(1-e^{-\vartheta})y$ for $y\in[0,1]$. Take conditional expectations and use $1+x\le e^x$. (b) By (a), $M_s=\exp\sum_{t\le s}(\vartheta Y_t-(e^\vartheta-1)\mu_t)$ is a nonnegative supermartingale with $M_0=1$. Let $\sigma$ be the first $s\le n$ with $M_s\ge e^u$, and $\sigma=n$ if there is none. Optional stopping at the bounded time $\sigma$ gives $\E M_\sigma\le1$, and $M_\sigma\ge e^u$ on the event in question, so its probability is at most $e^{-u}$. The second sum is treated in the same way.
\end{proof}

\section{Ownership and the coarse anchors}\label{app:anchor}

This appendix proves \cref{lem:anchor-supplied,lem:anchor-adaptive}. The basic tool is an ownership procedure for a general fixed loss $\lambda$; the anchors use it with the natural loss. Throughout, $\ell_h=\lambda(T,\tau_h)$, $R(h)=\E\ell_h$, $R^*=\inf_\cH R$ and $F(h)=\Pr(\tau_h\le M)$. All oracle calls are made on displayed queries only.

\subsection{Anchored ownership}

Let $a\in\cH$ be fixed before the procedure, and let $r,w\in[0,1]$, $0<t<1$ and $0<\alpha<1$ be numbers. Put
\[
 \eta=\frac t{32},\quad K=\max\Bigl\{1,\Bigl\lceil\frac{Dw}\eta\Bigr\rceil\Bigr\},\quad J=K+1,\quad
 \theta=\frac\eta{8(r+\eta)}\le\frac18,\quad \chi=\frac\eta{32J},
\]
let $H_\chi$ be as in \cref{lem:vc-net}, and let $B$ be the least positive integer with
\begin{align}
 B\eta&\ge16,&
 \frac{\log(8H_\chi/\alpha)}{B\theta}&\le\frac\eta4,&
 \frac{\log(8K/\alpha)}{B\theta}&\le\eta,\label{eq:anc-batch-a}\\
 &&2(2JBD+1)^{2d}e^{-B\eta/72}&\le\frac\alpha4.&&\label{eq:anc-batch-b}
\end{align}
These conditions are law-free; the net $\cN_\chi$ itself is not an input. For pilots $g_1,g_2,\ldots$ and a record $T$ define the \emph{ownership score} of $h$ in batch $j$,
\begin{equation}\label{eq:anc-score}
 Y_{j,h}(T)=\ell_h(T)\,\1\Bigl\{\max_{i<j}\tau_{g_i}(T)<\tau_h(T)\le\tau_{g_j}(T)\Bigr\},
\end{equation}
where the maximum of an empty list is $0$. If $T_{j,1},\ldots,T_{j,B}$ are the records of batch $j$, put $S_m(h)=B^{-1}\sum_{j\le m}\sum_{k\le B}Y_{j,h}(T_{j,k})$.

\begin{lemma}[Anchored ownership]\label{lem:ownership}
\Cref{alg:ownership} returns, on every history, a member of $\cH$ and an ordered interval in $[0,1]$ of width at most $t$. It uses at most $J$ batches of $B$ resets, at most $K$ oracle calls and at most $J$ distinct pilots, and
\begin{equation}\label{eq:anc-batch-order}
 B=O\!\left(\frac{r+\eta}{\eta^2}\Bigl[(d+1)\log\frac{e(d+1)D}\eta+\log\frac e\alpha\Bigr]\right).
\end{equation}
If $R^*\le r$ and $F(a)\le w$, then with probability at least $1-\alpha$ the interval contains $R^*$ and the risk of the returned function. The same holds conditionally on any earlier history that fixes $a,r,w$ and satisfies these two inequalities.
\end{lemma}

\begin{algorithm}[t]
\caption{$\textsc{Ownership}(a,r,w,t,\alpha)$}\label{alg:ownership}
\begin{algorithmic}[1]
\State compute $\eta,K,J,\theta,\chi,B$ as above; $g_1\gets a$; play $g_1$ on a batch of $B$ fresh resets
\For{$m=1,\ldots,K$}
 \State call the oracle on all queries displayed in batches $1,\ldots,m$; let $h_m$ be a returned member minimizing $S_m$
 \State $g_{m+1}\gets h_m$; play $g_{m+1}$ on a batch of $B$ fresh resets; $\widetilde R_m\gets B^{-1}\sum_{k\le B}\ell_{h_m}(T_{m+1,k})$
 \State $L_m\gets(S_m(h_m)-\eta)/(1+\theta)$, \ $U_m\gets(\widetilde R_m+\eta)/(1-\theta)$; clip each to $[0,1]$ separately
 \If{the clipped pair is ordered and has width at most $t$} \Return $h_m$ and the clipped interval \EndIf
 \State \Comment{otherwise batch $m+1$ is kept as the next ownership batch}
\EndFor
\State \Return $a$ and $[0,0]$ \Comment{cap exhausted}
\end{algorithmic}
\end{algorithm}

\begin{proof}
\emph{Sample size.} Condition~(i) of \eqref{eq:anc-batch-a} is $B\ge16/\eta$, (ii) is $B\ge32(r+\eta)\log(8H_\chi/\alpha)/\eta^2$, and (iii) is $B\ge8(r+\eta)\log(8K/\alpha)/\eta^2$. Since $2JBD+1\le3JDB$, condition \eqref{eq:anc-batch-b} holds if $B\eta/72\ge\log(8/\alpha)+2d\log(3JDB)$, and by \cref{lem:inversion} (with $A=72/\eta$, $B_0=\log(8/\alpha)$, $k=2d$, $c=3JD$) this holds for every $B\ge(144/\eta)(\log(8/\alpha)+2d\log(864dJD/\eta))$; for $d=0$ the threshold is $(72/\eta)\log(8/\alpha)$. Hence $B$ is at most one more than the largest of these four thresholds. Since $w\le1$ and $D/\eta\ge32$, we have $J\le D/\eta+2\le3D/\eta$ and $K\le J$, so \cref{lem:vc-net} gives $\log H_\chi=O(d\log(e(d+1)D/\eta))$. Using $1/\eta\le(r+\eta)/\eta^2$, we obtain \eqref{eq:anc-batch-order}.

\emph{Observability.} Consider a record $T$ of batch $j$, displayed by $g_j$. If $h$ disagrees with some displayed pair, its first disagreement $q$ equals $\tau_h$, and $\ell_h$ is determined by the pairs up to $q$. The condition $\max_{i<j}\tau_{g_i}<q$ says that each earlier pilot disagrees with some pair strictly before $q$, which is decidable from the same display, and $\tau_h=q\le n_{g_j}\le\tau_{g_j}$. If $h$ agrees with all displayed pairs and $g_j$ errs, then $\tau_h>\tau_{g_j}$ and the score is zero; nothing hidden is evaluated. If $g_j$ survives, the whole record and its descriptor are displayed and the score is computable. On an empty record all functions survive at rank one, so only the first batch owns its loss; a discrepancy at the last actual query is treated as a discrepancy, not as survival. Consequently $S_m(h)$ depends on $h$ only through its values at the queries displayed in batches $1,\ldots,m$, and a minimizer over the representatives returned by the oracle is an exact empirical minimizer over $\cH$; at most $(mBD+1)^d$ representatives are needed.

\emph{Totalization.} If the procedure returns in the loop, the interval is ordered and has width at most $t$ by the test. Otherwise it returns $a$ and $[0,0]$. In both cases the output is a member of $\cH$. An invalid endpoint pair is never accepted; the fallback $[0,0]$ is a convention for histories on which no certificate was found and carries no coverage claim.

\emph{Predictable means.} Let $\mu_j(h)$ be the conditional mean of one score of batch $j$ given everything before that batch. The pilots $g_1,\ldots,g_j$ are then fixed and the record is fresh, so
\[
 \mu_j(h)=\E_T\Bigl[\ell_h\,\1\Bigl\{\max_{i<j}\tau_{g_i}<\tau_h\le\tau_{g_j}\Bigr\}\Bigr],
\]
where $\E_T$ integrates a separate reference record with the realized pilots held fixed. The same value is the conditional mean of every row of batch $j$ given the earlier rows of that batch. Let $\bar\tau_m(T)=\max_{j\le m}\tau_{g_j}(T)$ and $C_m(h)=\sum_{j\le m}\mu_j(h)$. On one reference record, the events in the indicators are disjoint for different $j$, and their union is $\{\tau_h\le\bar\tau_m\}$ (the first $j$ with $\tau_{g_j}\ge\tau_h$ owns the loss). Hence
\begin{equation}\label{eq:anc-compensator}
 C_m(h)=\E_T\bigl[\ell_h\1\{\tau_h\le\bar\tau_m\}\bigr]\le R(h).
\end{equation}
This is an identity of predictable compensators with the realized functions held fixed; it is not a statement about the conditional law of old records given the final list of pilots.

\emph{Uniform martingale bound.} For the analysis, continue the procedure after it returns by playing arbitrary predictable pilots until $J$ batches have been used; this does not change any event concerning the actual run. Fix $h'\in\cN_\chi$. The rows $Y_{j,h'}(T_{j,k})$, taken in order, lie in $[0,1]$ and have conditional means $\mu_j(h')$. Apply \cref{lem:supermartingale}(b) with $\vartheta=\theta$, $u=\log(8H_\chi/\alpha)$, to both signs. Since $(e^\theta-1)/\theta\le1+\theta$ and $(1-e^{-\theta})/\theta\ge1-\theta$ for $0<\theta\le1$, at every batch boundary $m\le J$,
\begin{equation}\label{eq:anc-net-martingale}
 |S_m(h')-C_m(h')|\le\theta C_m(h')+\frac{\log(8H_\chi/\alpha)}{B\theta}\le\theta C_m(h')+\frac\eta4,
\end{equation}
except with probability $2e^{-u}$. A union bound over $h'\in\cN_\chi$ gives failure at most $2H_\chi e^{-u}=\alpha/4$. There is no factor for the number of batches.

\emph{Transfer to the class.} Apply \cref{lem:vc-rare} to the $N=JB$ independent latent records of the padded run, with mass $\chi$ and threshold $\kappa=B\eta/4$. Here $\kappa\ge4$ and $4N\chi=B\eta/8\le\kappa$, and by \eqref{eq:anc-batch-b}, outside probability $\alpha/4$, every event $E(h,h')$ of mass at most $\chi$ contains at most $\kappa$ of these records. For $h\in\cH$ choose $h'\in\cN_\chi$ with $\cP(E(h,h'))\le\chi$. By \cref{rem:vc-transfer}, the two ownership scores can differ only on records in $E(h,h')$, whatever the pilots, and each difference is at most one, so $|S_m(h)-S_m(h')|\le\kappa/B=\eta/4$. For the compensators, the integrands in \eqref{eq:anc-compensator} for $h$ and $h'$ agree outside $E(h,h')$ and lie in $[0,1]$, so $|C_m(h)-C_m(h')|\le\chi$, not $m\chi$. Combining with \eqref{eq:anc-net-martingale} and $(1+\theta)\chi\le\eta/2$,
\begin{equation}\label{eq:anc-uniform}
 |S_m(h)-C_m(h)|\le\theta C_m(h)+\eta\qquad\text{for every }m\le J\text{ and every }h\in\cH .
\end{equation}

\emph{Validation.} Each candidate $h_m$ is fixed before batch $m+1$, whose $B$ losses $\ell_{h_m}(T_{m+1,k})$ are then independent with mean $R(h_m)$. \Cref{lem:supermartingale}(b) at the final time with $\vartheta=\theta$ and $u=B\theta\eta$, for both signs, gives
\begin{equation}\label{eq:anc-validation}
 |\widetilde R_m-R(h_m)|\le\theta R(h_m)+\eta
\end{equation}
except with conditional probability $2e^{-B\theta\eta}\le\alpha/(4K)$, by the last condition in \eqref{eq:anc-batch-a}. A union bound over the at most $K$ tested candidates costs $\alpha/4$. Thus \eqref{eq:anc-uniform} and \eqref{eq:anc-validation} hold together with probability at least $1-3\alpha/4$. Neither event refers to which validations failed, so reusing a failed validation batch as the next ownership batch is legitimate.

\emph{Coverage.} Work on this event. For the empirical minimizer $h_m$, \eqref{eq:anc-uniform} and \eqref{eq:anc-compensator} give, for every $h$, $S_m(h_m)\le S_m(h)\le(1+\theta)R(h)+\eta$; taking an infimizing sequence, $S_m(h_m)\le(1+\theta)R^*+\eta$, so $L_m\le R^*$. No population minimizer is needed. By \eqref{eq:anc-validation}, $U_m\ge R(h_m)$. Hence $L_m\le R^*\le R(h_m)\le U_m$. Separate clipping to $[0,1]$ preserves this chain and does not increase the width, so every accepted interval covers.

\emph{Acceptance within the cap.} Assume now $R^*\le r$ and write $C=C_m(h_m)$, $R=R(h_m)$. The lower half of \eqref{eq:anc-uniform} and the bound on $S_m(h_m)$ give $(1-\theta)C-\eta\le(1+\theta)r+\eta$, so
\[
 C\le\frac{(1+\theta)r+2\eta}{1-\theta}\le r+3\eta,
\]
where the last step is equivalent to $\theta(2r+3\eta)\le\eta$, which holds because $\theta=\eta/(8(r+\eta))$. Suppose $R-C\le\eta$. Then $R\le r+4\eta$, and from \eqref{eq:anc-uniform}, \eqref{eq:anc-validation},
\[
 C-L_m\le\frac{2\theta C+2\eta}{1+\theta},\qquad U_m-R\le\frac{2\theta R+2\eta}{1-\theta}.
\]
Since $\theta C\le\theta(r+3\eta)\le3\eta/8$ and $\theta R\le\theta(r+4\eta)\le\eta/2$,
\[
 U_m-L_m=(U_m-R)+(R-C)+(C-L_m)\le\frac{24\eta}7+\eta+\frac{11\eta}4=\frac{201\eta}{28}<8\eta<t ,
\]
so the candidate is accepted. Consequently, on the event above, a rejected candidate satisfies $R(h_m)-C_m(h_m)>\eta$.

For a nonempty finite list $G$ of functions put $\Psi(G)=\E_T\max_{g\in G}\tau_g$. Let $G_m=\{g_1,\ldots,g_m\}$. Since ranks are integers and $\ell\le1$, \eqref{eq:anc-compensator} gives for a rejected candidate
\begin{equation}\label{eq:anc-progress}
 \Psi(G_m\cup\{h_m\})-\Psi(G_m)\ge\Pr_T\bigl(\tau_{h_m}>\bar\tau_m\bigr)\ge R(h_m)-C_m(h_m)>\eta .
\end{equation}
Every list contains $a$. On a record on which $a$ survives, $\max_G\tau_g=M+1=\tau_a$, so
\begin{equation}\label{eq:anc-potential}
 0\le\Psi(G)-\Psi(\{a\})=\E_T\Bigl[\bigl(\max_{g\in G}\tau_g-\tau_a\bigr)\1\{\tau_a\le M\}\Bigr]\le D\,F(a)\le Dw .
\end{equation}
If all $K$ candidates were rejected, the potential would increase by more than $K\eta\ge Dw$, which is impossible. (If $w=0$, even one rejection is impossible.) Hence on the event above a candidate is accepted within the cap.

\emph{Accounting.} The first batch, every rejected candidate's batch and the accepted validation batch are actual resets and are all counted; there are at most $J$ batches and $J$ pilots. There are at most $K$ oracle calls. Every argument holds after conditioning on an earlier history fixing $a,r,w$, which gives the conditional statement.
\end{proof}

\subsection{The anchor with a supplied ceiling}

\begin{proof}[Proof of \cref{lem:anchor-supplied}]
Let $\epsb=\min(1,f+\eps)$. If $\epsb=1$, call the oracle on the empty list, return its member and $w=1$; no reset is used and $\eps\le w=1\le2(f+\eps)$.

Let $\epsb<1$, so $\epsb=f+\eps$. Put $I=\lceil\log_2(1/\epsb)\rceil$, $t_0=1$ and $t_i=\max(\epsb,2^{-i})$ for $1\le i\le I$. Then $t_I=\epsb$, $t_{i-1}\le2t_i$, and $\sum_{i=1}^It_i^{-1}\le3/\epsb$, because $t_i=2^{-i}$ for $i<I$ and $\sum_{i<I}2^i<2^I<2/\epsb$. Start with a member $a_0$ from an empty-list call and $U_0=1$. At stage $i=1,\ldots,I$, run $\textsc{Ownership}(a_{i-1},U_{i-1},U_{i-1},t_i,\alpha/I)$ for the natural loss. Let $a_i$ and $[L_i,\widetilde U_i]$ be its output and put
\[
 U_i=\min\bigl(\widetilde U_i,\ f+t_i\bigr)\in[0,1].
\]
Return $a=a_I$ and $w=\max(\eps,\min(1,f+\epsb))$.

\emph{Costs on every history.} At the first stage $U_0=1\le2t_1$. At a later stage, $U_{i-1}\le f+t_{i-1}\le\epsb+2t_i\le3t_i$, because $f\le\epsb\le t_i$. So the call at stage $i$ has inputs $r=w=U_{i-1}\le\bar u_i:=\min(1,3t_i)$. These inputs depend on the history, but $K\le\lceil D\bar u_i/\eta\rceil\le96D$, so the call has at most $96D+1$ batches and pilots. By the proof of \cref{lem:ownership}, its batch size is at most one more than the largest of four thresholds. These thresholds are nondecreasing in $r$ and in $w$: $\theta$ decreases in $r$, $K$ and $J$ increase in $w$, and $\nu(\chi)$, hence $H_\chi$, increases as $\chi=\eta/(32J)$ decreases. Evaluating them at $r=w=\bar u_i$ gives a bound on the batch size that depends only on $d$, $D$, $t_i$ and $\alpha/I$, and that is $\Ot((d+1)/t_i)$ by \eqref{eq:anc-batch-order}. Summing over the stages gives $\Ot((d+1)D/\epsb)$ resets and $O(DI)=\Ot(D)$ pilots and oracle calls, plus one initial call. These deterministic bounds hold on every history, including histories on which a call fails or $f$ is not a valid ceiling.

\emph{Validity.} Suppose $F^*\le f$. If the call at stage $i$ succeeds and its inputs are valid, both $F^*$ and $F(a_i)$ lie in $[L_i,\widetilde U_i]$, an interval of width at most $t_i$, so
\[
 F^*\le F(a_i)\le\widetilde U_i\le L_i+t_i\le F^*+t_i\le f+t_i ;
\]
in particular the clipping at $f+t_i$ is inactive, $F(a_i)\le U_i$ and $F^*\le U_i$, so the inputs of the next call are valid. The first call has valid inputs because $U_0=1$. Summing the conditional failure probabilities of the calls, each at most $\alpha/I$ given a history with valid inputs, all calls succeed with probability at least $1-\alpha$. On this event $F(a)\le U_I\le f+\epsb$ and $F(a)\le1$, so $F(a)\le w$. Finally $\eps\le2(f+\eps)$ and $f+\epsb=2f+\eps\le2(f+\eps)$, so $\eps\le w\le\min(1,2(f+\eps))$.
\end{proof}

\subsection{The anchor without a supplied ceiling}

\begin{proof}[Proof of \cref{lem:anchor-adaptive}]
If $\eps=1$, return an empty-list member and $w_0=1$. Let $\eps<1$, $I=\lceil\log_2(1/\eps)\rceil$ and $t_i=\max(\eps,2^{-i})$ for $1\le i\le I$. Start with an empty-list member $a_0$ and $U_0=1$. At stage $i$ run $\textsc{Ownership}(a_{i-1},U_{i-1},U_{i-1},t_i,\alpha/I)$ for the natural loss, and let $a_i$ and $[L_i,U_i]$ be its (clipped and totalized) output. Stop at the first $i$ with $t_i=\eps$ or $U_i>2t_i$; call it $c$, and return $a=a_c$ and $w_0=\max(\eps,U_c)\in[\eps,1]$.

\emph{Costs on every history.} The first call has bounds $1\le2t_1$. A later stage $i$ is reached only if stage $i-1$ did not stop, so $U_{i-1}\le2t_{i-1}\le4t_i$. As in the supplied case, the call at stage $i$ therefore has at most $128D+1$ batches and pilots, and its batch size is at most a number that depends only on $d$, $D$, $t_i$ and $\alpha/I$ and is $\Ot((d+1)/t_i)$. Since $\sum_{i\le c}t_i^{-1}\le3/t_c$ and $t_c\ge\eps$, the claimed caps follow. All numerical inputs stay in $[0,1]$ even after an unsuccessful call; the clipping and the proper fallback with $[0,0]$ are what make this all-history statement possible.

\emph{Validity.} As in the supplied case, with probability at least $1-\alpha$ all executed calls succeed with valid inputs, and then $F^*\le F(a_i)\le U_i\le F^*+t_i$ at each executed stage. If $t_c>\eps$, the stopping rule gives $U_c>2t_c$, so $F^*\ge U_c-t_c>t_c$ and $w_0=U_c\le F^*+t_c<2F^*$. If $t_c=\eps$, then $w_0\le\max(\eps,F^*+\eps)\le F^*+\eps$. In both cases $F(a)\le U_c\le w_0\le2(F^*+\eps)$.

\emph{Local reset count.} If $c=1$, then $t_c\ge1/2$ and $F^*+\eps\le2\le4t_c$. If $c>1$, success at stage $c-1$ and the fact that it did not stop give $F^*\le U_{c-1}\le2t_{c-1}\le4t_c$, and also $\eps\le t_c$. In all cases $t_c\ge(F^*+\eps)/5$ on the success event.
\end{proof}

\begin{remark}[Deployment and failure accounting]\label{rem:anchor-accounting}
An empty-list representative is obtained without being played. When the coarse schedule runs, its initial pilot and every candidate, including the final validated anchor, are played and counted. When the schedule is bypassed because $\epsb=1$, the anchor may be unplayed; its first use by a logger is then a counted training pilot. Enumerating a dictionary plays nothing by itself. The bounds above control the union of actual deployments by the total number of pilot slots over all calls. Coverage and the local bounds are used only while all earlier calls have succeeded; the numerical caps, properness and the ordered intervals of the stated widths hold on every history. No conditioning on the final adaptive list, no free exploration batch and no evaluation of a hidden terminal is used.
\end{remark}

\section{Proof of the acquisition lemma}\label{app:acquisition}

We prove \cref{lem:acquisition}. All probabilities are conditional on the history before the stage, which fixes $a,L,w,\zeta,b,\alpha$; assume $F(a)\le w$ for part~(b). The stopped-data content of the proof is the legal gain of \cref{lem:gain} and the uniform transfer through observed patterns. Given these, the rest is the standard conversion through a multiplicative potential, bounded martingale concentration and one randomized sparsification.

\paragraph{Part (a) and the sample sizes.}
The procedure plays a pilot drawn from $L$ on each of its $m+n$ resets, makes one oracle call on the $m$ dictionary displays, and draws $k$ slot identities with private coins; the oracle call and the draws play nothing. Since the dictionary displays contain at most $Dm$ queries, \cref{lem:vc-traces}(i) gives $K:=|\cG|\le K_0=(Dm+1)^d$. For part (c): since $2Dm+1\le3Dm$, the defining condition on $m$ holds if $(m\log2)/(2W_0)\ge\log(8/\alpha)+2d\log(3Dm)$, and by \cref{lem:inversion} (with $A=2W_0/\log2$, $B=\log(8/\alpha)$, $k=2d$, $c=3D$) this holds for every
\[
 m\ \ge\ \frac{4W_0}{\log2}\Bigl(\log\frac8\alpha+2d\log\frac{24dDW_0}{\log2}\Bigr)\qquad(d\ge1),
\]
and for every $m\ge(2W_0/\log2)\log(8/\alpha)$ if $d=0$. Together with $m\ge8W_0$ this shows $m=O(W_0[\log(e/\alpha)+d\log(e(d+1)DW_0)])$. Next $n=\lceil W\iota\rceil$ with $\iota=\log(4/\alpha)+d\log(Dm+1)$, and $W=2W_0/\Lam\le2W_0/\log3$. Hence $m+n=\Ot((d+1+\log(1/\alpha))W_0)$ with $W_0=1/(\zeta wb)$, where the hidden factor is polylogarithmic in $D$, $d+1$, $W_0$ and $1/\alpha$. The bound on $k$ is immediate.

\paragraph{The observed dictionary.}
For $h,h'\in\cH$ consider the event $O_{hh'}$ of \cref{lem:gain}(d) on one display of a pilot drawn from $L$. It depends on the pair only through the values of $h,h'$ at the at most $D$ displayed queries, together with data of the display, so by \cref{lem:vc-traces}(ii) these events realize at most $(2Dm+1)^{2d}$ indicator vectors on $2m$ displays. The $m$ dictionary displays are independent with a common law. By \cref{lem:vc-missed} with $\chi=1/W_0$ (note $m\chi\ge8$), except with probability $2(2Dm+1)^{2d}2^{-m/(2W_0)}\le\alpha/4$, every pair $(h,h')$ such that no dictionary display lies in $O_{hh'}$ satisfies $\Pr(O_{hh'})\le1/W_0$. This uniform statement is established before any representative is chosen.

Every $h\in\cH$ agrees with its representative $h'\in\cG$ at all displayed dictionary queries. Then no dictionary display lies in $O_{hh'}$, because $O_{hh'}$ requires exactly one of the two to agree with the label at some displayed query. Integrating \eqref{eq:observable-disagreement} over $T$ therefore gives, on the event above, simultaneously for all $h$,
\begin{equation}\label{eq:dictionary-transfer}
 \Pr_T\bigl(\rho_L(h)\ne\rho_L(h')\bigr)\le\frac{\Pr(O_{hh'})}\zeta\le\frac1{\zeta W_0}=wb .
\end{equation}
The representatives are algorithmic objects obtained from displayed queries; no latent net is used here.

\paragraph{The exponential-weights pass.}
From now on condition also on the dictionary, which is fixed during the pass. Let $\mathcal F_{t-1}$ contain the dictionary and the first $t-1$ rounds of the pass, so that $\pi_t$ is $\mathcal F_{t-1}$-measurable while the pilot and the record of round $t$ are fresh. Put
\[
 G_{t,h}=\E[\gamma_{t,h}\mid\mathcal F_{t-1}],\qquad \bar\gamma_t=\sum_{h\in\cG}\pi_t(h)\gamma_{t,h}.
\]
By \cref{lem:gain}(b),(c) and \cref{lem:rank-tail},
\[
 G_{t,h}=\frac{\E_T\Phi_{\pi_t}(T,\rho_L(h))}{w\Lam},\qquad
 0\le\gamma_{t,h}\le\frac2{\zeta bw\Lam}=W,\qquad
 \E[\bar\gamma_t\mid\mathcal F_{t-1}]\le\frac{F(a)\Lam}{w\Lam}\le1 .
\]
In the last bound the same numerical lottery $\pi_t$ appears as weight and in the potential, which is exactly the setting of \cref{lem:rank-tail}.

\emph{Pathwise bound.} Recall $K=|\cG|\le K_0$, and let $Z_t=\sum_{h\in\cG}K^{-1}\exp(\sum_{s<t}\gamma_{s,h}/W)$, so that $Z_1=1$ and $\pi_t(h)=K^{-1}\exp(\sum_{s<t}\gamma_{s,h}/W)/Z_t$. Since $e^x\le1+(e-1)x$ for $x\in[0,1]$,
\[
 \frac{Z_{t+1}}{Z_t}=\sum_{h\in\cG}\pi_t(h)e^{\gamma_{t,h}/W}\le1+(e-1)\frac{\bar\gamma_t}W\le\exp\Bigl((e-1)\frac{\bar\gamma_t}W\Bigr).
\]
As $K^{-1}\exp(\sum_t\gamma_{t,h}/W)\le Z_{n+1}$ for every $h\in\cG$, taking logarithms gives
\begin{equation}\label{eq:eg-pathwise}
 \sum_{t=1}^n\gamma_{t,h}\le W\log K+(e-1)\sum_{t=1}^n\bar\gamma_t\qquad(h\in\cG).
\end{equation}

\emph{Two supermartingales.} Apply \cref{lem:supermartingale}(a) with $\vartheta=1$ to the $[0,1]$-valued variables $\bar\gamma_t/W$ and $\gamma_{t,h}/W$:
\[
 \E\bigl[e^{\bar\gamma_t/W}\bigm|\mathcal F_{t-1}\bigr]\le\exp\Bigl((e-1)\frac{\E[\bar\gamma_t\mid\mathcal F_{t-1}]}W\Bigr),
 \qquad
 \E\bigl[e^{-\gamma_{t,h}/W}\bigm|\mathcal F_{t-1}\bigr]\le\exp\Bigl(-(1-e^{-1})\frac{G_{t,h}}W\Bigr).
\]
Multiplying over the rounds gives two nonnegative supermartingales starting at one. Markov's inequality at the final time, with threshold $e^\iota$, and a union over the $K\le K_0$ coordinates in the second case, show that outside an event of probability at most $e^{-\iota}+K_0e^{-\iota}\le\alpha/2$,
\[
 \sum_t\bar\gamma_t\le(e-1)\sum_t\E[\bar\gamma_t\mid\mathcal F_{t-1}]+W\iota\le(e-1)n+W\iota,
 \qquad
 (1-e^{-1})\sum_tG_{t,h}\le\sum_t\gamma_{t,h}+W\iota\quad(h\in\cG).
\]
Since $W\log K\le W\iota\le n$, combining with \eqref{eq:eg-pathwise} yields
\[
 (1-e^{-1})\sum_tG_{t,h}\le W\iota+(e-1)\bigl((e-1)n+W\iota\bigr)+W\iota\le\bigl(2+(e-1)e\bigr)n,
\]
that is, $n^{-1}\sum_tG_{t,h}\le\CEG$ for every $h\in\cG$, with $\CEG=(2+(e-1)e)/(1-e^{-1})=10.5530\ldots$.

\emph{Averaging.} For fixed $T$ and $l$, $\pi\mapsto\Phi_\pi(T,l)$ is convex, and neither the indicator $\1\{l>\tau_a\}$ nor the rank $\rho_L(h)$ depends on the numerical lottery. Jensen's inequality for the average $\bar\pi$ of the pre-update vectors gives, pointwise, $\Phi_{\bar\pi}(T,\rho_L(h))\le n^{-1}\sum_t\Phi_{\pi_t}(T,\rho_L(h))$, hence
\begin{equation}\label{eq:dense-certificate}
 \E_T\Phi_{\bar\pi}(T,\rho_L(h))\le\frac1n\sum_tw\Lam G_{t,h}\le\CEG\,w\Lam\qquad(h\in\cG).
\end{equation}

\paragraph{One sparsification.}
Condition on the whole pass, and let $p$ be the uniform lottery on $k$ independent draws from $\bar\pi$. Define the bad set
\[
 \Csp=\bigl\{T:\ \tau_a\le M\ \text{and}\ \exists\,l\le M+1\ \text{with}\ Q_{\bar\pi}(T,l)\ge b\ \text{and}\ Q_p(T,l)<Q_{\bar\pi}(T,l)/2\bigr\}.
\]
For a fixed record and rank, $Q_p(T,l)$ is the average of $k$ independent indicators with mean $Q_{\bar\pi}(T,l)$, so by the multiplicative Chernoff bound $\Pr(Q_p<Q_{\bar\pi}/2)\le e^{-kQ_{\bar\pi}/8}\le e^{-kb/8}$ when $Q_{\bar\pi}\ge b$. A union over the at most $D+1$ ranks of the record and integration over a fresh record with $\tau_a\le M$ give
\[
 \E_{\text{coins}}\Pr_T(\Csp)\le F(a)(D+1)e^{-kb/8}\le w(D+1)e^{-kb/8}\le\frac{\alpha wb}4
\]
by the choice of $k$. By Markov's inequality, $\Pr_T(\Csp)\le wb$ except with probability $\alpha/4$ over the coins. This bad set does not depend on the comparator.

On records with $\tau_a\le M$ outside $\Csp$ we have $Q_p(T,l)+b\ge\frac12(Q_{\bar\pi}(T,l)+b)$ for every $l$: if $Q_{\bar\pi}(T,l)\ge b$ this follows from $Q_p\ge Q_{\bar\pi}/2$, and otherwise from $b\ge(Q_{\bar\pi}+b)/2$. Hence $\Phi_p\le2\Phi_{\bar\pi}$ there. No such comparison is claimed on records on which the anchor survives, and none is needed, because there all potentials at ranks at most $M+1$ vanish. On $\Csp$ the sparse potential is at most $1/b$. With \eqref{eq:dense-certificate},
\[
 \E_T\Phi_p(T,\rho_L(h))\le2\CEG w\Lam+\frac{\Pr_T(\Csp)}b\le(2\CEG\Lam+1)\,w\qquad(h\in\cG).
\]

\paragraph{Transfer to the class.}
Let $h\in\cH$ and let $h'\in\cG$ be its representative. Both potentials lie in $[0,1/b]$ and coincide when the truncated ranks coincide, so
\[
 \bigl|\Phi_p(T,\rho_L(h))-\Phi_p(T,\rho_L(h'))\bigr|\le\frac1b\1\{\rho_L(h)\ne\rho_L(h')\},
\]
and \eqref{eq:dictionary-transfer} adds at most $w$. This proves \eqref{eq:truncated-certificate} with $C(b)=2\CEG\Lam+2$, simultaneously for all $h\in\cH$. The three failure probabilities add up to $\alpha/4+\alpha/2+\alpha/4=\alpha$. The transfer is paid once, after the sparsification. No function is played because of its numerical weight or because it was enumerated; only pilots drawn from the frozen logger are played during the procedure.

\section{A sparse graph with the required path property}\label{app:graph}

We prove \cref{lem:graph-interface}. The argument is a reconstruction of the sparse-connector, good-vertex and joining method of Erd\H{o}s, Graham and Szemer\'edi \citep{EGS1975_egs1975_journal_static_renyi}, written in the local-expander language of Alwen, Blocki and Pietrzak \citep[Section~4]{ABP2017_2018_abp2017_arxiv_v2}. Those graph lemmas were already public in the May 2017 version of the preprint; the cited July 2017 version supplies the locators. The proof below fixes constants and boundary cases and is not a new graph-theoretic result.

Let $m_0\ge5$ be an integer, $\xi=1/m_0$ and $\gamma\in[5\xi,1)$. A directed graph on $[S]$ with edges $(x,y)$, $x<y$, is a \emph{$\xi$-local expander} if it contains every edge $(x,y)$ with $1\le y-x\le2m_0-1$, and, for every two adjacent intervals
\[
 I=[v,v+r-1],\qquad J=[v+r,v+2r-1]\subseteq[S],
\]
all $X\subseteq I$ and $Y\subseteq J$ with $|X|,|Y|\ge\xi r$ are joined by an edge from $X$ to $Y$. For $r\le m_0$ the short edges already give this, since then $X$ and $Y$ are nonempty and every $x\in X$, $y\in Y$ satisfy $1\le y-x\le2r-1$.

\begin{lemma}[Random local expansion]\label{lem:graph-construction}
Include all edges $(x,y)$ with $1\le y-x\le2m_0-1$. For each $y\in\{2,\ldots,S\}$ and each integer $i$ with $m_0\le2^i\le2S$, independently draw
\[
 k_G=\Bigl\lceil4m_0\Bigl(2\log(em_0)+\log\frac{2S^2}\beta\Bigr)\Bigr\rceil
\]
uniform points from the window $[\max(1,y-2^{i+1}),\,y-1]$, which is nonempty because $y\ge2$, and add the edges from these points to $y$. With probability at least $1-\beta$ the graph is a $\xi$-local expander, and on every draw every indegree is at most $\min\{S-1,\ 2m_0-1+k_G(\lfloor\log_2(2S)\rfloor+1)\}$.
\end{lemma}

\begin{proof}
The indegree bound is immediate: there are at most $2m_0-1$ short edges into $y$, at most $\lfloor\log_2(2S)\rfloor+1$ scales, at most $k_G$ draws per scale (duplicates only reduce the count), and at most $y-1\le S-1$ possible predecessors. Vertex $1$ receives no random edge and has no predecessors.

Fix $r>m_0$ and adjacent intervals $I,J$ as above, and put $s=\lceil\xi r\rceil\ge2$. It suffices to consider $|X|=|Y|=s$. Let $i=\lceil\log_2r\rceil$, so that $m_0<r\le2^i<2r\le S$; hence scale $i$ is drawn. Every $y\in Y$ satisfies $y\ge v+r\ge2$, so $y$ receives random edges; and $y\le v+2r-1$ with $2^{i+1}\ge2r$ gives $y-2^{i+1}<v$, so the window of $y$ at scale $i$ contains $I\supseteq X$ and has fewer than $4r$ points. One draw hits $X$ with probability at least $s/(4r)\ge\xi/4$. The draws are independent over $y\in Y$ and over repetitions, so
\[
 \Pr(\text{no edge from }X\text{ to }Y)\le(1-\xi/4)^{k_Gs}\le e^{-k_Gs/(4m_0)} .
\]
There are at most $\binom rs^2\le(er/s)^{2s}\le(em_0)^{2s}$ choices of $(X,Y)$, so the failure probability for this pair of intervals is at most
\[
 \exp\Bigl[-s\Bigl(\frac{k_G}{4m_0}-2\log(em_0)\Bigr)\Bigr]\le\exp\Bigl[-s\log\frac{2S^2}\beta\Bigr]\le\frac\beta{2S^2}.
\]
A union bound over at most $S^2/2$ pairs of adjacent intervals gives failure at most $\beta/4<\beta$. The cases $r\le m_0$ are covered by the short edges.
\end{proof}

Given a deletion set $B\subseteq[S]$, call $x\in[S]$ \emph{$\gamma$-good} if every interval in $[S]$ with endpoint $x$ has at most a $(1-\gamma)$-fraction of deleted vertices:
\[
 |B\cap[x,x+\varrho-1]|\le(1-\gamma)\varrho\quad\text{and}\quad|B\cap[x-\varrho+1,x]|\le(1-\gamma)\varrho
\]
for every $\varrho\ge1$ for which the interval lies in $[S]$. Taking $\varrho=1$ shows that a good vertex is not deleted.

\begin{lemma}[Few bad vertices]\label{lem:graph-count}
At most $2|B|/(1-\gamma)$ vertices are not $\gamma$-good.
\end{lemma}

\begin{proof}
If $B$ is empty, every vertex is good. Otherwise call $x$ forward-bad if some interval $[x,x+\varrho-1]\subseteq[S]$ has more than $(1-\gamma)\varrho$ deleted vertices. Choose the least forward-bad vertex and such a witness interval; then the least forward-bad vertex beyond this interval and a witness interval; and so on. The chosen intervals are disjoint and cover all forward-bad vertices, and each of them contains more than a $(1-\gamma)$-fraction of deleted vertices, so their total length is less than $|B|/(1-\gamma)$. Backward-bad vertices are counted in the same way with the order reversed. The sum of the two bounds proves the lemma.
\end{proof}

\begin{lemma}[Joining good vertices]\label{lem:graph-joining}
In a $\xi$-local expander with $\gamma\ge5\xi$, any two $\gamma$-good vertices $x<y$ are joined by a directed path whose vertices lie in $[x,y]\setminus B$.
\end{lemma}

\begin{proof}
Let $n=y-x+1$. If $y-x\le2m_0-1$, the short edge $(x,y)$ works, as $x,y\notin B$. Otherwise $n>2m_0$. Let $R_f$ be the set of vertices of $[x,y]\setminus B$ reachable from $x$ by a path inside $[x,y]\setminus B$, and $R_b$ the set of vertices of $[x,y]\setminus B$ from which $y$ is reachable inside $[x,y]\setminus B$.

\emph{Expansion step.} If $2\varrho\le n$ and $|R_f\cap[x,x+\varrho-1]|\ge\xi\varrho$, then fewer than $\xi\varrho$ undeleted vertices of $[x+\varrho,x+2\varrho-1]$ lie outside $R_f$. Otherwise local expansion would give an edge from a vertex of $R_f$ to an undeleted vertex of $[x+\varrho,x+2\varrho-1]$ outside $R_f$, and that vertex would be in $R_f$.

\emph{Doubling.} We show by induction that for every $i\ge0$ with $2^i\le n$, fewer than $\xi2^i$ undeleted vertices of $[x,x+2^i-1]$ lie outside $R_f$, and $|R_f\cap[x,x+2^i-1]|\ge(\gamma-\xi)2^i$. For $i=0$ the interval is $\{x\}$ and $x\in R_f$. If the claim holds for $i$ and $2^{i+1}\le n$, then $|R_f\cap[x,x+2^i-1]|\ge(\gamma-\xi)2^i\ge\xi2^i$ because $\gamma\ge2\xi$, so the expansion step with $\varrho=2^i$ shows that fewer than $\xi2^i$ undeleted vertices of the second half lie outside $R_f$; adding the two halves gives fewer than $\xi2^{i+1}$. Goodness of $x$ gives at least $\gamma2^{i+1}$ undeleted vertices in $[x,x+2^{i+1}-1]$, so at least $(\gamma-\xi)2^{i+1}$ of them are in $R_f$. The same argument with the order reversed, using backward goodness of $y$, gives $|R_b\cap[y-2^i+1,y]|\ge(\gamma-\xi)2^i$ for $2^i\le n$.

\emph{Connection.} Put $r=\lfloor n/2\rfloor\ge m_0$ and let $p$ be the largest power of two with $p\le r$, so $p>r/2$. Let $I_1=[x,x+r-1]$ and $I_2=[x+r,x+2r-1]\subseteq[x,y]$. Then $|R_f\cap I_1|\ge(\gamma-\xi)p>(\gamma-\xi)r/2\ge\xi r$, since $\gamma\ge3\xi$. Moreover $[y-p+1,y-1]\subseteq I_2$, because $r+p\le2r\le n$ and $2r\ge n-1$. Hence
\[
 |R_b\cap I_2|\ge(\gamma-\xi)p-1>\frac{(\gamma-\xi)r}2-1\ge\xi r,
\]
where the last step uses $r\ge m_0=1/\xi$ and $\gamma\ge5\xi$. Local expansion gives an edge from $R_f\cap I_1$ to $R_b\cap I_2$. Concatenating a path from $x$ to its tail, the edge, and a path from its head to $y$ gives the required path.
\end{proof}

\begin{proof}[Proof of \cref{lem:graph-interface}]
Use \cref{lem:graph-construction} with $m_0=10$, so that $\xi=1/10$, $2m_0-1=19$ and $k_G$ is as in \eqref{eq:degree-cap}, and take $\gamma=1/2=5\xi$. The short edge $(s-1,s)$ gives $s-1\in A_s$ for $s\ge2$, and the indegree bound is $k_A$. On the local-expansion event, fix any $B\subseteq[S]$ and list the good vertices in increasing order. By \cref{lem:graph-count} there are at least $S-4|B|$ of them. Join consecutive good vertices by \cref{lem:graph-joining}; the joining paths lie in intervals with disjoint interiors, so their concatenation is a directed path avoiding $B$ through all good vertices, with at least $S-4|B|$ vertices. A single good vertex is a path with one vertex, and if the bound $S-4|B|$ is nonpositive there is nothing to prove. Local expansion holds simultaneously for all pairs of intervals, so the conclusion holds for every deletion set at once, including sets that depend on the graph and on later data.
\end{proof}

\begin{remark}[Variants of the stage graph]\label{rem:graph-variants}
(i) Since \cref{lem:graph-construction} shows that a suitable graph exists, a deterministic exhaustive search over graphs with the same short edges and indegree cap also finds one; testing local expansion is a finite computation and uses no reset.

(ii) Blocki, Cinkoske, Lee and Son \citep[Section~4, Theorem~9]{BCLS2021_2022_blocki2022_explicit} give explicit graphs $G(\delta,N)$ on $N=2^n$ vertices with maximum indegree $O(\log N)$ that are $\delta$-local expanders in the sense that every two adjacent intervals of equal length with subsets of size at least $\delta$ times the length are joined by an edge; they credit Theorem~9 to the argument of Erd\H{o}s, Graham and Szemer\'edi applied to their explicit bipartite expanders. Take the prefix $[S]$ of their graph on the next power of two, with $\delta=1/10$, and add every edge of length at most $19$. Every test of adjacent intervals inside $[S]$ keeps its edge, and the added edges give the short-edge condition. \Cref{lem:graph-count,lem:graph-joining} then give the path property on every draw, with indegree $O(\log S)$ and no failure allowance. This uses the published graph property, not an implementation of the pseudocode. The construction was public in 2021 and published at STACS in 2022.

(iii) The complete graph on $[S]$, with $k_A=S-1$, is a stage graph with the path property on every draw (\cref{sec:composition}). Running \cref{alg:main} with the complete graph instead of the random one gives a learner whose resets and distinct pilots are at most $\Ot((d+1)[D^2(f+\eps)+r]/\eps^2)$ and $\Ot(D(1+f/\eps))$ on every history, and which covers with probability at least $1-\delta$ when $F^*\le f$ and $R^*\le r$. Indeed \cref{prop:precision} applies with $k_A=4D-1$ and no graph event, so \eqref{eq:stage-cost} has $Dk_Aw=O(D^2(f+\eps))$, while the pilot bound in \cref{prop:precision}(a) does not depend on $k_A$; the failure probability is at most $\delta/4+2\delta/16<\delta$. We do not use this weaker variant in the main theorem.
\end{remark}

\section{Uniform relative estimation and the joint interval}\label{app:estimation}

This appendix proves \cref{lem:estimate-interval}. We first state a generic estimate for a frozen lottery, then bound its sample size, then construct the interval. All randomness before the frozen lottery is conditioned on. The latent-record nets and rare counts of \cref{app:vc} are used only in the analysis.

\begin{lemma}[Frozen-lottery relative estimate]\label{lem:est-relative}
Condition on a history that fixes a lottery $\Pi$ and a measurable function $\psi\ge0$, the scoring rule. In each of $n$ rounds a pilot $g_i$ is played on a fresh record $T_i$, and, conditionally on the history, the pairs $(g_1,T_1),\ldots,(g_n,T_n)$ are independent, with $g_i\sim\Pi$ and $T_i\sim\cP$ independent of $g_i$. The score of $h\in\cH$ in round $i$ is $Y_{h,i}=\psi(g_i,T_i,\tau_h(T_i))$. For a generic pair $(g,T)$ with this law write $Y_h=\psi(g,T,\tau_h(T))$, and assume:
\begin{enumerate}[label=(\roman*)]
\item $Y_h$ is a function of the display of $g$ on $T$ and of the values of $h$ at the displayed queries;
\item $0\le Y_h\le W_Y$ for a number $W_Y\ge1$;
\item $\E[Y_h\mid T]\in[0,1]$ and $R(h)-\beta_b\le\E Y_h\le R(h)$ for a number $\beta_b\ge0$;
\item $\E Y_h^2\le c_1R(h)+c_2$ for numbers $c_1>0$, $c_2\ge0$.
\end{enumerate}
Let $v\in[0,1]$, $0<t\le1$ and $0<\beta<1$, define $s_E$ and $\theta$ by \eqref{eq:est-s} and $\nu_n,H_n,\iota_n,\kappa_n$ by \eqref{eq:net-parameters}, and suppose $n\ge2$ satisfies the two conditions of \eqref{eq:est-n}. Let $\widehat R(h)$ be the average of the $n$ scores of $h$. Then, with probability at least $1-\beta$,
\begin{equation}\label{eq:estimator-generic}
 |\widehat R(h)-R(h)|\le\theta R(h)+\frac{3t}{16}+\beta_b\qquad\text{for every }h\in\cH .
\end{equation}
\end{lemma}

\begin{proof}
Let $\cN=\cN_{1/n}$ be the net of \cref{lem:vc-net} with mass $1/n$; it has at most $H_n$ members, and every $h$ has a neighbor $h'\in\cN$ with $\cP(E(h,h'))\le1/n$. No member of the net is given to the learner.

\emph{Concentration on the net.} For a fixed $h'\in\cN$ the $n$ scores $Y_{h',i}=\psi(g_i,T_i,\tau_{h'}(T_i))$ are values of one fixed function at the independent pairs $(g_i,T_i)$, so they are independent and identically distributed. Since $0\le s_EY\le1$ and $e^x\le1+x+x^2$ for $x\le1$, for either sign
\[
 \E e^{\pm s_E(Y-\E Y)}\le e^{\mp s_E\E Y}\bigl(1\pm s_E\E Y+s_E^2\E Y^2\bigr)\le e^{s_E^2\E Y^2}.
\]
Hence, by the exponential Markov inequality, for every $x>0$,
\[
 \Pr\Bigl(\pm\bigl(\widehat R(h')-\E Y_{h'}\bigr)\ge s_E\E Y_{h'}^2+\frac x{s_En}\Bigr)\le e^{-x}.
\]
A union bound over both signs and over $\cN$, with $x=\iota_n=\log(4H_n/\beta)$, costs at most $\beta/2$.

\emph{Rare disagreements.} By \cref{lem:vc-rare}, outside an event of probability at most $\beta/2$, every event $E(h,h')$ of mass at most $1/n$ contains at most $\kappa_n$ of the $n$ latent records. The two exceptional events may be dependent; the union bound suffices.

\emph{Transfer.} Fix $h$ and its neighbor $h'$. Outside $E(h,h')$ the two functions agree at every query of the record, so $\tau_h(T)=\tau_{h'}(T)$ and their scores agree for every pilot draw. Hence
\[
 |\widehat R(h)-\widehat R(h')|\le\frac{W_Y\kappa_n}n,\qquad
 |\E Y_h-\E Y_{h'}|\le\frac1n,\qquad
 R(h')\le R(h)+\frac1n ,
\]
where the middle bound uses that the conditional means lie in $[0,1]$ and agree outside $E(h,h')$, and the last uses $\ell\in[0,1]$. Combining these with the concentration bound and with (iv),
\[
 |\widehat R(h)-\E Y_h|\le s_Ec_1R(h)+s_Ec_2+\frac{\iota_n}{s_En}+\frac{W_Y\kappa_n+1+s_Ec_1}n .
\]
Now $s_Ec_2\le t/16$, $\iota_n/(s_En)\le t/16$ by \eqref{eq:est-n}, and $s_Ec_1\le t/(16(v+t))\le1/16$, so the last term is at most $(W_Y\kappa_n+2)/n\le t/16$. Adding the bias bound from (iii) gives \eqref{eq:estimator-generic}.
\end{proof}

\begin{lemma}[Sample size of the final estimate]\label{lem:est-order}
In \cref{lem:est-relative}, the least integer $n\ge2$ satisfying \eqref{eq:est-n} exists, and it is at most the largest of $2$ and the two numbers
\[
 \frac{32}{s_Et}\Bigl(\log\frac4\beta+2d\log\frac{3072\,d^2D}{s_Et}\Bigr),
 \qquad
 \frac{32}t\Bigl(W_Y\Bigl(4+18\log\frac4\beta\Bigr)+2+36W_Yd\log\frac{3456\,dDW_Y}t\Bigr)
\]
(for $d=0$ the logarithmic terms with factor $d$ are absent), plus one. Consequently
\[
 n_F=\Ot\!\left(\bigl(d+1+\log(1/\beta)\bigr)\Bigl[\frac{W_Y}t+\frac{c_1(v+t)}{t^2}+\frac{c_2+t}{t^2}\Bigr]\right),
\]
where the hidden factor is polylogarithmic in $d+1$, $D$, $W_Y$, $1/s_E$ and $1/t$.
\end{lemma}

\begin{proof}
Let $d\ge1$. By \cref{lem:vc-net} with $\chi=1/n$, and since $\log4\le d\log(48dDn)$, we have $\nu_n\le24dn\log(48dDn)+1$. With $x=48dDn$ this gives $\nu_nD+1\le(x/2)\log x+D+1\le x^2$, so $\log(\nu_nD+1)\le2\log x$ and $\iota_n\le\log(4/\beta)+2d\log(48dDn)$. The first condition of \eqref{eq:est-n} therefore holds if $n\ge(16/(s_Et))(\log(4/\beta)+2d\log(48dDn))$, and by \cref{lem:inversion} (with $A=16/(s_Et)$, $k=2d$, $c=48dD$) this holds for every $n$ above the first displayed number. Next, $\kappa_n\le4+18\log(4/\beta)+36d\log(3Dn)$ because $2nD+1\le3Dn$, so the second condition holds if $n\ge(16/t)(W_Y(4+18\log(4/\beta))+2+36W_Yd\log(3Dn))$, and \cref{lem:inversion} (with $A=16/t$, $k=36W_Yd$, $c=3D$) shows that this holds above the second displayed number. For $d=0$ we have $H_n=1$, $\iota_n=\log(4/\beta)$ and $\kappa_n=\max\{4,18\log(4/\beta)\}$, and the conditions are explicit. Every integer above both thresholds satisfies \eqref{eq:est-n}, which proves the first claim. For the order, use $1/s_E\le W_Y+16c_1(v+t)/t+16(c_2+t)/t$.
\end{proof}

\begin{lemma}[Proper minimization and the totalized interval]\label{lem:est-interval}
Suppose that \eqref{eq:estimator-generic} holds with $\beta_b\le t/16$, so that $|\widehat R(h)-R(h)|\le\theta R(h)+t/4$ for all $h$. Enumerate the patterns of $\cH$ on all queries displayed in the $n$ rounds, let $\widehat h$ be a returned member minimizing $\widehat R$, let $q=\widehat R(\widehat h)$, and put
\[
 l=\frac{q-t/4}{1+\theta},\qquad u=\frac{q+t/4}{1-\theta}.
\]
Then $l\le R^*\le R(\widehat h)\le u$. If moreover $R^*\le v$, then $u-l\le2t/3$. Clip $l$ and $u$ separately to $[0,1]$ and replace the result by $[0,0]$ if it is unordered or wider than $t$. On every history the result is an ordered interval in $[0,1]$ of width at most $t$ and $\widehat h\in\cH$; under the two conditions above it contains $R^*$ and $R(\widehat h)$.
\end{lemma}

\begin{proof}
The empirical score of $h$ depends only on its values at the displayed queries, so its minimum over the returned representatives is its minimum over $\cH$. For every $h$, $q\le\widehat R(h)\le(1+\theta)R(h)+t/4$, hence $l\le R(h)$ for every $h$ and so $l\le R^*$. Taking an infimizing sequence also gives $q\le(1+\theta)R^*+t/4$; no population minimizer is needed. The lower half of the estimate at $\widehat h$ gives $q\ge(1-\theta)R(\widehat h)-t/4$, so $R(\widehat h)\le u$. Since $R^*\le R(\widehat h)$, both numbers lie in $[l,u]$.

Now let $R^*\le v$. Since $\theta=c_1s_E\le t/(16(v+t))\le1/16$ by \eqref{eq:est-s},
\[
 u-l=\frac{2\theta q+t/2}{1-\theta^2}
 \le\frac{2\theta(1+\theta)v+\theta t/2+t/2}{1-\theta^2}
 \le\frac{(17/128+4/128+64/128)\,t}{255/256}=\frac{2t}3,
\]
where we used $2\theta(1+\theta)v\le(17/8)\,tv/(16(v+t))\le17t/128$ and $\theta t/2\le t/32$. Clipping preserves membership of numbers in $[0,1]$ and does not increase the width of an ordered pair, so on this event the clipped pair is ordered, has width at most $2t/3$, contains both numbers, and the fallback is not used. On every history the fallback guarantees an ordered interval of width at most $t$; off the event no coverage is claimed. Every selected representative and the fallback output are members of $\cH$.
\end{proof}

\begin{proof}[Proof of \cref{lem:estimate-interval}]
\Cref{alg:precision} plays $n_F$ pilots drawn independently from $\Pi$ on fresh resets, so given the history before the final estimate the pairs $(g_i,T_i)$ are independent with $g_i\sim\Pi$ and $T_i\sim\cP$, and the scoring rule \eqref{eq:final-score} is fixed by that history. The scores satisfy (i)--(iv) of \cref{lem:est-relative} with $W_Y=2/b\ge1$, $c_1=2$, $c_2=2C_*w$ and $\beta_b=64bC(b)w\le t/16$, by \cref{lem:final-score}, when \eqref{eq:final-certificate} holds. \Cref{lem:est-relative} then gives \eqref{eq:relative-estimate} with probability at least $1-\beta$, since $3t/16+\beta_b\le t/4$, and \cref{lem:est-interval} gives the remaining claims. The sample size $n_F$ of \cref{alg:precision} is the one of \cref{lem:est-order}. No trace bound in this argument uses more than $D$ queries per latent record, so the argument is valid for every horizon.
\end{proof}

\section{Unknown ceilings, expected costs and accounting}\label{app:adaptation}

This appendix proves \cref{thm:adaptive} and \cref{prop:auxiliary}.

\Cref{alg:adaptive} reuses one anchor and one graph at all precisions, and uses fresh dictionaries, fresh numerical passes, fresh sparse slots and fresh final-estimation data at every precision. The scales satisfy $t_I=\eps$, $t_i=2^{-i}$ for $i<I$, $t_{i-1}\le2t_i$ (with $t_0=1$), and
\begin{equation}\label{eq:scale-sums}
 \sum_{i=1}^I\frac1{t_i}\le\frac3\eps,\qquad\sum_{i=1}^I\frac1{t_i^2}\le\frac3{\eps^2},
\end{equation}
because $2^{I-1}<1/\eps$ gives $\sum_{i<I}2^i<2/\eps$ and $\sum_{i<I}4^i<4^I/3<4/(3\eps^2)$. On every history $t_i\le w_i\le1$ and $0\le v_i\le1$, since $w_0\in[\eps,1]$ and the endpoints of $\widehat I_{i-1}$ lie in $[0,1]$.

\begin{algorithm}[t]
\caption{Learner of \cref{thm:adaptive} (no supplied ceilings)}\label{alg:adaptive}
\begin{algorithmic}[1]
\Require $d,D,\eps,\delta$
\If{$\eps=1$} \Return an empty-list representative and $[0,1]$ \EndIf
\State $I\gets\lceil\log_2(1/\eps)\rceil$; \ $t_i\gets\max(\eps,2^{-i})$ for $1\le i\le I$; \ $\beta\gets\delta/(32(I+1))$
\State run the procedure of \cref{lem:anchor-adaptive} with failure $\alpha=\delta/4$; obtain $a$ and $w_0$
\State draw one graph $\Gamma$ by \cref{lem:graph-interface} on $[4D]$ with allowance $\delta/8$
\For{$i=1,\ldots,I$}
 \State $w_i\gets\max(t_i,w_0)$; \ $v_i\gets1$ if $i=1$, and $v_i\gets$ the upper endpoint of $\widehat I_{i-1}$ if $i>1$
 \State $(\widehat h_i,\widehat I_i)\gets\textsc{PrecisionStage}(a,t_i,w_i,v_i,\beta,\Gamma)$ \Comment{fresh data at every stage}
\EndFor
\State \Return $(\widehat h_I,\widehat I_I)$
\end{algorithmic}
\end{algorithm}

\begin{proof}[Proof of \cref{thm:adaptive}]
If $\eps=1$ the claims are trivial, so let $\eps<1$.

\emph{(a) Every history.} Each output of \cref{alg:precision} is a member of $\cH$ with an ordered interval in $[0,1]$ of width at most its precision (\cref{prop:precision}(a)); the last precision is $\eps$. The graph is drawn with allowance $\delta/8$, so its cap is $k_A=O((1+\log D)\log(eD/\delta))$. Replacing $w_i$ and $v_i$ by their bound $1$ in \eqref{eq:stage-cost}, stage $i$ uses $\Ot((d+1)(Dk_A+3)/t_i^2)$ resets and plays at most $\Ot(D/t_i)$ functions. By \eqref{eq:scale-sums} the sums are $\Ot((d+1)D/\eps^2)$ and $\Ot(D/\eps)$. The anchor adds $\Ot((d+1)D/\eps)$ resets and $\Ot(D)$ pilots by \cref{lem:anchor-adaptive}. Every function played in a stage is the anchor or one of its slots, so the union of the functions played over all stages is at most one plus the total number of slots; these are the global caps. They are not instancewise bounds.

\emph{(b) The success event.} Let $E_0$ be the event that the anchor procedure succeeds, $E_\Gamma$ the event that $\Gamma$ has the path property, and $E_i$ the event that all designs and the final estimate of stage $i$ satisfy their guarantees. Then $\Pr(E_0^c)\le\delta/4$ and $\Pr(E_\Gamma^c)\le\delta/8$. On $E_0$ we have $F(a)\le w_0\le2(F^*+\eps)$, hence $F(a)\le w_i$ and
\[
 w_i=\max(t_i,w_0)\le2F^*+3t_i .
\]
We claim that on $E_0\cap E_\Gamma\cap E_1\cap\cdots\cap E_{i-1}$ we have $R^*\le v_i\le R^*+2t_i$. For $i=1$ this holds because $t_1\ge1/2$ and $v_1=1$. For $i>1$, stage $i-1$ had valid inputs and succeeded, so $\widehat I_{i-1}$ contains $R^*$ and has width at most $t_{i-1}\le2t_i$; its upper endpoint $v_i$ therefore lies in $[R^*,R^*+2t_i]$. Thus on this event stage $i$ has valid inputs, and \cref{prop:precision}(b), applied conditionally on the history before stage $i$ and on the realized graph, gives $\Pr(E_i^c\cap E_0\cap E_\Gamma\cap E_1\cap\cdots\cap E_{i-1})\le2\beta$. No fresh graph event is charged at later precisions; the single graph is charged once. The total failure probability is at most
\[
 \frac\delta4+\frac\delta8+2I\beta<\frac\delta4+\frac\delta8+\frac\delta{16}<\delta .
\]
On the joint event the last interval covers. Moreover \eqref{eq:stage-cost} with $w_i\le2F^*+3t_i$ and $v_i\le R^*+2t_i$ gives
\[
 N_i=\Ot\!\left((d+1)\,\frac{D(F^*+t_i)+R^*+t_i}{t_i^2}\right),\qquad
 V_i=\Ot\!\left(D\,\frac{F^*+t_i}{t_i}\right)
\]
for the resets and new pilots of stage $i$. Summing with \eqref{eq:scale-sums} gives $\Ot((d+1)[D(F^*+\eps)+R^*]/\eps^2)$ resets, since $D\sum_it_i^{-1}\le3D/\eps\le3D\eps/\eps^2$, and $\Ot(D(F^*/\eps+I))=\Ot(D(1+F^*/\eps))$ pilots, since $I=O(\log(e/\eps))$. On $E_0$ the anchor uses $\Ot((d+1)D/(F^*+\eps))\le\Ot((d+1)D(F^*+\eps)/\eps^2)$ resets and $\Ot(D)$ pilots. All new sparse slots and all anchor pilots are counted; reuse can only make the union smaller.

\emph{(c) Expected costs.} Run the same learner with internal failure $\delta'=\min(\delta,\eps^2)$, recomputing all allocations and logarithms; then (a) and (b) hold with $\delta'$, and $\log(1/\delta')\le\log(1/\delta)+2\log(1/\eps)$ is still polylogarithmic. On the success event the costs obey the local bounds of (b); on its complement, which has probability at most $\delta'$, they obey the global caps of (a). Hence
\[
 \E N\le\Ot\!\left((d+1)\frac{D(F^*+\eps)+R^*}{\eps^2}\right)+\eps^2\,\Ot\!\left(\frac{(d+1)D}{\eps^2}\right),
 \qquad
 \E V\le\Ot\!\left(D\Bigl(1+\frac{F^*}\eps\Bigr)\right)+\eps^2\,\Ot\!\left(\frac D\eps\right).
\]
The second terms are $\Ot((d+1)D)$ and $\Ot(D\eps)$, and they are absorbed because $D\le D(F^*+\eps)/\eps^2$ and $D\eps\le D(1+F^*/\eps)$. This is an expectation argument based on the global caps, not a relabelling of the high-probability statement.
\end{proof}

\begin{proof}[Proof of \cref{prop:auxiliary}]
(i) Every reset displays at most $M\le D$ pairs.

(ii) For \cref{alg:main}: the anchor procedure of \cref{lem:anchor-supplied} makes $O(D\log(e/\eps))$ ownership minimizations and one initial call, and the precision stage makes $S=4D$ design calls and one selection call. For \cref{alg:adaptive}: the anchor procedure of \cref{lem:anchor-adaptive} makes $\Ot(D)$ calls including initialization, and the $I$ precision stages make $I(4D+1)=O(D\log(e/\eps))$ calls. In both cases the total is $\Ot(D(1+\log(1/\eps)))$; the term $1$ covers the endpoint $\eps=1$, where exactly one empty-list call is made although no reset is used.

(iii) In design $s$ the oracle is called on the queries of $m_s$ displays, at most $Dm_s$ query occurrences, so by \cref{lem:vc-traces}(i) it returns at most $(Dm_s+1)^d$ representatives. The final selection call sees at most $Dn_F$ queries and returns at most $(Dn_F+1)^d$ representatives. These counts do not bound the memory used by the anchor procedures, the running time of the oracle, or the bit complexity of exact arithmetic.

(iv) This is the last part of \cref{prop:precision}(a) and its proof: every pilot of a stage is the anchor or a slot, and enumeration plays nothing by itself; every actual use of a function, in whatever role, is counted in $V$.

(v) This is the first line of \cref{alg:main,alg:adaptive}: width $1\le\eps$, and $[0,1]$ contains $R^*$ and $R(\widehat h)$.

For $d=0$ all trace counts equal one and every argument remains valid with the factor $d+1$; no lower-bound sharpness at $d=0$ is claimed.
\end{proof}

\section{Proofs of the lower bounds}\label{app:lower}

This appendix proves \cref{thm:lower,thm:natural-lower}. Throughout, $d$ and $D$ are positive integers and the ceilings $f$ and $r$ lie in $[0,1]$; individual lemmas add further cutoffs. Pilots, displays and outputs are as in \cref{sec:model}; in particular the output belongs to the class but need not have been played. $N$ is the number of resets and $V$ the number of distinct functions actually played as pilots: computing a mixture, enumerating functions or returning the output plays nothing by itself, and every function actually used as a pilot is counted. For the bounds on $\E V$ we assume, as stated in the theorems, that runs terminate almost surely on every instance of the finite families below; this holds in particular for every learner with deterministic caps.

The hard families below are finite: their classes, query sets and sets of worlds are finite. (The proof of \cref{thm:natural-lower}(c) at the end of the appendix is an upper bound and holds for every class satisfying \cref{ass:oracle,ass:measurable}.) Each family uses one known loss, fixed before learning, and its hidden parameters are carried by the truth function and by the record law. In every family an empty record has a fixed constant descriptor. We write $\E^0$ and $\Pr^0$ for expectations and probabilities in a comparison experiment.

\subsection{Threshold paths and two information tools}

The fresh-bit calculation for threshold paths is standard; related symmetric-counterexample constructions appear in \citet[Example~2]{braverman2026}. A \emph{block of depth} $L\ge1$ has ordered queries $1,\ldots,2^L-1$ and the threshold functions $h_\varsigma(j)=\1\{j\ge\varsigma\}$, $1\le\varsigma\le2^L$. A hidden leaf is written
\[
 \mathfrak b=1+\sum_{s=1}^LB_s2^{L-s},\qquad B_s\in\{0,1\},
\]
and the truthful labels are those of $h_{\mathfrak b}$. At level $s$ put
\[
 o_s=1+\sum_{q<s}B_q2^{L-q},\qquad j_s=o_s+2^{L-s}-1 .
\]
The record of the block is the binary-search path $j_1,\ldots,j_L$, and the label of $j_s$ is $1-B_s$ (if $B_s=0$ then $\mathfrak b\le j_s$, and if $B_s=1$ then $\mathfrak b\ge j_s+1$). The address $j_s$ depends only on $B_1,\ldots,B_{s-1}$. A threshold $h_\varsigma$ agrees with the whole path if and only if $\varsigma=\mathfrak b$, because the path separates all $2^L$ leaves. By convention, $L=0$ means that there are no path queries and a single trivial threshold.

The families of \cref{lem:low-realizable,lem:low-free-gate,lem:low-natural-tree}, and the comparison experiments used in their proofs, have the following structure, which we call an \emph{arm family}. There is a finite set $\mathcal A_{\rm all}$ of \emph{arms}; each arm $a$ is a block of depth $L_a\ge0$ on its own queries, with its own hidden leaf. A world consists of the vector $\mathbf B$ of all leaves and of further parameters $\varphi$ (winners, activities). Under the prior, the leaf bits are independent and uniform and $\mathbf B$ is independent of $\varphi$. Each record is generated in two steps. First a \emph{type} $\iota\in\mathcal A_{\rm all}\cup\{\varnothing\}$ is drawn from a fixed law that is the same in all worlds; the type $\iota=a$ names the arm $a$, so each arm has exactly one type. Then the record is a deterministic function of $\iota$, of $\varphi$ and, if $\iota\ne\varnothing$, of the leaf of the arm $\iota$, of the following form. If $\iota=\varnothing$, the record is a function of $\varphi$. If $\iota=a$, the record consists of an initial part whose queries and labels are functions of $(a,\varphi)$ (for instance a gate query; the initial part may be empty), followed by the binary-search path of the leaf of $a$, followed by a final part (further queries and the descriptor) that is a function of $a$, $\varphi$ and that leaf. Thus in each world every arm has one fixed record. We say that the $s$-th bit of the leaf of $a$ is \emph{displayed} when the pair at level $s$ of the path of $a$ is displayed. For the analysis we realize the types by a sequence $\iota_1,\iota_2,\ldots$ of independent draws from the type law, made before learning: the $n$-th reset, if it takes place, has type $\iota_n$, and the other draws are not used. Let $\Xi$ consist of $\varphi$, the whole sequence $(\iota_n)_{n\ge1}$ and the learner's private randomness. Under the prior, $\mathbf B$, $\varphi$, the sequence $(\iota_n)_{n\ge1}$ and the private randomness are independent; in particular $\mathbf B$ is independent of $\Xi$. This realization changes neither the law of the experiment nor what the learner observes.

\begin{lemma}[Fresh bits and charging to distinct pilots]\label{lem:fresh-bits}
In an arm family:
\begin{enumerate}[label=(\alph*)]
\item conditionally on $\Xi$ together with the displays of the first $n$ resets, for any $n$, or together with the complete transcript of a run that terminates, the leaf bits that have not been displayed are independent and uniform;
\item at a reset of type $a\ne\varnothing$, the number of bits of $a$ displayed for the first time has conditional expectation at most $2$ given $\Xi$ and the earlier displays, and it is $0$ if the pilot errs in the initial part of the record of $a$;
\item a fixed function displays the same prefix on all resets of the same type $a$, so only its first such reset can display new bits of $a$;
\item for a set $\mathcal A$ of arms let $U_{\mathcal A}$ be the number of pairs $(a,s)$ with $a\in\mathcal A$ and $1\le s\le L_a$ such that the $s$-th bit of $a$ is displayed on at least one reset (each such pair counts once, however often its bit is displayed), and let $X_{\mathcal A}$ be the number of resets of a type $a\in\mathcal A$ whose pilot $g$ passes the initial part of the record of $a$ and was not the pilot of an earlier reset of type $a$. Then $\E U_{\mathcal A}\le2\E X_{\mathcal A}$. In particular $\E U_{\mathcal A}$ is at most twice the expected number of resets of a type in $\mathcal A$ whose pilot passes the initial part of its record, and at most twice the expected number of pairs $(g,a)$, $a\in\mathcal A$, such that the played function $g$ passes the initial part of $a$.
\end{enumerate}
\end{lemma}

\begin{proof}
Fix $\Xi$. The transcript is then a deterministic function of $\mathbf B$: the pilot of each reset is a function of the private randomness and of the earlier displays, the type of the $n$-th reset is the coordinate $\iota_n$ of $\Xi$, and the record is a function of the type, of $\varphi$ and of the leaf of the arm named by the type. We show by induction on $n$ that for every history $h_n$ of the first $n$ displays, the set of leaf vectors that produce $h_n$ is a \emph{cylinder}: there are numbers $s_a\ge0$ and values of the first $s_a$ bits of every arm $a$, determined by $(\Xi,h_n)$, such that $\mathbf B$ produces $h_n$ if and only if its bits take these values. For $n=0$ take all $s_a=0$. Let the claim hold for $n$, and let $g$ and $\iota$ be the pilot and the type of reset $n+1$; both are determined by $(\Xi,h_n)$. If $\iota=\varnothing$, the display is determined by $(\Xi,h_n)$, so the set for $h_{n+1}$ is the cylinder for $h_n$ or empty. Let $\iota=a$ be an arm. The display of the initial part is determined by $(\Xi,h_n)$, and if $g$ errs there nothing else is displayed. Otherwise the display continues along the path of $a$. At level $s$ it shows the address $j_s$, which is determined by $B_1,\ldots,B_{s-1}$ of $a$, and the label $1-B_s$, and it stops at the first level at which $g$ errs. Hence, given $(\Xi,h_n)$, the displayed path and the first $q$ bits of $a$ determine each other, where $q$ is the last displayed level ($q=L_a$ if $g$ agrees with the whole path). If $g$ agrees with the whole path, the final part is a function of $\Xi$ and of the leaf of $a$, which the displayed path already determines. So the set for $h_{n+1}$ is the cylinder for $h_n$ intersected with the condition that the first $q$ bits of $a$ take the displayed values, again a cylinder or empty. Since $\mathbf B$ is uniform and independent of $\Xi$, its conditional law given $\Xi$ and $h_n$ is uniform on the cylinder. The decision to stop after $n$ resets is determined by $(\Xi,h_n)$, so the same holds given $\Xi$ and a complete transcript with $n$ resets. This proves (a).

For (b), let $s$ bits of $a$ be displayed before reset $n+1$. To display $j\ge1$ new bits, $g$ must pass the initial part and agree with the path at levels $s+1,\ldots,s+j-1$. Given $\Xi$, $h_n$ and $B_{s+1},\ldots,B_{s+i-1}$, the address $j_{s+i}$ and the prediction of $g$ there are fixed, while $B_{s+i}$ is uniform by (a); so $g$ agrees at level $s+i$ with conditional probability $1/2$. Hence the number $G$ of new bits satisfies $\Pr(G\ge j\mid\Xi,h_n)\le2^{-(j-1)}$, and $\E[G\mid\Xi,h_n]\le\sum_{j\ge1}2^{-(j-1)}=2$.

For (c), in a world all resets of type $a$ have the same record, consisting of the fixed initial part, path and final part of $a$, and a fixed function stops at the same position whenever it meets the same record. So if $g$ was the pilot of an earlier reset of type $a$, then either it erred in the initial part then and errs there again, or it displays the same prefix again.

For (d), $U_{\mathcal A}$ is the sum over resets of the number of bits of arms in $\mathcal A$ displayed for the first time at that reset. Whether reset $n+1$ counts in $X_{\mathcal A}$ is determined by $(\Xi,h_n)$, and so is the event that reset $n+1$ takes place. By (b) and (c), the number of bits of arms in $\mathcal A$ displayed for the first time at reset $n+1$ is zero unless the reset counts in $X_{\mathcal A}$, and its conditional expectation given $(\Xi,h_n)$ is at most twice the indicator that it counts. Summing over $n$ by Tonelli's theorem, which needs no bound on $N$, gives $\E U_{\mathcal A}\le2\E X_{\mathcal A}$. The two further bounds follow because $X_{\mathcal A}$ is at most each of the two counts.
\end{proof}

For laws $P,Q$ let $\KL(P\Vert Q)$ be the relative entropy and $\TV(P,Q)=\sup_A|P(A)-Q(A)|$.

\begin{lemma}[Testing with padded transcripts]\label{lem:stopped-kl}
Suppose two worlds share the learner's action and stopping rules and the law of its private randomness, and that before stopping the conditional relative entropy of the next observation is at most $\kappa$. If $\E_PN<\infty$, the complete stopped transcripts, including the output, satisfy
\[
 \KL(P_{\rm tr}\Vert Q_{\rm tr})\le\kappa\,\E_PN,\qquad\TV(P_{\rm tr},Q_{\rm tr})\le\sqrt{\kappa\,\E_PN/2}.
\]
Moreover, if a hidden vector is uniform on $\{-1,1\}^d$ and the transcript laws of every two vectors that differ in one coordinate are within total variation $\tau_0$, then every decoder has prior-average fraction of incorrect coordinates at least $(1-\tau_0)/2$.
\end{lemma}

\begin{proof}
After stopping, fill every later position of the transcript with a fixed symbol. Given a common finite past, the private randomness and the chosen actions have the same conditional kernels in both worlds and contribute no relative entropy; padded positions contribute none either. The chain rule therefore gives, for the first $m$ positions,
\[
 \KL(P_{[m]}\Vert Q_{[m]})\le\kappa\sum_{t=1}^m\Pr_P(N\ge t)=\kappa\,\E_P\min(N,m).
\]
Relative entropy on an increasing sequence of $\sigma$-fields converges to the relative entropy on the generated $\sigma$-field, as follows from its variational formula by approximating bounded measurable functions with their conditional expectations given the finite prefixes. Letting $m\to\infty$ gives the first bound. Appending the output is a common randomized kernel of the stopped transcript and cannot increase relative entropy; equivalently, the output randomness can be included in the initial private seed.

For the second bound we recall Pinsker's inequality. For an event $A$, data processing follows from the log-sum inequality applied on $A$ and on its complement. For Bernoulli parameters $p,q$, the binary relative entropy has second derivative $1/(p(1-p))\ge4$ in $p$ and vanishes together with its first derivative at $p=q$, so it is at least $2(p-q)^2$. Taking events that approach the total variation gives $\KL\ge2\TV^2$. Finally, for two vectors that differ only in coordinate $i$, the probabilities of decoding that coordinate incorrectly under the two laws sum to at least $1-\tau_0$. Pair the $2^d$ vectors along coordinate $i$ and average over $i$.
\end{proof}

\begin{remark}[Exact dimension and padding]\label{rem:padding}
A product of $d$ threshold blocks on disjoint query sets has VC dimension exactly $d$: one query per block is shattered, and two queries of one block cannot be shattered by thresholds. The same holds for a block in which a function selects one arm, accepts that arm's gate, and is positive on a final segment of that arm's path queries: within one arm the positive sets are nested, and points of different arms cannot both be positive. The product of $d$ such blocks again has dimension exactly $d$.

If a construction has a class of VC dimension one, add $d-1$ further queries on which the class takes all $2^{d-1}$ combinations of values independently of the rest. The result has VC dimension exactly $d$: a shattered set restricts to a shattered set on the original domain, so it has at most one original point and at most $d-1$ added points, and one shattered original point together with all added points is shattered. Fix arbitrary labels on the added queries and never sample them. They reveal no information, and the variants of a function that differ only there count as distinct functions, which can only increase $V$; on the supported queries every variant behaves like the original function.
\end{remark}

\subsection{A realizable floor for both resources}

\begin{lemma}[Rare threshold paths]\label{lem:low-realizable}
For all $d,D\ge1$ and $0<\eps\le1/16$ there is a finite family with the natural loss, a class of VC dimension exactly $d$ and $F^*=R^*=0$, on which every learner that succeeds with probability at least $4/5$ in every world has
\[
 \sup\E N\ge\frac{dD}{40\eps},\qquad\sup\E V\ge\frac D5 .
\]
\end{lemma}

\begin{proof}
Take $d$ blocks of depth $D$ on disjoint query sets, with leaves $\mathfrak b_1,\ldots,\mathfrak b_d$, and let the class be the product of thresholds. With probability $\theta=8\eps\le1/2$ a record chooses a uniform block and is its path, of length $D$; otherwise it is empty. Descriptors are constant. The product of the true thresholds is perfect, so $F^*=R^*=0$, and a member of the class has risk exactly $\theta$ times the fraction of blocks whose threshold is wrong. A successful output therefore has the correct threshold in at least $7/8$ of the blocks.

Give the leaves independent uniform priors. This is an arm family whose arms are the $d$ blocks and whose types are the blocks and $\varnothing$, with empty initial parts and no further parameters. Runs terminate almost surely: for the reset bound we may assume $\E N<\infty$, and for the pilot bound this is the convention of \cref{thm:lower}. Let $U$ be the number of distinct leaf bits displayed, that is, $U_{\mathcal A}$ of \cref{lem:fresh-bits}(d) for the set $\mathcal A$ of all blocks, and $J$ the number of blocks whose path is completely displayed; then $J\le U/D$. The output is a function of $\Xi$ and of the transcript, so by \cref{lem:fresh-bits}(a) the threshold it chooses for an incompletely displayed block is correct with conditional probability at most $1/2$. Hence the expected fraction $C$ of correct blocks satisfies
\[
 \frac7{10}=\frac45\cdot\frac78\le\E C\le\frac12+\frac{\E J}{2d}\le\frac12+\frac{\E U}{2dD},
\]
so $\E U\ge2dD/5$. By \cref{lem:fresh-bits}(d) applied to all arms, $\E U\le2\theta\,\E N$, since the type of each reset names an arm with probability $\theta$, independently of the event that the reset takes place; and $\E U\le2d\,\E V$, since a played function and an arm form at most $dV$ pairs. Substituting gives the bounds for the prior means, hence for some world; if an expectation is infinite there is nothing to prove. The argument includes $D=1$.
\end{proof}

\subsection{Free gates: the natural-ceiling term at zero general optimum}

Comparisons between a hidden winner and a null environment are standard information arguments; related contextual combination locks appear in \citet{jia2023}. The proof below checks the precise proper, frozen-pilot interface and its resource accounting.

\begin{lemma}[A diluted free-gate family]\label{lem:low-free-gate}
Let $0<\eps\le1/16$, $f\ge16\eps$, $d,D\ge1$, and put
\[
 K=\Bigl\lfloor\frac f{4\eps}\Bigr\rfloor\ge4,\qquad L=D-1,\qquad L_+=\max(1,L).
\]
There is a family with a class of VC dimension exactly $d$, one fixed binary loss, $F^*=f(1-1/K)\le f$ and $R^*=0$, on which every learner that succeeds with probability at least $4/5$ in every world has
\[
 \sup\E N\ge\frac{11\,dK^2L_+}{800f},\qquad\sup\E V\ge\frac{19\,KL_+}{1600}.
\]
\end{lemma}

\begin{proof}
\emph{Class and worlds.} Each of $d$ blocks has $K$ arms. Arm $(i,c)$ has a gate query $e_{ic}$ with label one and a threshold block of depth $L$ with its own hidden leaf $\mathfrak b_{ic}$. A member of the class chooses in each block one arm $a_i$ and one threshold, predicts one at the gate of $a_i$, uses the chosen threshold on that arm, and predicts zero at every query of the other arms of the block. (For $L=0$ it just chooses a gate.) By \cref{rem:padding} the VC dimension is exactly $d$.

A world specifies one good arm $C_i$ in each block and the leaves of all arms. A record is empty, with the constant descriptor $0$, with probability $1-f$; otherwise it chooses $(i,c)$ uniformly from $[d]\times[K]$, asks the gate $e_{ic}$, follows the path of arm $(i,c)$, and has descriptor $\1\{c\ne C_i\}$. All labels are fixed functions of the queries within a world. The loss is
\[
 \lambda=\begin{cases}
 0 & \text{at a discrepancy at the gate},\\
 1 & \text{at a later discrepancy},\\
 \text{the descriptor} & \text{at a survival},
 \end{cases}
\]
and zero on the empty record. A gate discrepancy displays no path or descriptor, and a path discrepancy displays its label but not the descriptor. The loss is bounded, prefix-observable and the same for all worlds.

A direct computation gives, for a member $h$ of the class,
\begin{equation}\label{eq:low-free-risk}
 R(h)=\frac f{dK}\,\#\{i:\ a_i\ne C_i\ \text{or the threshold on }a_i\text{ is wrong}\},
\end{equation}
with the threshold condition omitted if $L=0$. Hence $R^*=0$. The natural loss is minimized by using a correct threshold on the chosen arm of every block, while the other arms fail at their gates, so $F^*=f(1-1/K)$. Since $K\le f/(4\eps)$, a successful output has the good arm in at least $3/4$ of the blocks.

Give the good arms independent uniform priors and the leaf bits independent uniform priors. The family is an arm family whose arms are the pairs $(i,c)$, whose types are these pairs and $\varnothing$, and whose further parameters are the good arms; the initial part of a record of arm $(i,c)$ is its gate, and its final part is its descriptor. For a fixed block $i$, the \emph{null experiment} is the experiment in which every arm of block $i$ has descriptor one, while all other parameters keep their priors; it is again an arm family, with the same arms and types. The null experiment need not satisfy $R^*=0$, so no assumption on the learner's costs is made there; instead the procedure is capped before it is run in it.

\emph{The null coupling.} Let a procedure have deterministic caps. Run it in the null experiment for block $i$ with its usual private randomness and record types, draw $C_i$ uniformly and independently, and couple with the original experiment using the same leaves, types and randomness. Let $A_i$ be the number of distinct arms of block $i$ whose whole path and descriptor were displayed in the null run. The two runs coincide until the first completed visit to arm $C_i$. If $C_i$ is not among the completed arms of the null run, the runs and their outputs coincide. Writing $\widehat a_i$ for the arm chosen in block $i$ by the output,
\begin{equation}\label{eq:low-null-coupling}
 \Pr(\widehat a_i=C_i)\le\frac{\E^0A_i+1}K .
\end{equation}
The first term bounds the probability that the independent uniform arm $C_i$ is completed in the null run, and the second the probability that the null output names it. We do not claim that $C_i$ is uniform given a miss. The term $+1$ allows an unplayed final answer.

\emph{Expected resets.} Let $n=\sup\E N$ over the family, and assume it is finite. Stop the procedure after $N_0=\lfloor10n\rfloor$ resets, returning any member of the class on truncation. In every world $\Pr(N>N_0)\le\E N/(N_0+1)\le1/10$, so the truncated procedure succeeds with probability at least $7/10$, and its prior-mean fraction of correct arms is at least $21/40$. The cap $N_0$ also holds in every null experiment. A member of the class passes the gate of exactly one arm of block $i$, and the type of each reset names that arm with probability $f/(dK)$, independently of the event that the reset takes place. If $L\ge1$, a completed arm needs all its $L$ bits displayed, so \cref{lem:fresh-bits}(d), applied to the arms of block $i$, gives $\E^0A_i\le2N_0f/(dKL)$; if $L=0$, a completed arm needs a matching gate, so $\E^0A_i\le N_0f/(dK)$. In both cases
\[
 \Pr(\widehat a_i=C_i)\le\frac1K+\frac{2N_0f}{dK^2L_+}.
\]
Averaging over $i$ and using $21/40-1/K\ge11/40$ gives $10n\ge N_0\ge11dK^2L_+/(80f)$, which is the reset bound. The floor and the denominator $N_0+1$ avoid an additive rounding loss.

\emph{Expected distinct pilots.} Let $v=\sup\E V$ over the family, and assume it is finite. Stop the procedure immediately before it would play an $(m+1)$-st distinct function, where $m=\lfloor10v\rfloor$; in every world this happens with probability at most $v/(m+1)\le1/10$. Impose in addition a common finite reset cap $N_1$ with $\sup_\omega\Pr_\omega(N>N_1)\le1/20$ over the worlds $\omega$ of the family; such an $N_1$ exists because the family is finite and every run terminates almost surely. No finite expected reset count is needed. The doubly capped procedure succeeds with probability at least $4/5-1/10-1/20=13/20$ in every world, so its prior-mean fraction of correct arms is at least $39/80$.

In the null experiment for block $i$, a played function passes the gate of exactly one arm of block $i$, and at most $m$ functions are played. \Cref{lem:fresh-bits}(d), applied to the arms of block $i$, therefore gives $\E^0A_i\le2m/L$ if $L\ge1$; if $L=0$, each distinct pilot can complete at most one arm, so $A_i\le m$. Hence \eqref{eq:low-null-coupling} is at most $1/K+2m/(KL_+)$. Averaging and using $39/80-1/K\ge19/80$ gives $10v\ge m\ge19KL_+/160$, which is the pilot bound. Dilution changes how long one waits for a matching gate, but not the information obtained per distinct pilot.
\end{proof}

\subsection{A hidden terminal mean with zero natural optimum}

\begin{lemma}[The general-risk term]\label{lem:low-terminal}
Let $0<\eps\le1/16$, $r\ge16\eps$ and $d,D\ge1$. There is a family with a class of VC dimension exactly $d$, $F^*=0$ and $R^*\le r$, on which every learner that succeeds with probability at least $4/5$ in every world has $\sup\E N\ge dr/(16384\eps^2)$.
\end{lemma}

\begin{proof}
Let $\Delta=4\eps/r\le1/4$. There are $d$ queries, all with label zero, and the class consists of all $2^d$ functions on them. A record is empty with probability $1-r$; otherwise it chooses $i$ uniformly and asks query $i$. Predicting one there is a discrepancy with loss $1/2$. Predicting zero survives and displays a terminal bit $Y\sim\Ber(1/2+\sigma_i\Delta)$, which is the loss; $\sigma\in\{-1,1\}^d$ is the hidden vector of the world. The empty record has loss zero. The loss rule is fixed: the unknown signs affect the record law, not the loss rule or the labels. The records have length at most one, so the family is admissible for every $D\ge1$.

The all-zero function has no discrepancy, so $F^*=0$. If $m_-$ signs are negative, the optimum predicts one exactly at the positive signs, and
\[
 R^*=\frac r2-\frac{r\Delta m_-}d\le r,\qquad
 R(h)-R^*=r\Delta\,\frac{\#\{i:\ h\text{ decides coordinate }i\text{ wrongly}\}}d .
\]
Success therefore requires at most $1/4$ of the coordinates to be wrong.

Give the learner the complete record, including $Y$, on every reset; this stronger experiment can simulate the stopped one by discarding information. For two sign vectors that differ in one coordinate, the relative entropy of one reset is at most $(r/d)\cdot2\Delta\log\frac{1+2\Delta}{1-2\Delta}$ in either direction, and
\[
 2\Delta\log\frac{1+2\Delta}{1-2\Delta}\le16\Delta^2,
\]
because $\log\frac{1+x}{1-x}=\int_0^x\frac{2\,du}{1-u^2}\le\frac{8x}3$ for $x=2\Delta\le1/2$. Let $n$ be the largest expected number of resets over the family. If $n<d/(1024r\Delta^2)$, \cref{lem:stopped-kl} gives total variation less than $\sqrt{1/128}<1/8$ for every neighboring pair, and the prior-average fraction of wrong coordinates is at least $7/16$. Uniform success bounds that fraction by $\frac45\cdot\frac14+\frac15=\frac25<\frac7{16}$, a contradiction. Hence $n\ge d/(1024r\Delta^2)=dr/(16384\eps^2)$. Data-dependent stopping, output randomization and private coins are covered by \cref{lem:stopped-kl}.
\end{proof}

\begin{proof}[Proof of \cref{thm:lower}]
\Cref{lem:low-realizable} applies for every pair of ceilings, because its family has $F^*=R^*=0$, and gives $\Omega(dD/\eps)$ resets and $\Omega(D)$ pilots. If $f\ge16\eps$, \cref{lem:low-free-gate} satisfies both ceilings (also when $r=0$), and since $K\ge f/(8\eps)$ and $L_+\ge D/2$ it gives $\Omega(dDf/\eps^2)$ resets and $\Omega(Df/\eps)$ pilots. If $f<16\eps$, these terms are at most sixteen times the realizable ones. If $r\ge16\eps$, \cref{lem:low-terminal} satisfies both ceilings (it has $F^*=0$) and gives $\Omega(dr/\eps^2)$ resets. If $r<16\eps$, this term is at most $16dD/\eps$. For each resource take the largest of the applicable lower bounds; the maximum of three (or two) nonnegative numbers is at least a third (or a half) of their sum. This proves \cref{thm:lower} with a universal constant. A covering $\eps$-certificate implies $R(\widehat h)\le R^*+\eps$, so the stronger output requirement does not avoid these bounds, and per-world success $1-\delta$ with $\delta\le1/5$ implies their hypothesis. They concern ceilings on two separate infima, not prescribed exact optima, and do not supply an optimizing hypothesis (see also \cref{rem:lower-reading}).
\end{proof}

\subsection{Natural discrepancy}

For the natural loss $R=F$, so the two ceilings amount to the single ceiling $\min(f,r)$, which this subsection denotes by $r$; the free-gate family of \cref{lem:low-free-gate} does not apply to this loss.

\begin{lemma}[Diluted active and null blocks]\label{lem:low-natural-tree}
Let $D\ge3$, $0<\eps\le1/128$ and $r\ge128\eps$, and put $k=D-2\ge1$ and $K=\lfloor r/(16\eps)\rfloor\ge8$. There is a finite natural-loss family with a class of VC dimension exactly $d$ and $F^*\le r$ on which every learner that succeeds with probability at least $4/5$ in every world satisfies, for the prior described below,
\[
 \E N\ge\frac{dkK^2}{16r},\qquad\E V\ge\frac{kK}{16}.
\]
\end{lemma}

\begin{proof}
Take $d$ blocks with $K$ gated arms each, every arm having its own threshold block of depth $k$, and the class of \cref{lem:low-free-gate}. Add two queries $t_0,t_1$ with fixed labels zero and one, at which every member of the class predicts one; these constant coordinates do not change the VC dimension. Each block is either \emph{null} or \emph{active}, and an active block has one winning arm $W_i$.

A record is empty with probability $1-r$. Otherwise it chooses a uniform block-arm pair, asks its gate (label one), follows its path of $k$ queries, and then asks $t_1$ if the arm is the winner of an active block, and $t_0$ otherwise. The descriptor is constant, and hard records have exactly $k+2=D$ queries. All labels are fixed; which final query occurs is part of the record law. The final query is hidden after any earlier discrepancy, including one at the last level of the path. A member of the class survives a hard record exactly when the record's arm is its chosen arm, its threshold is correct, and the arm is the winner of an active block. Hence, if $A$ blocks are active,
\begin{equation}\label{eq:low-natural-tree-risk}
 F^*=r\Bigl(1-\frac A{dK}\Bigr),\qquad
 F(h)-F^*=\frac r{dK}\,\#\{i\text{ active}:\ h\text{ misses the winner or its leaf in block }i\},
\end{equation}
and in particular $F^*\in[r(1-1/K),r]$.

Give the activities independent $\Ber(1/2)$ priors, the winners of active blocks independent uniform priors, and the leaves independent uniform priors. Call an arm \emph{probed} when its final query is displayed, which requires surviving its whole path, and let $S_i$ be the set of probed arms of block $i$. For a fixed $i$, couple a run in which block $i$ is forced to be null with a run in which it is forced to be active, with the winner drawn independently of the null run. Before termination the runs can differ only in the final query of the winning arm. Hence, writing $C_i$ for the event that the output has the winner and the correct leaf in block $i$,
\begin{equation}\label{eq:low-active-null}
 \Pr(C_i\mid i\text{ active})\le\frac{\E^0|S_i|+1}K ,
\end{equation}
where $\E^0$ refers to the run with block $i$ forced null. Both versions belong to the original family, so no out-of-promise budget assumption is involved; the coupling is valid for almost surely finite runs, and the $+1$ allows an unplayed final answer as in \eqref{eq:low-null-coupling}.

Let $\mathsf M$ be the number of active blocks whose winner or leaf is missed by the output. On the success event \eqref{eq:low-natural-tree-risk} gives $\mathsf M\le\eps dK/r\le d/16$, and always $\mathsf M\le d$, so $\E\mathsf M\le d/16+d/5=21d/80$. Since $\E A=d/2$, the expected number of active blocks identified correctly is at least $19d/80$, that is, $\sum_i\Pr(i\text{ active},C_i)\ge19d/80$. Multiplying \eqref{eq:low-active-null} by $K/2$ and summing over $i$,
\[
 \E\sum_{i:\,i\text{ is null}}|S_i|=\frac12\sum_i\E^0|S_i|\ge K\sum_i\Pr(i\text{ active},C_i)-\frac d2\ge d\Bigl(\frac{19K}{80}-\frac12\Bigr)\ge\frac{dK}8 ,
\]
using $K\ge8$. Every probed arm has displayed all $k$ of its bits, so the number $U$ of distinct leaf bits displayed ($U_{\mathcal A}$ of \cref{lem:fresh-bits}(d) for the set $\mathcal A$ of all arms) satisfies $\E U\ge dkK/8$. On the other hand, the family is an arm family whose arms are the block-arm pairs, whose types are these pairs and $\varnothing$, and whose further parameters are the activities and the winners, with the gates as initial parts and the final queries in the final parts; the forced-null and forced-active versions have the same arms and types. A member of the class passes the gate of exactly one arm in each block, and the type of each reset names one of these $d$ arms with probability $d\cdot r/(dK)=r/K$, independently of the event that the reset takes place. \Cref{lem:fresh-bits}(d) therefore gives $\E U\le(2r/K)\E N$ and $\E U\le2d\,\E V$. These inequalities prove both bounds; if an expectation is infinite there is nothing to prove.
\end{proof}

\begin{lemma}[Natural resets at horizons one and two]\label{lem:low-paired}
Let $D\in\{1,2\}$, $0<\eps\le1/16$ and $r\ge16\eps$. There is a natural-loss family with a class of VC dimension exactly $d$ and $F^*\le r$, on which every learner that succeeds with probability at least $4/5$ in every world has $\sup\E N>dr/(4096\eps^2)$.
\end{lemma}

\begin{proof}
Let $\gamma=4\eps/r\le1/4$. Block $i$ has two queries $x_{i0},x_{i1}$ with labels zero and one, fixed and known in all worlds. A member of the class predicts one common bit on both queries of each block, independently across blocks; the VC dimension is exactly $d$. A world has a hidden vector $\varpi\in\{0,1\}^d$. With probability $r$ a record selects a uniform block $i$ and then the query $x_{i\varpi_i}$ with probability $1/2+\gamma$, and the other query otherwise; with the remaining probability the record is empty. Records have length at most one, and every pilot sees the query and its label, even when it errs.

The best prediction in block $i$ is $\varpi_i$, and
\[
 F^*=r\Bigl(\frac12-\gamma\Bigr)\le r,\qquad F(h)-F^*=2r\gamma\,\frac{\#\{i:\ h(x_{i0})\ne\varpi_i\}}d .
\]
A uniformly successful learner therefore has expected wrong fraction at most $\eps/(2r\gamma)+1/5=13/40$. Neighboring worlds have one-reset relative entropy at most $16r\gamma^2/d$, by the Bernoulli computation in \cref{lem:low-terminal}. If every world had $\E N\le d/(256r\gamma^2)$, \cref{lem:stopped-kl} would give neighboring total variation at most $\sqrt{1/32}<1/4$, and the prior-average wrong fraction would exceed $3/8>13/40$, a contradiction. Hence some world has $\E N>d/(256r\gamma^2)=dr/(4096\eps^2)$. Properness is essential: a function outside the class could predict the known label on both queries.
\end{proof}

\begin{lemma}[Natural hidden arms at horizon two]\label{lem:low-hidden-arms}
Let $d\ge1$, $0<\eps\le1/16$ and $r\ge8\eps$. There is a finite natural-loss family of horizon two with a class of VC dimension exactly $d$ and $F^*\le r$, on which every learner that succeeds with probability at least $4/5$ in every world has $\sup\E V\ge r/(160\eps)$.
\end{lemma}

\begin{proof}
Let $K=\lfloor r/(2\eps)\rfloor\ge4$. The supported class consists of $h_1,\ldots,h_K$, where $h_i$ predicts one at $x_i$, zero at $x_j$ for $j\ne i$, and one at every second query $z_j$. In world $a\in[K]$ all $x_j$ have label one and $z_j$ has label $\1\{j=a\}$. A record is empty with probability $1-r$; otherwise it chooses a uniform $j$ and asks $x_j$ and then $z_j$. Descriptors are constant. The supported class has VC dimension one: a single first query is shattered, two first queries cannot both be positive, and the second queries are constant on the class. Apply the padding of \cref{rem:padding} to obtain dimension exactly $d$.

Only $h_j$ and its padded variants pass the first query of arm $j$, so $F(h_a)=r(1-1/K)$ and $F(h_i)=r$ for $i\ne a$. Since $r/K\ge2\eps$, success requires an output whose supported part is $h_a$.

Grant the learner, at the first selection of each arm, the answer to the question whether this arm is $a$, whether or not a matching record occurs. This dominates legal play: record types can be simulated independently, and the only world-dependent displayed label is that of the second query of a selected arm. Let $Q$ be the number of distinct arms selected; then $Q\le V$, also with padded variants. Fix all learner and simulation coins. A procedure that probes at most $u<K$ arms succeeds in at most $u+1$ of the $K$ worlds: on its all-no path it probes at most $u$ arms, and every world whose winner is not among them follows the all-no path and receives the same output, which names at most one further arm. Almost-sure termination on the finite family makes this valid outside a common null set of coins. Averaging over the coins and truncating immediately before the $(u+1)$-st new arm, which changes the run only if $Q>u$, gives under the uniform prior on the winner
\[
 \Pr(\text{success})\le\frac{u+1}K+\Pr(Q>u)\le\frac{u+1}K+\Pr(V>u).
\]
Take $u=\lfloor K/2\rfloor$. For $K\ge4$ we have $(u+1)/K\le3/4$, so success probability $4/5$ forces $\Pr(V>u)\ge1/20$ and $\E V\ge(u+1)/20\ge K/40\ge r/(160\eps)$, where the last step uses $K\ge r/(4\eps)$. The final output is not charged as a training pilot.
\end{proof}

\begin{proof}[Proof of \cref{thm:natural-lower}]
(a) For $D\ge3$ and $r\ge128\eps$, \cref{lem:low-natural-tree} with $K\ge r/(32\eps)$ and $k=D-2\ge D/3$ gives $\Omega(dDr/\eps^2)$ resets. For $D\in\{1,2\}$ and $r\ge16\eps$, \cref{lem:low-paired} gives $\Omega(dr/\eps^2)=\Omega(dDr/\eps^2)$ resets. In the remaining cases $r+\eps\le129\eps$, and \cref{lem:low-realizable}, whose family has $F^*=0$, gives $dD/(40\eps)\ge dD(r+\eps)/(5160\eps^2)$. Taking the maximum with the realizable floor gives the sum $dD(r+\eps)/\eps^2$ up to a constant. All lemmas used are valid in the common range $\eps\le1/128$; some of them hold in larger ranges.

(b) For $D\ge3$ and $r\ge128\eps$, \cref{lem:low-natural-tree} gives $\E V\ge kK/16=\Omega(Dr/\eps)$. For $D=2$ and $r\ge8\eps$, \cref{lem:low-hidden-arms} gives $\Omega(r/\eps)=\Omega(Dr/\eps)$. In the remaining cases $1+r/\eps\le129$, and the realizable floor $D/5$ suffices. Taking the maximum with $D/5$ gives the claim.

(c) Let $D=1$. Every nonempty record displays its single query and its label whatever the pilot, and empty records have loss zero, so on every reset the natural loss of every candidate is observed. Play one fixed member of the class on all resets, for instance an empty-list representative. The observations form an independent sample, and the natural losses of all candidates are functions of it. Let $\widehat F_n$ be the empirical natural loss. For $nt^2\ge2$,
\begin{equation}\label{eq:low-one-pilot}
 \Pr\Bigl(\sup_{h\in\cH}|\widehat F_n(h)-F(h)|>t\Bigr)\le4(2n+1)^de^{-nt^2/8}.
\end{equation}
Indeed, on the event in \eqref{eq:low-one-pilot} select a witness of the deviation, measurably by \cref{ass:measurable}; an independent copy of the sample has empirical mean within $t/2$ of the witness's mean with probability at least $1/2$, by Chebyshev's inequality and variance at most $1/(4n)$. So the probability in \eqref{eq:low-one-pilot} is at most twice the probability that the two empirical means differ by more than $t/2$ for some candidate. Conditionally on the $n$ pairs of observations, exchange the two members of each pair independently. For one pattern of losses, Hoeffding's inequality for the resulting independent signed differences bounds this event by $2e^{-nt^2/8}$. By \cref{lem:vc-traces}(i) there are at most $(2n+1)^d$ patterns of the class on the pooled queries (empty records add none), and a union bound proves \eqref{eq:low-one-pilot}.

Take $t=\eps/2$ and $n$ finite with $n\ge8/\eps^2$ and $4(2n+1)^de^{-n\eps^2/32}\le\delta$. Let $q=\min_h\widehat F_n(h)$, attained by a member returned by the oracle on the displayed queries. On the complementary event $\sup_{h\in\cH}|\widehat F_n(h)-F(h)|\le t$, which has probability at least $1-\delta$, both $F^*$ and the risk of the minimizer lie in $[\max(0,q-t),\min(1,q+t)]$; for $F^*$ use a sequence of functions whose risks approach $F^*$. This interval has width at most $2t=\eps$ on every history, and the learner uses one training pilot. A nonzero constant pilot floor is still necessary in the worst case, by \cref{lem:low-realizable} with $D=1$.
\end{proof}

\begin{remark}[Boundary cases]\label{rem:low-scope}
For general losses the $f$ term survives at $r=0$: the free-gate family has an attained zero general optimum and a positive natural optimum. At $f=r=0$ the remaining scales are $dD/\eps$ resets and $D$ pilots. For the natural loss, $r=0$ forces a zero natural infimum, which is attained in the finite classes used here, so there is no agnostic $f$ term to import. No lower-bound assertion is made for $d=0$ (see \cref{rem:lower-reading}), for $D=0$ or for accuracies outside the stated lower-bound ranges, and the reset lower bounds are not lower bounds on the number of displayed labels. The constructions establish the stated proper, frozen-pilot, stopped-record lower bounds; we do not claim novelty of their individual information-theoretic ingredients.
\end{remark}

\end{document}